\documentclass[pdflatex,sn-mathphys-num]{sn-jnl}

\usepackage{graphicx}%
\usepackage{multirow}%
\usepackage{amsmath,amssymb,amsfonts}%
\usepackage{amsthm}%
\usepackage{mathrsfs}%
\usepackage[title]{appendix}%
\usepackage{xcolor}%
\usepackage{colortbl}%
\definecolor{SeaGreen3}{RGB}{46,139,87}
\usepackage{textcomp}%
\usepackage{manyfoot}%
\usepackage{booktabs}%
\usepackage{algorithm}%
\usepackage{algorithmicx}%
\usepackage{algpseudocode}%
\usepackage{listings}%
\usepackage{capt-of}
\usepackage{rotating}
\usepackage{longtable}

\theoremstyle{thmstyleone}%
\theoremstyle{thmstyletwo}%

\theoremstyle{thmstylethree}%

\begin{document}

\title[Article Title]{Easy Peeler: Towards Efficient Modular Medical Image Registration}


\author[1]{\fnm{Xi} \sur{Jia}}

\author[1]{\fnm{Fuping} \sur{Wu}}

\author[1]{\fnm{Rima} \sur{Mustafa}}

\author[2]{\fnm{Mateusz} \sur{C. Florkow}}

\author[3]{\fnm{Tobias} \sur{H. Dovmark}}

\author[4,5]{\fnm{Katherine} \sur{R. Bull}}

\author[1]{\fnm{Jemma} \sur{C. Hopewell}}

\author[6]{\fnm{Joanna} \sur{M.M. Howson}}

\author[1,7]{\fnm{Thomas} \sur{E. Nichols}}

\author[1]{\fnm{Bart\l omiej} \sur{W. Papie\.z}}
\equalcont{The corresponding authors: \url{xi.jia@psych.ox.ac.uk};\url{bartlomiej.papiez@bdi.ox.ac.uk}}

\affil[1]{\orgdiv{Nuffield Department of Population Health, Big Data Institute, Li Ka Shing Centre for Health Information and Discovery}, \orgname{University of Oxford}, \orgaddress{\street{Old Road Campus}, \city{Oxford}, \postcode{OX3 7LF}, \country{United Kingdom}}}

\affil[2]{\orgdiv{Novo Nordisk R$\&$D Digital Hub London}, \orgname{Novo Nordisk}, \orgaddress{\street{1 Pancras Square}, \city{London}, \postcode{N1C 4AG}, \country{United Kingdom}}}


\affil[3]{\orgdiv{Computational Real-World Evidence}, \orgname{Novo Nordisk}, \orgaddress{\street{Novo Alle 1}, \postcode{2880 Bagsværd}, \country{Denmark}}}
\affil[4]{\orgdiv{Nuffield Department of Medicine}, \orgname{University of Oxford}, \orgaddress{\street{Old Road Campus}, \city{Oxford}, \postcode{OX3 7BN}, \country{United Kingdom}}}
\affil[5]{\orgdiv{Oxford Kidney Unit}, \orgname{Churchill Hospital}, \orgaddress{\country{United Kingdom}}}
\affil[6]{\orgdiv{Novo Nordisk Research Centre Oxford}, \orgname{Novo Nordisk}, \orgaddress{\street{Innovation Building, Old Road Campus}, \city{Oxford}, \postcode{OX3 7LF}, \country{United Kingdom}}}
\affil[7]{\orgdiv{Centre for Integrative Neuroimaging, FMRIB, Nuffield Department of Clinical Neurosciences}, \orgname{University of Oxford}, \orgaddress{\postcode{OX3 9DU}, \country{UK}}}

\abstract{
Deformable image registration estimates dense spatial correspondences between images and underpins longitudinal analysis, multimodal image fusion, atlas construction, population studies, treatment planning and image-guided intervention. Despite its broad utility, developing specialised registration solutions remains challenging because performance depends strongly on anatomy, imaging modality, dataset characteristics and available computational resources.

We introduce Easy Peeler, a modular image registration framework that decouples image representation from deformation estimation. Easy Peeler combines a fixed pretrained visual encoder with a lightweight transition--decoder that adapts extracted features for dense deformation prediction, eliminating the need to learn task-specific visual representations from scratch.

Across ten public biomedical image registration datasets spanning multiple anatomies, modalities and dimensionalities, Easy Peeler matched or exceeded widely used learning-based baselines while reducing training compute in most settings. These results demonstrate that pretrained visual representations can serve as a reusable foundation for dense correspondence, enabling efficient adaptation across heterogeneous registration tasks.

We release Easy Peeler as an open and reproducible framework designed to make high-performance medical image registration more accessible. By reducing task-specific training requirements and supporting the reuse of pretrained representations, Easy Peeler provides a practical approach to efficient and modular registration across biomedical imaging applications.
}


\maketitle

\section{Introduction}
\label{sec:intro}

Establishing dense spatial correspondences between images is a fundamental problem in medical image analysis, enabling multimodal image fusion, longitudinal analysis, atlas construction, population studies, treatment planning and image-guided interventions.

The rapid expansion of population-scale longitudinal imaging cohorts, including UK Biobank \cite{allen2024prospective}, large-scale brain research initiatives \cite{namburete2023normative, xie2025multi}, together with the increasing integration of multimodal imaging data in clinical trials \cite{ganjgahi2025ai}, has intensified the need for accurate, scalable and automated registration methods.

Classical registration methods formulate registration as an iterative optimisation problem that balances image similarity against spatial regularity \cite{rueckert2006diffeomorphic,beg2005computing,ashburner2007fast,vercauteren2009diffeomorphic,sotiras2013deformable}. Although these methods remain widely used, their performance and computational cost can depend strongly on the selected similarity measure, transformation model, regularisation, and optimisation strategy \cite{schnabel2016advances}. Deep learning-based methods substantially reduce inference time by replacing iterative optimisation with neural networks that directly predict deformation fields from image pairs \cite{balakrishnan2019voxelmorph,chen2022transmorph,jia2022u,chen2025survey,tian2024unigradicon}. However, most existing approaches learn image representations and deformation estimation jointly within task-specific architectures. Consequently, changes in anatomy, modality, or dimensionality commonly require a new network to be designed or trained from scratch.

This conventional formulation implicitly treats representation learning and deformation estimation as inseparable, although they address distinct problems. Representation learning determines which visual structures should be compared, whereas deformation estimation determines how those structures should be spatially aligned. The emergence of foundation models \cite{oquab2023dinov2,kirillov2023segment,ravi2024sam,simeoni2025dinov3}, particularly visual foundation models (VFMs), challenges the need to relearn both components for every registration task. Trained on large and diverse image collections, VFMs provide transferable representations that have proved effective in downstream medical imaging applications, including biomedical segmentation \cite{ma2024segment} and detection \cite{zhao2025foundation}. Their suitability for deep image registration models, however, remains unclear \cite{song2025dino}.

Here, we introduce a modular approach that lies between training a specialised end-to-end network for every registration task and using pretrained visual models as generic, unadapted feature extractors. Our framework separates reusable representation learning from task-specific deformation estimation, thereby reducing the architectural design and training required when adapting registration to a new dataset. The approach follows three principles:
\begin{enumerate}
    \item Reuse pretrained visual representations. Image representation is provided by a pretrained encoder (e.g. VFM) whose parameters remain unchanged during registration training. This avoids repeatedly learning low- and intermediate-level visual features for each anatomy, modality and dimensionality. 
    \item Adopt a common lightweight deformation pipeline. An architecturally identical transition--decoder filters, prompts, and fuses image features, learns a displacement-oriented representation and reconstructs a dense deformation field. This design limits task-specific architectural changes while maintaining a clear separation between representation and deformation learning.    
    \item Determine only the remaining task-dependent choices empirically. A compact set of candidate feature representations is evaluated on held-out validation data, and the feature achieving the highest registration accuracy is selected. Empirical adaptation is therefore focused on feature selection and optimisation settings rather than complete network redesign.
\end{enumerate}

We implemented these strategies in Easy Peeler (Fig.~\ref{fig:visual_results_and_easy_peeler}), a modular registration framework developed and evaluated across ten public biomedical imaging datasets spanning fluorescence microscopy, cardiac Magnetic Resonance Imaging (MRI), brain MRI, abdominal MRI, lung Computed Tomography (CT), abdominal CT, and multimodality CT--MR registration. Easy Peeler supports two- and three-dimensional registration, intra- and inter-subject alignment, and mono- and multimodal imaging. In contrast to conventional end-to-end networks, only the lightweight transition--decoder is trained; the visual encoder remains frozen, and full-resolution deformation fields are reconstructed using a parameter-free decoding operation.

Easy Peeler is modular rather than tied to a specific backbone. We systematically evaluated 47 feature representations from 14 foundation models and found that registration performance depends more strongly on the target task and feature depth than on model scale, pretraining domain or model generation. These results provide empirical guidance for narrowing the feature search space when adapting the framework to new datasets.

Across the ten evaluated datasets, Easy Peeler matched or exceeded widely used learning-based registration baselines while reducing training compute in most settings. Its modular design enables the integration of existing and emerging foundation models without redesigning the deformation pipeline while concentrating task-specific learning on the mapping from visual features to dense correspondence. 
We release the code, configurations, and trained weights, providing a reproducible route to high-performance registration without requiring a specialised end-to-end architecture for every new application.

\noindent
\begin{minipage}{\textwidth}
    \centering
    \includegraphics[width=0.99\linewidth]{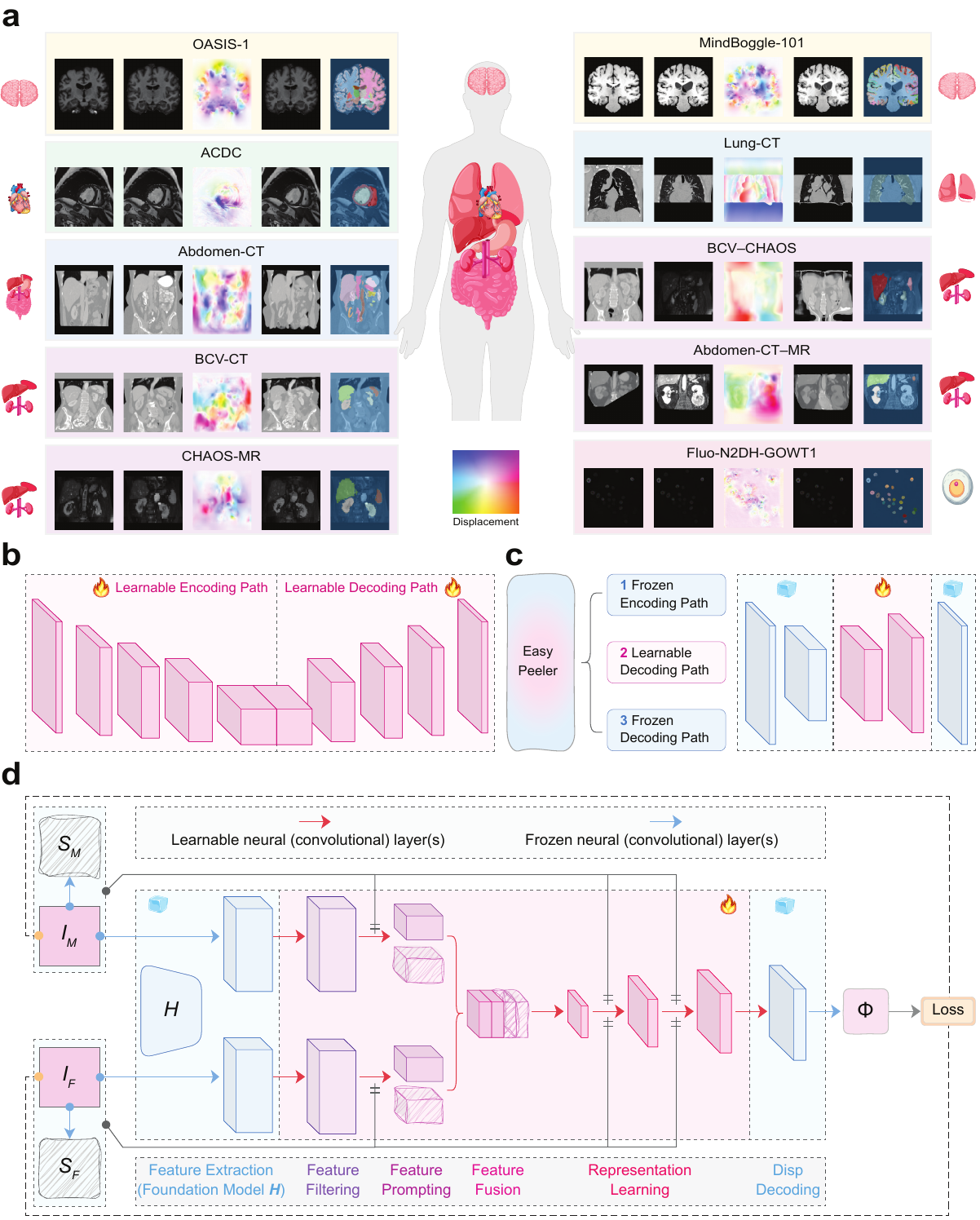}
\end{minipage}
\captionof{figure}{\textbf{Easy Peeler: a modular framework for reusable medical image registration.} \textbf{(a)} Representative examples from the ten public datasets used for evaluation, spanning fluorescence microscopy, cardiac MRI, brain MRI, lung CT, abdominal CT, abdominal MRI and cross-modality CT--MR registration. For each dataset, we show the moving image, fixed image, predicted displacement field, warped moving image and warped segmentation overlay. For 3D datasets, the central coronal slice is displayed. Displacement fields use the standard optical-flow colour coding \cite{baker2011database}. \textbf{(b)} Conventional deep registration jointly learns image representations and deformation estimation within a task-specific encoder--decoder architecture. \textbf{(c)} \textit{Easy Peeler} decouples these components by combining a frozen image encoder with a lightweight trainable transition--decoder and a parameter-free displacement reconstruction layer. Only the transition--decoder is optimised for a new registration task. \textbf{(d)} Overview of the Easy Peeler framework. A frozen foundation encoder ($\mathcal{H}$) extracts reusable image representations, which are adapted for registration through four trainable modules: feature filtering, feature prompting, feature fusion and displacement representation learning. A parameter-free displacement decoding layer reconstructs the full-resolution deformation field.
}
\label{fig:visual_results_and_easy_peeler}

\section{Results}
\subsection{Benchmarking established registration methods}

\begin{figure}[tb!]
    \centering
    \includegraphics[width=0.99\linewidth]{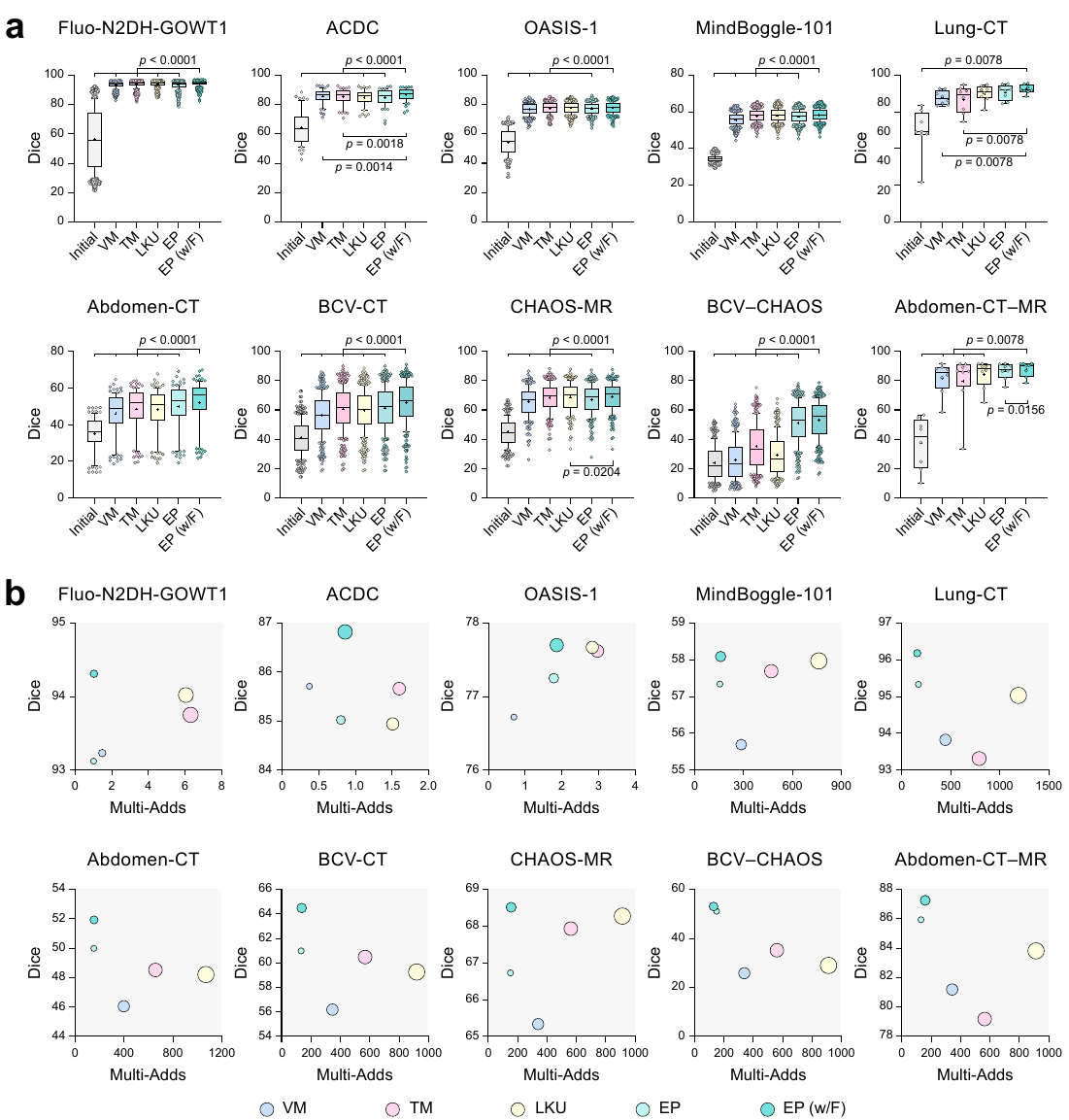}
    \caption{\textbf{Easy Peeler enables efficient registration without sacrificing accuracy.}
\textbf{(a)} Test-set Dice scores for Easy Peeler (using the best-performing frozen representation for each task) and four baselines: VoxelMorph, TransMorph, LKU-Net and Easy Peeler without pretrained features. Boxplots summarise performance across all test pairs; significance was assessed using paired Wilcoxon signed-rank tests. Sample sizes were: Fluo-N2DH-GOWT1 ($n=930$), ACDC ($n=50$), OASIS-1 ($n=400$), MindBoggle-101 ($n=870$), Lung-CT ($n=8$), Abdomen-CT ($n=90$), BCV-CT ($n=306$), CHAOS-MR ($n=210$), BCV--CHAOS ($n=270$) and Abdomen-CT--MR ($n=8$).
\textbf{(b)} Accuracy--compute trade-off for each benchmark. The vertical axis shows mean Dice, the horizontal axis training compute (Giga-Multi-Adds), and bubble size shows inference compute, including frozen feature extraction. Easy Peeler matched or exceeded the strongest baseline on most datasets while requiring substantially less training compute than TransMorph and LKU-Net in all benchmarks and than VoxelMorph in eight of ten benchmarks. Although inference efficiency depended on the chosen frozen encoder, the reduced training cost was consistent across tasks.
}

    \label{fig:performance}
\end{figure}

We benchmarked \textit{Easy Peeler} against three widely used learning-based registration frameworks: VoxelMorph \cite{balakrishnan2019voxelmorph}, TransMorph \cite{chen2022transmorph}, and LKU-Net \cite{jia2022u}, across ten public registration benchmarks spanning fluorescence microscopy, MRI and CT, multiple anatomical systems (brain, heart, lungs and abdomen), and both mono- and cross-modality registration tasks. All methods were evaluated under an identical protocol using three independent training seeds.
Dice served as the primary measure of registration accuracy (Fig.~\ref{fig:performance}), while surface distance, image similarity and Jacobian-based metrics were analysed separately to assess boundary alignment and deformation quality (Supplementary Note~\ref{sec:supp:benchmark}).

Across the ten benchmarks, the best-performing \textit{Easy Peeler} configuration achieved the highest average Dice score (Fig.~\ref{fig:performance}~\textbf{a}).
Improvements were particularly pronounced for abdominal CT and cross-modality CT--MR registration, whereas gains were smaller for OASIS-1 and CHAOS-MR, where existing methods were already highly competitive.
These improvements were accompanied by consistently accurate boundary alignment: Easy Peeler achieved the lowest ASSD on all eight datasets reporting surface-distance metrics and the lowest HD95 on seven of them (Supplementary Note~\ref{sec:supp:benchmark}). Measures of deformation regularity further showed that improved overlap generally did not come at the expense of unrealistic deformations. Easy Peeler reduced the proportion of folding voxels ($|J|_{<0}\%$) and Jacobian variability ($std(J)$) on seven of ten datasets. The principal exception was the challenging BCV–CHAOS CT--MR benchmark, where increased overlap was accompanied by greater local folding, illustrating an accuracy–topology trade-off in this particularly difficult cross-modality setting.

\textit{Easy Peeler} also substantially reduced training compute while maintaining competitive or superior registration accuracy (Fig.~\ref{fig:performance}~\textbf{b}).
Relative to TransMorph and LKU-Net, training compute was lower on all ten benchmarks, with reductions ranging from 37--84\% and 34--87\%, respectively. 
Compared with VoxelMorph, training compute was lower in eight of ten datasets. 
Inference efficiency depended on the selected frozen representation because feature extraction contributed to the overall computational cost, particularly on smaller datasets such as ACDC and OASIS-1. Dataset-specific results, computation costs and trainable parameter counts are provided in Supplementary Note~\ref{sec:supp:benchmark}.

\subsection{Overview of Foundation-Feature Representation}
\begin{figure}[tb!]
    \centering
    \includegraphics[width=0.99\linewidth]{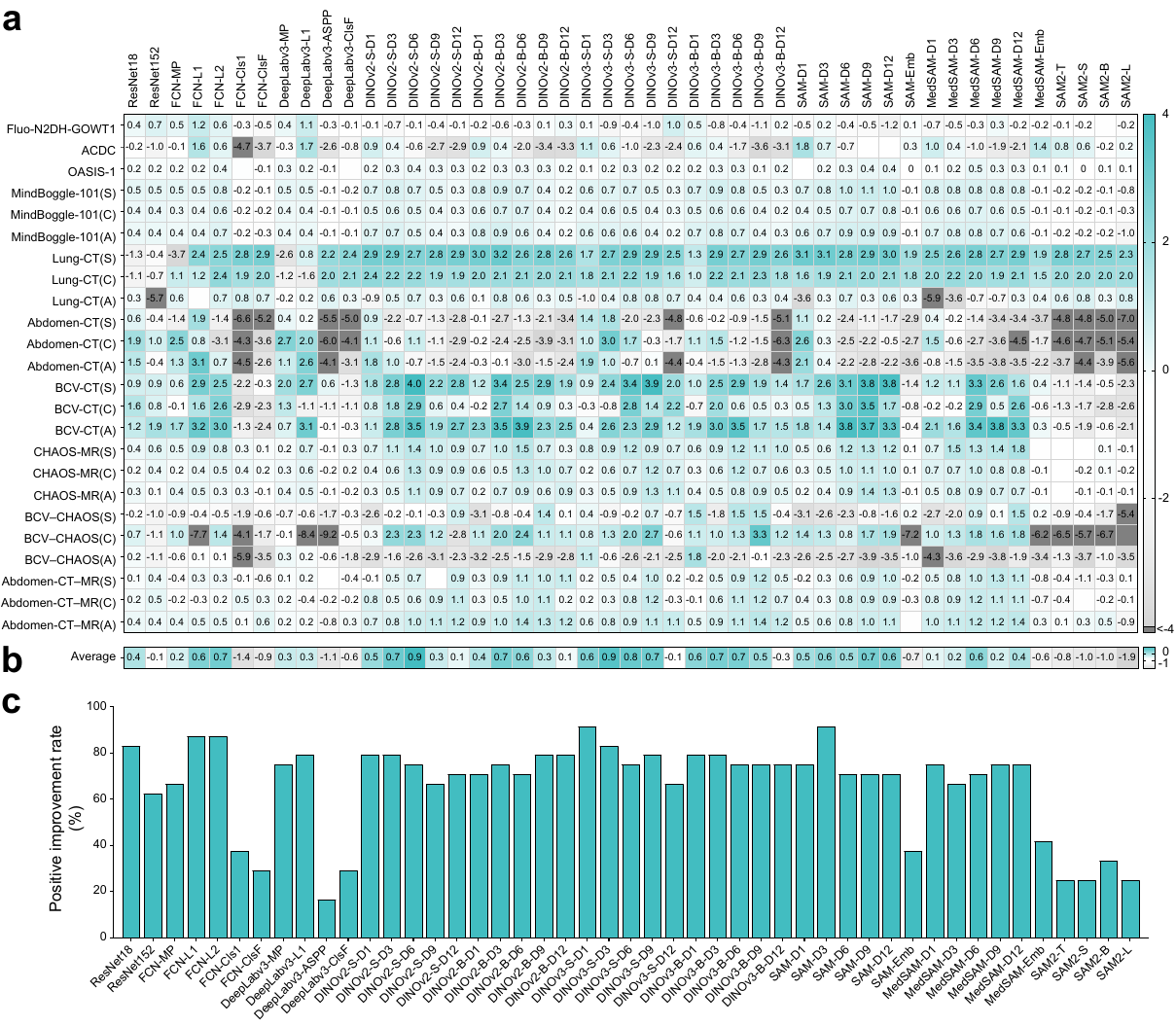}
    \caption{\textbf{Representation choice governs modular registration performance.}
(a) Performance of 47 frozen feature representations extracted from 14 pretrained encoders across 24 registration subtasks (three 2D tasks and seven 3D tasks evaluated in three orthogonal views). Heatmap values show the change in Dice relative to Easy Peeler without pretrained features. No single representation was optimal across all tasks, demonstrating that feature suitability depends on the target anatomy and imaging modality.
(b) Average performance improvement across all subtasks. Most representations (34/47) improved registration accuracy, whereas 13 reduced performance. Intermediate encoder features consistently outperformed classifier- and embedding-proximal representations, indicating that dense correspondence relies on different representations than high-level semantic recognition.
(c) Consistency of individual feature representations, quantified as the proportion of subtasks with $\Delta$Dice $>0$. Shallow and intermediate representations, including DINOv3-S-D1, SAM-D3 and FCN-L1/L2, improved performance across most tasks, whereas deeper semantic and embedding features showed substantially lower robustness.}
\label{fig:overview_47}
\end{figure}

We next investigated how the choice of frozen representation influences registration performance. Across 47 feature representations extracted from 14 pretrained encoders, 34 improved mean Dice relative to the baseline \textit{Easy Peeler} without foundation features, whereas 13 reduced performance (Fig.~\ref{fig:overview_47}). Rather than identifying a universally optimal encoder, these experiments revealed that feature suitability is strongly task dependent.

The best-performing representations varied across anatomies and imaging modalities (Fig.~\ref{fig:overview_47}a). No single encoder consistently dominated all benchmarks, with different CNN- and transformer-based models providing the highest accuracy on different tasks. When performance was averaged across all subtasks, several intermediate representations, including DINOv2-S-D6~\cite{oquab2023dinov2}, DINOv3-S-D3/D6 \cite{simeoni2025dinov3}, FCN-L2~\cite{long2015fully}, and SAM-D9~\cite{kirillov2023segment}, consistently improved registration performance, whereas deeper semantic or embedding-proximal features, such as the SAM2 variants, FCN-Cls1 and DeepLabv3-ASPP, generally reduced performance (Fig.~\ref{fig:overview_47}b).

We further quantified robustness using the proportion of subtasks for which each representation improved Dice. DINOv3-S-D1 and SAM-D3 produced the highest positive-impact rates (91.7\%), followed by FCN-L1 and FCN-L2 (87.5\%) (Fig.~\ref{fig:overview_47}c). Together, these findings demonstrate that representation selection is a critical component of modular registration. The most suitable features cannot be predicted solely from the encoder architecture or pretraining strategy, but instead depend on the target anatomy, imaging modality and registration task. Detailed feature-level analyses are provided in Supplementary Note~\ref{sec:supp:model_choice}.

\subsection{Does orientation matter in 2.5D extraction?}
\begin{figure}
    \centering
    \includegraphics[width=0.99\linewidth]{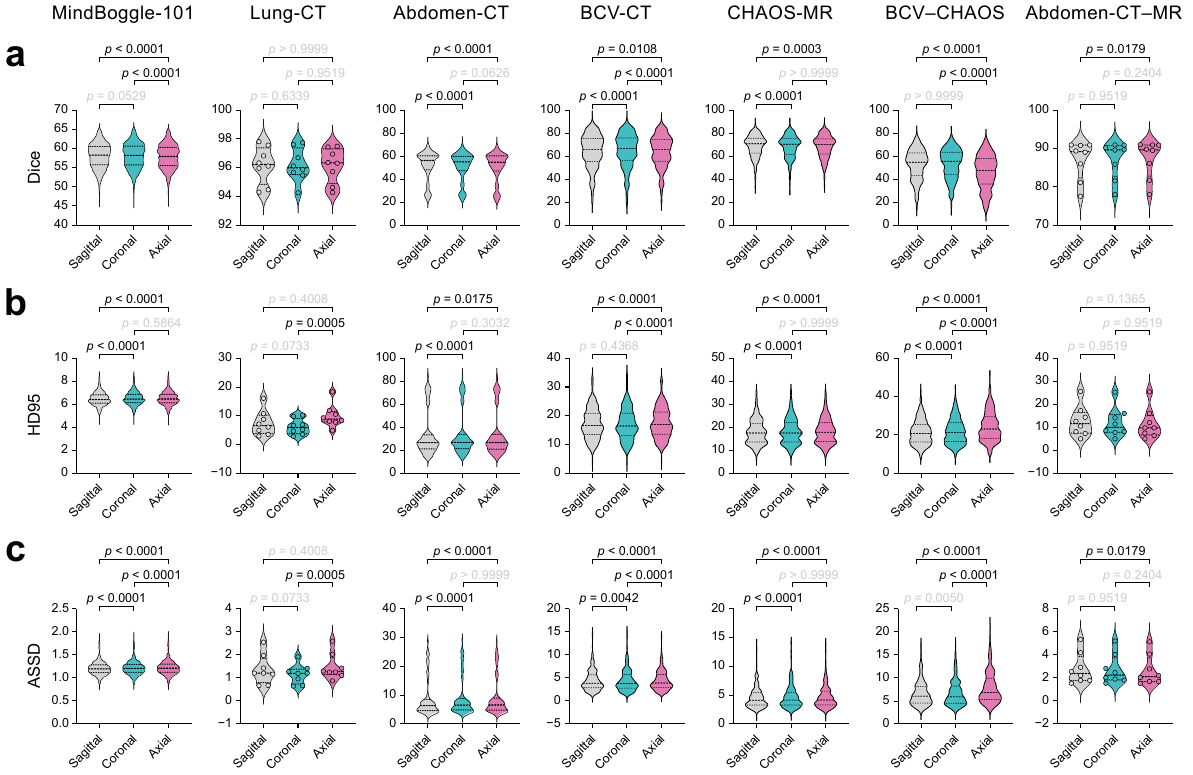}
    \caption{\textbf{Tri-planar 2.5D feature extraction is robust to orientation in volumetric registration.} Registration performance was evaluated using sagittal, coronal and axial feature extraction across seven 3D benchmarks. Panels \textbf{(a--c)} show Dice, HD95, and ASSD, respectively, after full-volume warping. Across most datasets, registration accuracy and boundary alignment were largely unaffected by extraction orientation, despite small statistically significant differences in some comparisons. The principal exception was the challenging BCV--CHAOS CT--MR benchmark, where axial extraction consistently underperformed sagittal and coronal views. These results establish tri-planar 2.5D feature extraction as a robust default for volumetric registration while highlighting that difficult cross-modality tasks may benefit from orientation-specific optimisation. Sample sizes were: MindBoggle-101 ($n=870$), Lung-CT ($n=8$), Abdomen-CT ($n=90$), BCV-CT ($n=306$), CHAOS-MR ($n=210$), BCV--CHAOS ($n=270$) and Abdomen-CT--MR ($n=8$).}
    \label{fig:feature_view_directions}
\end{figure}

Because the foundation models evaluated in this study were originally developed for 2D images, we investigated whether the orientation used for 2.5D feature extraction influences volumetric registration performance. 
Across seven 3D benchmarks, we compared sagittal, coronal and axial extraction using the best-performing feature representation for each orientation (Fig.~\ref{fig:feature_view_directions}). 

Overall, registration performance was remarkably insensitive to extraction orientation.
Although statistical differences (Friedman tests followed by Dunn's multiple-comparisons tests) were detected for several datasets, absolute changes in Dice were small, with the best and worst orientations differing by only 0.03–1.15 percentage points for six of the seven benchmarks. HD95 and ASSD showed the same overall trend, indicating that orientation had only a limited effect on both overlap and boundary alignment.

The principal exception was the challenging inter-subject multimodal BCV–CHAOS CT--MR benchmark, where axial extraction reduced Dice by approximately 7 percentage points and produced higher HD95 and ASSD than sagittal or coronal extraction. These results establish tri-planar 2.5D feature extraction as a robust default for volumetric registration, with orientation-specific optimisation required only for the most challenging cross-modality settings.
Dataset-specific analyses are provided in Supplementary Note~\ref{sec:supp:orientation}.

\subsection{Registration representations do not follow recognition scaling trends}

Modern VFMs are typically evaluated on image recognition, segmentation, and detection \cite{oquab2023dinov2,simeoni2025dinov3,kirillov2023segment,ravi2024sam,ravi2024sam,zhao2025foundation}, where performance has consistently benefited from the shift from CNNs to vision transformers, increased model scale, domain-specific pretraining and successive model generations. Whether these advances translate to dense image correspondence, which requires locally consistent geometric representations, remains unclear. We therefore systematically assessed how feature family, pretraining strategy, model scale and model generation affect registration performance.

\paragraph{Vision transformers often outperform CNNs, but the advantage is task dependent}
We first compared the strongest CNN- and ViT-derived representations for each registration benchmark (Fig.~\ref{fig:feature_cnnvit}~\textbf{a}). ViT features achieved the highest Dice on eight of ten datasets, particularly for brain MRI, abdominal MRI and cross-modality CT--MR registration. However, CNN features remained competitive on fluorescence microscopy and abdominal CT, where they also produced smoother deformation fields. Surface-distance metrics broadly supported these observations. Together, these results suggest that transformer representations provide a strong default for registration, although convolutional features remain advantageous for selected imaging modalities and anatomies.

\paragraph{Medical-domain pretraining does not consistently improve registration.}

We next compared SAM and MedSAM to determine whether medical-domain pretraining improves frozen registration features (Fig.~\ref{fig:feature_cnnvit}~\textbf{b}). MedSAM achieved higher Dice scores on five datasets, whereas SAM performed better on the other five, although the difference was not significant for two of these datasets. Improvements in Dice, surface distance and deformation regularity were similarly inconsistent across datasets. These findings indicate that medical-domain pretraining alone is not a reliable predictor of registration performance; instead, feature utility depends on the anatomy, modality and feature representation.

\paragraph{Larger encoders do not consistently improve registration.}
Increasing model size from DINO Small to Base for ViT or from ResNet18 to ResNet152 for CNN did not produce systematic improvements (Fig.~\ref{fig:feature_cnnvit}~\textbf{c}). Larger models occasionally generated smoother deformation fields, but these gains were not consistently accompanied by improved anatomical overlap or boundary alignment. Registration performance therefore appears to depend more strongly on feature quality than encoder scale.

\paragraph{Newer foundation models do not uniformly improve registration.}
Comparing DINOv3 with DINOv2 revealed no universal advantage for the newer model family (Fig.~\ref{fig:feature_cnnvit}~\textbf{d}). Each model performed best on different datasets, and improvements in Dice were frequently uncoupled from improvements in deformation quality. Together with the preceding analyses, these findings indicate that neither model generation, encoder size nor pretraining strategy reliably predicts registration performance. Instead, frozen representations should be selected empirically using a compact search-and-validation strategy.

\noindent
\begin{minipage}{\textwidth}
    \centering
    \includegraphics[width=0.99\linewidth]{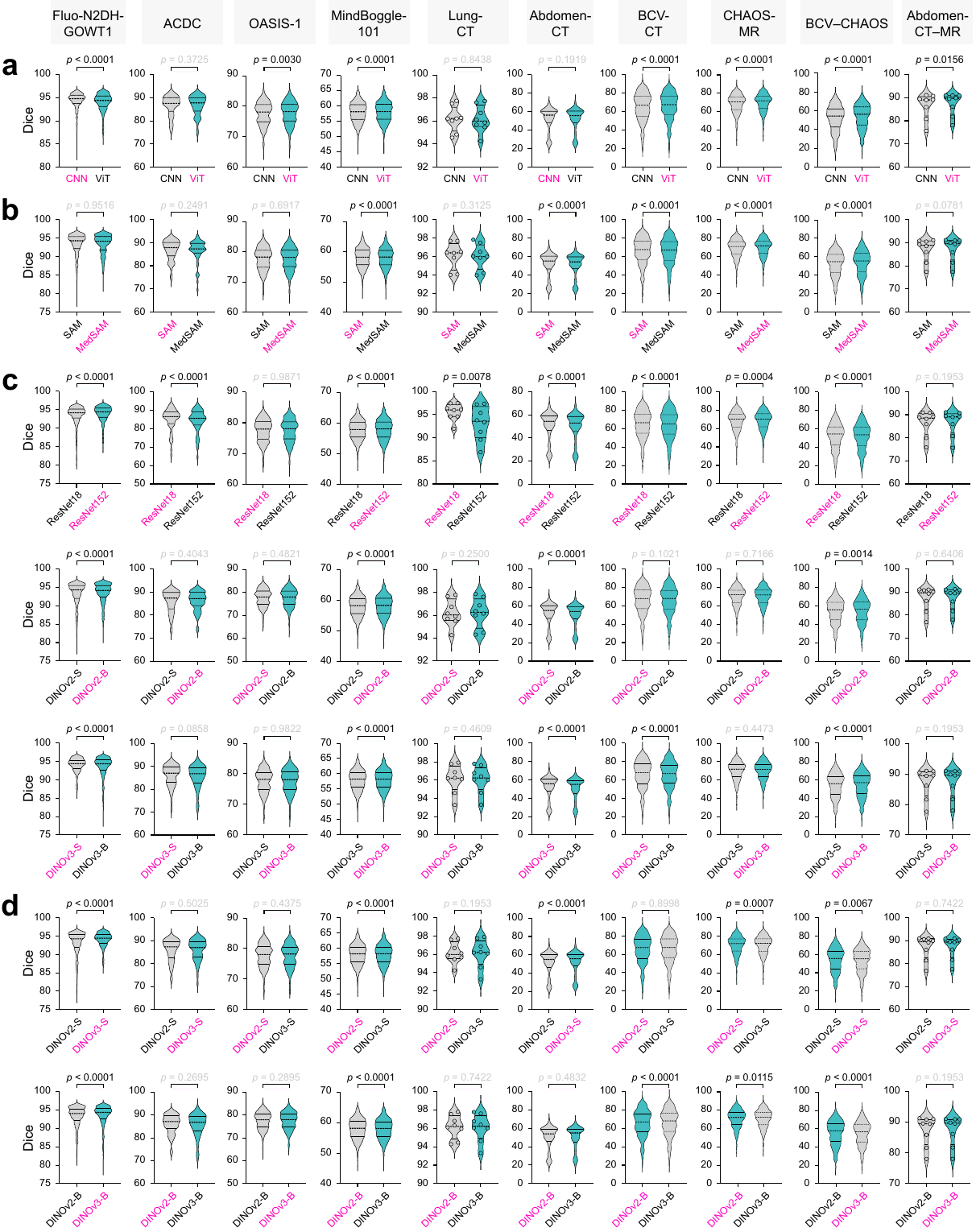}
\end{minipage}
\captionof{figure}{\textbf{Registration performance cannot be predicted solely from feature family, pretraining strategy or model scale.} \textbf{(a)} Comparison of the best-performing convolutional neural network (CNN) and vision transformer (ViT) representations for each registration task. ViT features achieved the highest accuracy in most benchmarks, although CNNs remained competitive for selected microscopy and abdominal CT tasks. 
    \textbf{(b)} Comparison of the best-performing features extracted from SAM and MedSAM. Medical-domain pretraining did not consistently improve registration performance, with each model outperforming the other on different tasks.
    \textbf{(c)} Effect of model scale. Increasing encoder capacity (ResNet18 versus ResNet152; DINOv2-Small versus DINOv2-Base; DINOv3-Small versus DINOv3-Base) did not produce consistent improvements in registration accuracy, indicating that larger encoders are not inherently better for dense correspondence.
    \textbf{(d)} Effect of model generation. DINOv3 did not consistently outperform DINOv2 across the evaluated benchmarks, demonstrating that newer foundation models do not necessarily provide superior registration features.
    For each comparison, the best-performing feature representation within each model family was selected using the validation set and evaluated on the test set. Violin plots show Dice distributions across test pairs; significance was assessed using paired Wilcoxon signed-rank tests. Sample sizes were: Fluo-N2DH-GOWT1 ($n=930$), ACDC ($n=50$), OASIS-1 ($n=400$), MindBoggle-101 ($n=870$), Lung-CT ($n=8$), Abdomen-CT ($n=90$), BCV-CT ($n=306$), CHAOS-MR ($n=210$), BCV--CHAOS ($n=270$) and Abdomen-CT--MR ($n=8$).}
    \label{fig:feature_cnnvit}

\paragraph{Intermediate representations influence registration performance}

We next investigated whether the depth at which frozen representations are extracted influences registration performance. Across encoder families, intermediate feature depths consistently produced the largest improvements, whereas the deepest classifier- or embedding-proximal features were rarely optimal and often reduced accuracy. This suggests that representations beneficial for semantic recognition are not necessarily optimal for dense correspondence, which instead relies on preserving local geometric information.

The dataset-averaged analyses in Fig.~\ref{fig:overview_47}\textbf{a,b} indicated that the largest gains generally arose from intermediate layers. To determine whether this trend also held at the image level, we analysed every test image across all datasets. For each image, we computed the Dice improvement relative to the Easy Peeler baseline,
$$\Delta \text{Dice} = \text{Dice}_{\text{EP w/ F}}-\text{Dice}_{\text{EP w/o F}}$$
and compared this quantity across feature depths for eight encoder families (FCN, DeepLabv3, DINOv2-Small/Base, DINOv3-Small/Base, SAM and MedSAM; Supplementary Fig.~\ref{supp:fig:depth}).

The image-level analysis confirmed the same pattern. For CNN-based encoders, intermediate layers (e.g. FCN-L2) consistently outperformed classifier features, whereas for transformer-based encoders the optimal depth typically occurred before the deepest semantic layers (DINOv2-Small/DINOv3-Small at D6, DINOv2-Base at D3 and DINOv3-Base at D1). Similarly, intermediate representations from SAM and MedSAM consistently exceeded the performance of the final embedding features.

Together, these findings indicate that dense medical correspondence benefits from intermediate representations that preserve local geometry rather than the deepest semantic abstractions. Consequently, feature depth should be considered a primary design choice when adapting frozen foundation models for image registration. Dataset-specific analyses are provided in Supplementary Note~\ref{sec:supp:fature_families}.

\subsection{What makes a good feature for image registration?}

Finally, we investigated whether mutual information (MI) between the input image and frozen feature representations could predict registration performance. Across 24 task--view combinations and 47 feature maps, the relationship between MI and Dice was generally weak and highly task dependent (Fig.~\ref{fig:mi_dice_correlation}~\textbf{f}). The strongest associations were observed for Abdomen-CT, where MI explained a moderate proportion of the variation in Dice across all three orientations. In contrast, most other datasets showed only weak correlations when all candidate features were considered.

To determine whether MI is more informative for selecting promising candidates, we repeated the analysis using only the five highest- and five lowest-MI feature representations for each task--view combination. Under this restricted setting, the relationship became substantially stronger, with 21 of 24 task--view combinations exhibiting an absolute correlation greater than 0.5 (Fig.~\ref{fig:mi_dice_correlation}). However, important exceptions remained. For example, Lung-CT showed an inverse relationship, in which lower-MI features consistently produced higher Dice scores, demonstrating that stronger image--feature similarity does not necessarily translate into better dense correspondence.

Together, these results suggest that MI provides a useful first-pass criterion for narrowing the feature search space, but is insufficient as a standalone predictor of registration quality. The optimal frozen representation remains task dependent and should ultimately be selected using validation data.

Additional analyses are provided in Supplementary Note~\ref{sec:supp:correlation}, further supporting the conclusion that mutual information can narrow the candidate feature space but cannot replace task-specific validation.

\subsection{How to use \textit{Easy Peeler} in practice?}
\textit{Easy Peeler} should be viewed as a lightweight modular framework for reusing frozen feature representations in registration rather than as a fixed architecture with a universal best backbone. In practice, we recommend starting with the baseline version without foundation features, then evaluating a compact set of candidate feature maps on a validation split and retaining the representation that improves overlap accuracy while preserving deformation regularity. Across our benchmark, intermediate feature depths were generally more reliable than deeper classifier- or embedding-proximal layers, although the optimal choice remained anatomy- and modality-dependent. For 3D applications, tri-planar 2.5D extraction provides a robust default, because orientation had little effect in most tasks; however, difficult cross-modality cases may still benefit from orientation-specific tuning. We adopt a single-scale formulation in the present study, while the proposed framework is compatible with cascaded or pyramidal architectures that enable coarse-to-fine registration, as discussed further in the Supplementary Note~\ref{sec:supp:pyramid_extension}.
 Additional practical guidance for selecting candidate feature maps is provided in Supplementary Note~\ref{sec:supp:pratical_advice}.

\noindent
\begin{minipage}{\textwidth}
    \centering
    \includegraphics[width=0.99\linewidth]{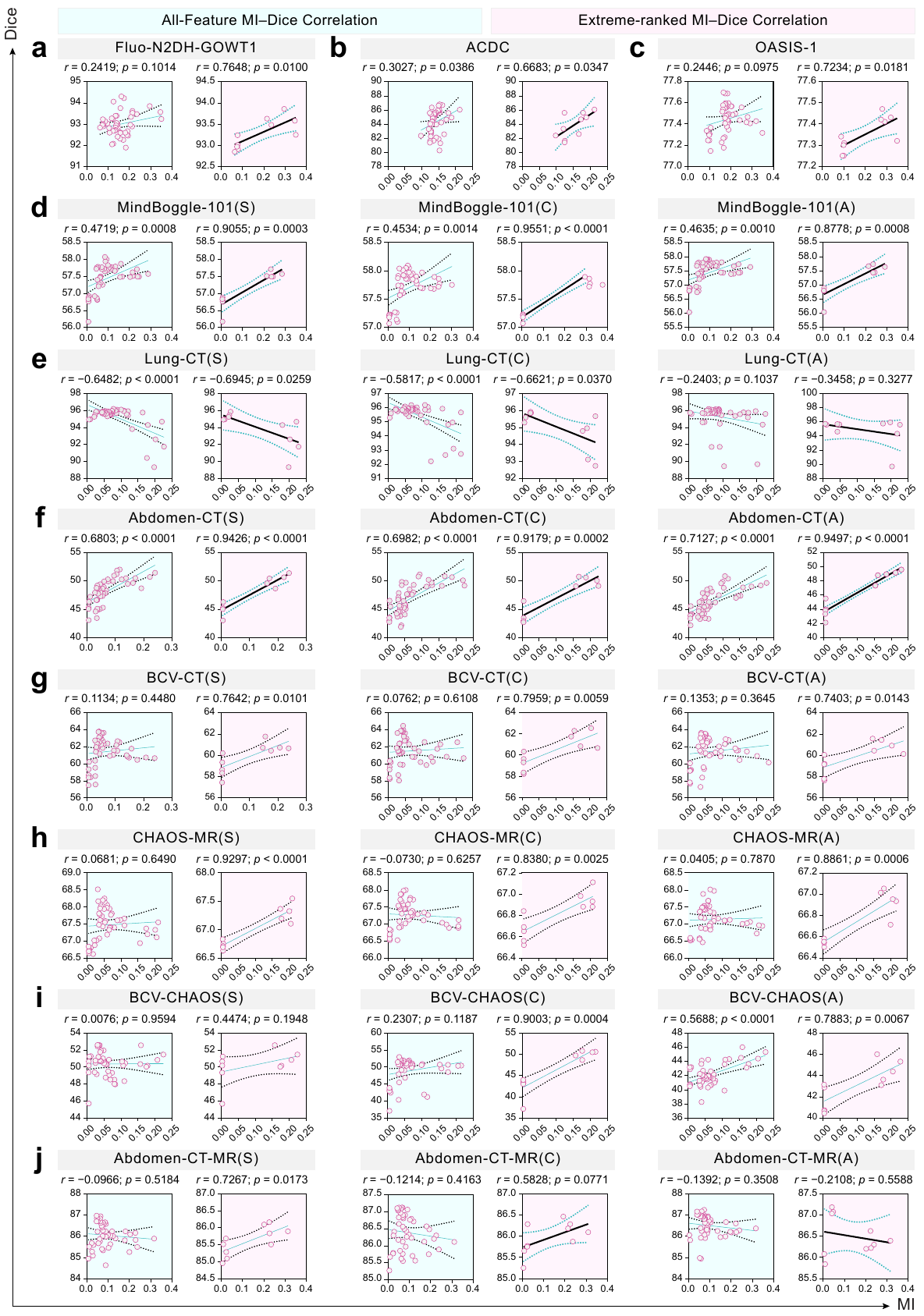}
\end{minipage}
\captionof{figure}{\textbf{Mutual information provides a coarse but imperfect indicator of registration performance.}
\textbf{a--x}, Sub-task-level associations between image--feature mutual information (MI) and registration accuracy. Each panel corresponds to one dataset and task-view combination. The left column shows the association across all 47 feature representations: for each model, MI between the input image and the corresponding extracted feature map was averaged across test pairs and compared with the mean Dice score. Points denote individual feature representations and green curves show linear regression fits. The right column shows the same analysis restricted to the five highest-MI and five lowest-MI features within each task-view combination. Pearson correlation coefficients (\textit{R}) and associated \textit{P} values are shown at the top of each panel.}
\label{fig:mi_dice_correlation}

\section{Discussion}

Most deep-learning approaches to medical image registration jointly learn image representations and deformation estimation within a task-specific network. Our results show that this coupling is not essential for effective registration. Foundation-model representations can support dense correspondence, allowing task-specific learning to focus on deformation estimation. Across diverse registration tasks, this modular formulation achieved competitive or superior accuracy while substantially reducing training costs, showing that representation learning and deformation estimation can be treated as separate design problems.

One of the clearest findings is that deeper features are not necessarily better for registration. Across 41 feature maps from eight encoder families, shallow to intermediate features generally provided the most consistent gains, whereas the deepest features were often less effective and, in some cases, degraded performance. This pattern was observed across both CNN-based encoders, including FCN \cite{long2015fully} and DeepLabv3 \cite{chen2017rethinking}, and ViT-based encoders, including SAM \cite{kirillov2023segment}, MedSAM \cite{ma2024segment}, DINOv2 \cite{oquab2023dinov2} and DINOv3 \cite{simeoni2025dinov3}. These findings suggest that dense correspondence places different demands on visual representations from recognition and semantic segmentation. Whereas recognition typically benefits from increasingly abstract representations \cite{zeiler2014visualizing}, registration appears to favour intermediate features that retain local geometric and boundary information.

A second important observation is that registration does not consistently follow the scaling trends commonly observed in visual recognition. Larger encoders, newer model generations and medical-domain pretraining did not systematically improve registration performance. Comparisons between DINOv2 and DINOv3, ResNet18 and ResNet152, and SAM and MedSAM (Fig.~\ref{fig:feature_cnnvit}) all showed that the most effective representation depended on the registration task. In some datasets, larger models improved Dice scores; in others, they provided little or no benefit, and in several cases, simpler or shallower alternatives performed better. Feature selection for image registration should therefore be treated as an empirical design problem rather than inferred from model scale, generation or pretraining domain alone.

The plane orientation analysis further supports the robustness of this modular approach. Because the foundation models evaluated here were developed for 2D images, we examined sagittal, coronal, and axial feature extraction for volumetric registration. Across most datasets, extraction orientation had little effect on final performance, suggesting that \textit{Easy Peeler} is relatively robust to the choice of plane. An exception occurred for cross-modality BCV--CHAOS CT--MR registration, in which one orientation underperformed the others. Together, these results support tri-planar 2.5D feature extraction as a robust default for volumetric registration, while indicating that challenging cross-modality settings may benefit from orientation-specific validation.

We also investigated whether mutual information (MI) could provide an exploratory indicator of feature suitability for image registration. MI was positively associated with Dice scores in some tasks, but the relationship was weak or inconsistent across many feature representations. When candidate features were restricted to subsets with the highest and lowest MI, the association strengthened in several settings, suggesting that MI may help narrow the feature search space. However, important exceptions remained, including cases where lower-MI features achieved better registration performance. MI may therefore provide a coarse screening criterion but is insufficient as a standalone predictor of registration quality. In practice, feature selection remains task dependent and should be guided by held-out validation data.

Our benchmark has several limitations. First, \textit{Easy Peeler} was deliberately designed as a single-scale registration framework to isolate the value of a modular encoder while learning only a lightweight transition--decoder. Performance could potentially be improved by extending the framework to cascaded \cite{Zhao_2019_ICCV} or pyramidal \cite{mok2020large} formulations, both of which are well established in image registration. Possible pyramid extensions are discussed in Supplementary Note~\ref{sec:supp:pyramid_extension}.

Second, our main benchmark compares against VoxelMorph~\cite{balakrishnan2019voxelmorph}, TransMorph~\cite{chen2022transmorph}, and LKU-Net~\cite{jia2022u}, which are widely used, publicly available, and support both 2D and 3D registration under a unified protocol. These established methods provide representative benchmarks for evaluating Easy Peeler across diverse registration tasks under a unified experimental protocol. Future comparisons with emerging task-specific architectures and classical diffeomorphic methods could further elucidate the settings in which pretrained visual representations offer the greatest benefits. A comparison with ANTs SyN~\cite{avants2008symmetric} is provided in Supplementary Note~\ref{sec:supp:pyramid_extension}.

Third, we deliberately retained fixed pretrained representations rather than adopting end-to-end fine-tuning, enabling a modular framework in which representation learning and deformation estimation remain decoupled. Although this design enabled systematic evaluation of a large representation space at relatively low training cost, it does not establish whether fine-tuning selected encoders could provide additional gains. Moreover, the current framework still requires empirical feature selection. Developing principled criteria that predict feature suitability without exhaustive registration experiments represents an important direction for future work.

Despite these limitations, the central message is clear. Frozen foundation features can be effectively reused for modular medical image registration, but their utility depends critically on the representation selected rather than simply on model scale, generation or domain specificity. 
\textit{Easy Peeler} separates representation choice from deformation learning, enabling new encoders and feature representations to be evaluated without redesigning the registration pipeline. As foundation models continue to evolve \cite{cao2026all}, this modular design provides a practical route for incorporating new representations into registration systems. More broadly, these findings motivate a shift in emphasis from developing increasingly complex task-specific architectures towards understanding which pretrained representations encode the information required for dense correspondence, how these representations should be selected and how they can be efficiently adapted to individual registration tasks.

\section{Methods and Experimental Details}
\label{sec:method}
\subsection{Methods}

The overall pipeline of the proposed \textit{Easy Peeler} is illustrated in Fig.~\ref{fig:visual_results_and_easy_peeler} (d). Given an input pair of fixed and moving images, $I_M$ and $I_F$, we first input them to the frozen foundation model (denoted as $\mathcal{H}$) to extract two separate sets of features, the moving features $\mathcal{H}(I_M)$ and the fixed features $\mathcal{H}(I_F)$.

We note that the features extracted from foundation models may contain many channels. We therefore perform feature filtering and feature prompting on the whole spectrum of features and learn the registration-oriented intensity and structural features. We then fuse the resulting features with a fusion block to form a compact feature representation for the input image pair. Lastly, we learn the representation of displacement from the fused feature with a small representation learning module, followed by a final frozen parameter-free decoding layer to decode the desired deformation field.

\subsubsection{Frozen Encoding Path}

Given a \(d\)-dimensional input image \(I\), with \(d\in\{2,3\}\), defined on a spatial domain \(\Omega\subset\mathbb{R}^{d}\), the frozen encoder \(\mathcal{H}\) extracts a \(c\)-channel feature representation,
$$
\mathcal{H}(I)\in\mathbb{R}^{c\times N_1\times\cdots\times N_d},
$$
where \(N_1,\ldots,N_d\) denote the spatial dimensions of the feature map. Thus, for a 2D image, \(\mathcal{H}(I)\in\mathbb{R}^{c\times m\times n}\), with the formulation extending directly to 3D by including an additional spatial dimension. In the context of image registration, the same frozen encoder is applied independently to the moving and fixed images, yielding feature representations \(\mathcal{H}(I_M)\) and \(\mathcal{H}(I_F)\), respectively.

To probe how generic visual features transfer to registration, we evaluate 14 publicly available foundation-models, ResNet18, ResNet152 \cite{zheng2024residual}, FCN \cite{long2015fully}, DeepLabv3 \cite{chen2017rethinking}, DINOv2-Small, DINOv2-Base \cite{oquab2023dinov2}, DINOv3-Small, DINOv3-Base \cite{simeoni2025dinov3}, SAM \cite{kirillov2023segment}, MedSAM \cite{ma2024segment}, and four SAM2 variants \cite{ravi2024sam}, yielding a total of \textbf{47} candidate feature maps. Input images $I_M$ and $I_F$  are resized per-backbone so that encoder outputs can be reshaped to a common \(32\times32\) spatial resolution; channel dimensionality varies by architecture. All encoders remain frozen throughout training, without any additional training or fine-tuning; consequently, the encoded features can be extracted once before decoder training and reused throughout the entire training process. For 3D volumes, we adopt a tri-planar 2.5D scheme  \cite{song2025dino}: features are extracted from non-overlapping \(3\)-slice blocks along the sagittal, coronal, and axial planes; if the number of slices is not a multiple of three, the final block uses the very last three slices.

\begin{itemize}
  \item \textbf{ResNet~\cite{he2016deep}.} We used \texttt{ResNet18} and \texttt{ResNet152} with ImageNet pretrained weights from PyTorch \footnote{\url{https://docs.pytorch.org/vision/main/models/resnet.html}}. Images were resized to \(128\times128\); we extracted the post–\textit{max-pool} feature map with resolution of \(64\times32\times32\), leading to \textbf{2} features denoted as ResNet18 and ResNet152.
  \item \textbf{FCN~\cite{long2015fully}.} We used \texttt{fcn-resnet101} weights with public COCO/VOC pretraining \footnote{\url{https://docs.pytorch.org/vision/main/models/fcn.html}}. We extracted \textbf{5} features using FCN, specifically, from \(128\times128\) inputs we took features after the initial \textit{max-pool} (\(64\times32\times32\)) and \textit{layer1} (\(256\times32\times32\)); from \(256\times256\) inputs we took \textit{layer2} (\(512\times32\times32\)), the first classifier (\(512\times32\times32\)), and the final classifier (\(21\times32\times32\)). The \textbf{5} features are denoted as FCN-MP, FCN-L1,  FCN-L2, FCN-Cls1, and FCN-ClsF, respectively.
  \item \textbf{DeepLabv3~\cite{chen2017rethinking}.} We used \texttt{deeplabv3-resnet101} with COCO/VOC pretraining \footnote{\url{https://docs.pytorch.org/vision/main/models/deeplabv3.html}}. We extracted \textbf{4} features using DeepLabv3. From \(128\times128\) inputs we took \textit{max-pool} (\(64\times32\times32\)) and \textit{layer1} (\(256\times32\times32\)); from \(256\times256\) inputs we took \textit{ASPP} (\(256\times32\times32\)) and the \textit{final classifier} (\(256\times32\times32\)). The \textbf{4} features are denoted as DeepLabv3-MP, DeepLabv3-L1,  DeepLabv3-ASPP, and DeepLabv3-ClsF, respectively.
  \item \textbf{DINOv2}~\cite{oquab2023dinov2}. We used two different backbones, i.e., DINOv2-Small and DINOv2-Base \footnote{\url{https://github.com/facebookresearch/dinov2}}. With \(448\times448\) inputs; tokens from transformer blocks \(\{1,3,6,9,12\}\) are reshaped to \(32\times32\). Channel dimensions are 384 (Small) and 768 (Base). The two variants \(\times\) five depths leads to a total of \textbf{10} features, denoted as DINOv2-S-D1,  DINOv2-S-D3,  DINOv2-S-D6, DINOv2-S-D9, DINOv2-S-D12, DINOv2-B-D1,  DINOv2-B-D3,  DINOv2-B-D6, DINOv2-B-D9, and DINOv2-B-D12.
  \item \textbf{DINOv3}~\cite{simeoni2025dinov3}. We adopted the DINOv3-Small and DINOv3-Base backbones \footnote{\url{https://github.com/facebookresearch/dinov3}}. Following the DINOv2 protocol, we extracted blocks \(\{1,3,6,9,12\}\), giving feature maps with resolution of \(384\times32\times32\) (Small) and \(768\times32\times32\) (Base). Two variants \(\times\) five depths results in \textbf{10} features: DINOv3-S-D1,  DINOv3-S-D3,  DINOv3-S-D6, DINOv3-S-D9, DINOv3-S-D12, DINOv3-B-D1,  DINOv3-B-D3,  DINOv3-B-D6, DINOv3-B-D9, and DINOv3-B-D12.
  \item \textbf{SAM}~\cite{kirillov2023segment}. Using the ViT-Base image encoder (MAE-pretrained) with \(512\times512\) inputs  \footnote{\url{https://github.com/facebookresearch/segment-anything}}, we collect features from depths \(\{1,3,6,9,12\}\) (\(768\times32\times32\)) and the final image embedding (\(256\times32\times32\)). The extracted \textbf{6} features are denoted as SAM-D1, SAM-D3, SAM-D6, SAM-D9, SAM-D12, and SAM-Emb.
  \item \textbf{MedSAM}~\cite{ma2024segment} \footnote{\url{https://github.com/bowang-lab/MedSAM}}. MedSAM is a fine-tuned SAM on medical images; we extracted features with the same protocol as SAM, producing five depth features (\(768\times32\times32\)) plus the embedding (\(256\times32\times32\)). The \textbf{6} features are denoted as MedSAM-D1, MedSAM-D3, MedSAM-D6, MedSAM-D9, MedSAM-D12, and MedSAM-Emb.
  \item \textbf{SAM2~\cite{ravi2024sam}.} We use four different MAE-pretrained Hiera image encoder, i.e., Tiny, Small, Base+ and Large variants. Each yield a \(256\times32\times32\) feature map from a \(512\times512\) input, resulting in  \textbf{4} features denoted as SAM2-T, SAM2-S, SAM2-B, and SAM2-L.
  
\end{itemize}

In total, we evaluated 47 feature representations.  Feature names follow the format Model-Variant-Stage, where S, B, T and L denote small, base, tiny and large variants, respectively; D1–D12 denote features extracted from encoder depths 1, 3, 6, 9 and 12; MP denotes max-pooling features; ASPP denotes atrous spatial pyramid pooling features; Cls1 and ClsF denote the first and final classifier features; and Emb denotes the final image embedding. We adopt these frozen encoders for three reasons: (i) official, traceable public weights and documentation; (ii) pretraining on large and diverse data yields contrast-, orientation-, and texture-robust features; and (iii) reusing these representations obviates training millions of parameters, substantially reducing compute for downstream registration.

\paragraph{Baseline \textit{Easy Peeler} (no foundation encoder)}

When no foundation encoder is used, we replace the encoder $\mathcal{H}$ with a fixed resampling operator. For a 2D image $I\in\mathbb{R}^{H\times W}$, we define

$$
\mathcal{H}(I)
=
\operatorname{resize}_{\mathrm{bilinear}}
\!\left(I,\,32\times32\right),
$$
where $\operatorname{resize}_{\mathrm{bilinear}}$ denotes bilinear interpolation to a spatial resolution of $32\times32$ pixels.

For a 3D volume $I\in\mathbb{R}^{H\times W\times D}$, we construct three orientation-specific representations using trilinear interpolation:

$$
\mathcal{H}_{\mathrm{sagittal}}(I)
=
\operatorname{resize}_{\mathrm{trilinear}}
\!\left(
I,\,
\left\lceil H/3\right\rceil\times32\times32
\right),
$$

$$
\mathcal{H}_{\mathrm{coronal}}(I)
=
\operatorname{resize}_{\mathrm{trilinear}}
\!\left(
I,\,
32\times\left\lceil W/3\right\rceil\times32
\right),
$$

and

$$
\mathcal{H}_{\mathrm{axial}}(I)
=
\operatorname{resize}_{\mathrm{trilinear}}
\!\left(
I,\,
32\times32\times\left\lceil D/3\right\rceil
\right).
$$

For each orientation, the corresponding dimension is resampled to approximately one third of its original size, while the two orthogonal dimensions are resampled to $32\times32$. For example, an input volume of size $120\times120\times120$ produces representations of size $40\times32\times32$, $32\times40\times32$, and $32\times32\times40$ for the sagittal, coronal, and axial orientations, respectively.

We refer to this encoder-free configuration as \emph{baseline Easy Peeler}, providing a direct baseline for assessing the contribution of foundation-model representations.

\subsubsection{Learnable transition-decoder}

The learnable transition-decoder converts moving and fixed features into a coarse displacement representation through four modules: feature filtering, feature prompting, feature fusion, and displacement representation learning. The resulting representation is subsequently mapped to the target image resolution using a parameter-free Fourier decoding layer.

Given moving and fixed feature maps
\[
x = \mathcal{H}(I_M), \qquad
y = \mathcal{H}(I_F),
\]
or, for an orientation-specific branch \(o\) in the 3D setting,
\[
x^{(o)} = \mathcal{H}_o(I_M), \qquad
y^{(o)} = \mathcal{H}_o(I_F),
\]
the transition-decoder predicts a coarse displacement representation \(z\). For simplicity, \(x\) and \(y\) below denote either the 2D feature maps or the features from a single 3D orientation branch. Detailed feature dimensions and orientation-specific operations are provided in Supplementary Note~\ref{sec:supp:method}.

\paragraph{Feature filtering.}
To emphasize spatial locations at which the moving and fixed representations differ, we first compute their absolute feature difference,
\[
d = |x-y|.
\]
A lightweight gating module predicts an attention mask
\[
m = \sigma\!\left(\mathcal{G}_{\theta}(d)\right),
\]
where \(\mathcal{G}_{\theta}\) is a depthwise convolution and \(\sigma(\cdot)\) denotes the sigmoid function. The mask is applied independently to the moving and fixed features:
\[
g_M = x \odot m, \qquad
g_F = y \odot m,
\]
where \(\odot\) denotes element-wise multiplication.

\paragraph{Feature prompting.}
The filtered representations are augmented with complementary image-intensity and local-structure information. Resolution-matched moving and fixed images are obtained by adaptive average pooling and are denoted by \(I_M^{(\ell)}\) and \(I_F^{(\ell)}\), respectively, where \(\ell\) indexes decoder scale from coarse to fine. Corresponding structural cues are obtained from the gradient magnitude,
\[
S_M^{(\ell)}
=
\operatorname{Sobel}\!\left(I_M^{(\ell)}\right),
\qquad
S_F^{(\ell)}
=
\operatorname{Sobel}\!\left(I_F^{(\ell)}\right).
\]
For 3D inputs, fixed Sobel kernels are applied along all three spatial axes and the resulting responses are squared, summed, and square-rooted to obtain a single-channel gradient-magnitude image. The analogous two-axis operation is used for 2D inputs.

At the coarsest decoder scale, the intensity and structural cues are concatenated with the corresponding filtered features and processed by separate prompting modules:
\[
\begin{aligned}
g_M^{\mathrm{ins}}
&=
\mathcal{P}^{\mathrm{ins}}_{\phi}
\!\left([g_M, I_M^{(0)}]\right),\\
g_M^{\mathrm{str}}
&=
\mathcal{P}^{\mathrm{str}}_{\eta}
\!\left([g_M, S_M^{(0)}]\right),
\end{aligned}
\]
with analogous operations for the fixed representation,
\[
\begin{aligned}
g_F^{\mathrm{ins}}
&=
\mathcal{P}^{\mathrm{ins}}_{\phi}
\!\left([g_F, I_F^{(0)}]\right),\\
g_F^{\mathrm{str}}
&=
\mathcal{P}^{\mathrm{str}}_{\eta}
\!\left([g_F, S_F^{(0)}]\right).
\end{aligned}
\]
Here, \([\cdot,\cdot]\) denotes channel-wise concatenation. The intensity and structure prompting modules use the same bottleneck architecture but learn separate features. Architectural specifications are provided in Supplementary Note~\ref{sec:supp:method}.

\paragraph{Feature fusion.}
The prompted moving and fixed features are first fused separately for the intensity and structure branches. For the intensity branch,
\[
\begin{aligned}
\tilde{f}_{\mathrm{ins}}
&=
[g_M^{\mathrm{ins}}, g_F^{\mathrm{ins}}],\\
a_{\mathrm{ins}}
&=
\sigma\!\left(
\mathcal{A}_{\mathrm{ins}}(\tilde{f}_{\mathrm{ins}})
\right),\\
f_{\mathrm{ins}}
&=
\tilde{f}_{\mathrm{ins}} \odot a_{\mathrm{ins}},
\end{aligned}
\]
and, analogously, for the structure branch,
\[
\begin{aligned}
\tilde{f}_{\mathrm{str}}
&=
[g_M^{\mathrm{str}}, g_F^{\mathrm{str}}],\\
a_{\mathrm{str}}
&=
\sigma\!\left(
\mathcal{A}_{\mathrm{str}}(\tilde{f}_{\mathrm{str}})
\right),\\
f_{\mathrm{str}}
&=
\tilde{f}_{\mathrm{str}} \odot a_{\mathrm{str}}.
\end{aligned}
\]

The resulting intensity and structural features are subsequently combined:
\[
\begin{aligned}
\tilde{f}
&=
[f_{\mathrm{ins}}, f_{\mathrm{str}}],\\
a
&=
\sigma\!\left(\mathcal{A}_{\mathrm{cue}}(\tilde{f})\right),\\
f
&=
\tilde{f} \odot a.
\end{aligned}
\]

Each attention operator \(\mathcal{A}\) is implemented using a depthwise convolution followed by sigmoid gating.

\paragraph{Displacement representation learning.}
The fused feature \(f\) is decoded progressively using three spatial scales. Each scale contains three bottleneck residual blocks. Between consecutive scales, a transposed convolution increases the spatial resolution, with orientation-specific kernels in the 3D model such that the two in-plane dimensions are upsampled while the orientation axis is preserved.

At each finer scale, the upsampled decoder representation is concatenated with resolution-matched moving and fixed intensity images and their corresponding structural cues. A depthwise gating operation modulates the resulting representation before residual processing. Thus, image appearance and local structural information are reintroduced throughout decoding rather than being provided only at the initial feature scale. Full definitions of the residual blocks, scale resolutions, cue injection, and orientation-specific upsampling operations are provided in Supplementary Note~\ref{sec:supp:method}.

After processing at the finest decoder scale, a \(3\times3\) convolution for 2D registration or a \(3\times3\times3\) convolution for 3D registration projects the decoded representation to
\[
z \in \mathbb{R}^{2\times H'\times W'}
\]
for 2D registration and
\[
z \in \mathbb{R}^{3\times H'\times W'\times D'}
\]
for 3D registration.

\paragraph{Parameter-free decoding.}
The predicted displacement representation \(z\) is defined at the finest decoder resolution rather than at the original image resolution. To obtain a full-resolution deformation field without introducing additional learned parameters, we use resolution-agnostic Fourier upsampling~\cite{jia2023fourier}. Specifically, \(z\) is treated as a band-limited spatial representation, and its Fourier spectrum is zero-padded to the target spatial dimensions before applying the inverse Fourier transform:
\[
\begin{aligned}
\text{2D:}\qquad
u
&=
\mathcal{F}^{-1}
\!\left[
\operatorname{ZeroPad}
\left(
\mathcal{F}(z), H\times W
\right)
\right],\\
\text{3D:}\qquad
u
&=
\mathcal{F}^{-1}
\!\left[
\operatorname{ZeroPad}
\left(
\mathcal{F}(z), H\times W\times D
\right)
\right].
\end{aligned}
\]
This produces a full-resolution displacement field
\[
u \in \mathbb{R}^{2\times H\times W}
\]
in 2D or
\[
u \in \mathbb{R}^{3\times H\times W\times D}
\]
in 3D without introducing additional learned parameters.

\subsection{Experimental Details}
\subsubsection{Datasets and registration tasks}
We evaluated Easy Peeler on ten deformable registration tasks constructed from public datasets including OASIS-1 (T1-weighted brain MRI), ACDC (cine cardiac MRI), Abdomen-CT, BCV-CT, CHAOS-MR (T2-SPIR Abdomen MRI), MindBoggle-101 (T1-weighted brain MRI), Lung-CT,  BCV--CHAOS (CT--MR), Abdomen-CT--MR, and Fluo-N2DH-GOWT1 (microscopy of GFP--GOWT1 mouse embryonic stem cells). The evaluation covers intra- and inter-subject registration, mono- and cross-modality settings, and both 2D and 3D registration. For population-level experiments, data were partitioned into training, validation, and test sets before registration pairs were constructed. The Abdomen-CT--MR task was instead used to evaluate intra-subject cross-modality registration through instance-level optimization initialized with weights learned from the inter-subject BCV--CHAOS task. Dataset characteristics, preprocessing, data splits, pair construction, and evaluation labels are summarized in Supplementary Table~\ref{supp:tab:datasets} and detailed in Supplementary Note~\ref{sec:supp:datasets}.

\subsubsection{Optimizations}

We can formulate \textit{Easy Peeler} as a neural network $\mathcal{EP}$ parameterized by $\boldsymbol{\Theta}$, which then can be optimized by minimizing the objective function ${\mathcal L}({\boldsymbol{\Theta}})$, defined as:
\begin{equation}\label{eq:1}
\arg\!\min_{\boldsymbol{\Theta}}  \frac{1}{N} \sum_{i=1}^N \left[ {\mathcal L}_{S}\left( I_{M_i} \circ  T_i, I_{F_i} \right) + \lambda {\mathcal L}_{R}({\boldsymbol{u}_i}) \right],  
\end{equation}
where the transformation $T_i = {\boldsymbol{u}_i} + {\rm{Id}}$ is composed of the displacement field ${\boldsymbol{u}_i}$ predicted by the network $\mathcal{EP}(\boldsymbol{\Theta}; I_{M_i}, I_{F_i})$ and the identity grid ${\rm{Id}}$. Here, $N$ denotes the number of training pairs, $\circ$ represents the spatial warping operator, ${\mathcal L}_{S}$ is the similarity metric quantifying image alignment, and ${\mathcal L}_{R}$ is the regularization term enforcing smooth deformation fields. The hyperparameter $\lambda$ controls the trade-off between alignment accuracy and deformation regularity.

For most datasets, we employed a direct displacement field formulation. However, for the Lung-CT dataset characterized by partial anatomical coverage and potential topological challenges, we adopted a diffeomorphic registration framework using stationary velocity fields (SVF) with scaling-and-squaring integration~\cite{arsigny2006log,ashburner2007fast}. The diffeomorphic objective function is given by:
\begin{equation}\label{eq:2}
   \arg\!\min_{\boldsymbol{\Theta}}  \frac{1}{N} \sum_{i=1}^N \left[ {\mathcal L}_{S}\left( I_{M_i} \circ {\boldsymbol{\rm{Exp}}}(\boldsymbol{v}_i), I_{F_i} \right) + \lambda {\mathcal L}_{R}(\boldsymbol{v}_i) \right],
\end{equation}
where ${\boldsymbol{v}_i} = \mathcal{EP}(\boldsymbol{\Theta}; I_{M_i}, I_{F_i})$ represents the stationary velocity field, and ${\boldsymbol{\rm{Exp}}}(\boldsymbol{v}_i)$ denotes the exponential map computed using seven scaling-and-squaring steps to encourage diffeomorphic properties.

We systematically evaluated multiple similarity metrics based on their suitability for different registration scenarios:

\begin{itemize}
    \item \textbf{Mean Squared Error (MSE):} Employed for monomodal registration tasks (Fluo-N2DH-GOWT1, ACDC, OASIS-1, MindBoggle-101, Lung-CT, CHAOS-MR) where intensity correspondence is expected.
    \item \textbf{Local Normalized Cross-Correlation (LNCC):} Utilized for the Abdomen-CT and BCV-CT datasets to account for local intensity variations ($9 \times 9 \times 9$ windows) while maintaining structural correspondence.
    \item \textbf{Mutual Information (MI):} Applied to the BCV--CHAOS dataset to handle multimodal registration challenges.
    \item \textbf{Combined MI and MIND:} For the Abdomen-CT--MR cross-modality registration, we employed a weighted combination of MI and MIND ~\cite{heinrich2012mind} to capture both statistical dependencies and local self-similarity patterns.
\end{itemize}

We employed a first-order diffusion regularizer ${\mathcal L}_{R}({\boldsymbol{u}}) = \|\nabla {\boldsymbol{u}}\|_2^2$ throughout all experiments, which penalizes spatial gradients of the displacement field to ensure smooth deformations. This choice provides a balance between deformation flexibility and physical plausibility.

\subsubsection{Dataset-Specific Configurations}
\paragraph{Compared methods.}
We compared \textit{Easy Peeler} with VoxelMorph, TransMorph, and LKU-Net, which were selected as representative and widely adopted baselines for learning-based deformable registration. These methods officially support both 2D and 3D inputs, have publicly accessible implementations, and can be evaluated under a unified training protocol across all datasets considered in this study.
This selection enables standardized comparisons without introducing task-specific architectures or
dimensionality-specific methods. We therefore use these baselines as representative reference methods rather than as an exhaustive survey of recent registration models.
All comparative methods (VM, TM, LKU, and the proposed EP) were implemented in PyTorch and trained with consistent experimental protocols. Each experiment was repeated with three different random seeds. Unless otherwise specified, models were trained for 500 epochs with a batch size of 1 and a fixed learning rate of $1\times10^{-4}$. Model selection was performed on the held-out validation set using the Dice similarity coefficient, and the checkpoint achieving the highest validation Dice was retained for final evaluation. Table~\ref{tab:training_config} summarizes the dataset-specific configurations.

\subsubsection{Evaluation}

For each task, we report \emph{Dice score}, \emph{Hausdorff distance at the 95th percentile (HD95)} and \emph{average symmetric surface distance (ASSD)} on the available anatomical labels. We also report deformation regularity via the \emph{standard deviation of the Jacobian determinant} (std($J$)) and the \emph{percentage of voxels with negative Jacobian} ($|J|_{ < 0}\%$). Higher is better for Dice; lower is better for HD95, ASSD, std($J$), and $|J|_{ < 0}\%$. Unless noted, values are mean$\pm$s.d.\ over test cases (aggregated across three random seeds); statistical procedures are detailed in Results.

\backmatter

\bmhead{Supplementary information}

\begin{itemize}
    \item Supplementary Note~\ref{sec:supp:datasets} provides the detailed preprocessing, data splits, pair construction, and evaluation labels for the ten datasets.
    \item Supplementary Note~\ref{sec:supp:benchmark} provides dataset-specific results supporting the main-text section ``Benchmarking established registration methods.''
    \item Supplementary Note~\ref{sec:supp:model_choice} provides the full feature-level analysis supporting the main-text section ``Overview of Foundation-Feature Representation.''
    \item Supplementary Note~\ref{sec:supp:orientation} provides the dataset-specific Dice, HD95 and ASSD analysis supporting the main-text section on ``Does orientation matter in 2.5D extraction?''
    \item Supplementary Note~\ref{sec:supp:fature_families} provides dataset-specific analyses supporting the main-text comparison of CNN versus ViT features, general versus medical features, model scale, and newer versus older foundation models.
    \item Supplementary Note~\ref{sec:supp:correlation} provides more details on MI--Dice associations.
    \item Supplementary Note~\ref{sec:supp:method} provides more details on the methodological design of Easy Peeler.
    \item Supplementary Note~\ref{sec:supp:pyramid_extension} provides more details on the methodological design of Pyramid Easy Peeler, together with comparison with single- and multi-scale ANTs SyN.
    \item Additional practical guidance for selecting candidate feature maps is provided in Supplementary Note~\ref{sec:supp:pratical_advice}.
    \item Additional training details  are provided in Supplementary Note~\ref{sec:supp:trainingdetails}.
\end{itemize}

\bmhead{Acknowledgments}

\textbf{This work is supported by funding from Novo Nordisk.} We thank the organizers and data providers of the datasets used in this work for releasing the data or granting permission for their use. In particular, we thank the Cell Tracking Challenge organizers for permission to use the Fluo-N2DH-GOWT1 dataset, the OASIS project for making the OASIS-1 dataset publicly accessible, and the Learn2Reg organizers for constructing and maintaining multiple benchmark datasets for the evaluation of medical image registration methods. Further details on dataset sources and availability are provided in ``Data availability''.

\section*{Declarations}

\begin{itemize}
\item Funding: This work was supported by funding from Novo Nordisk.
\item Conflict of interest/Competing interests: Mateusz C Florkow is a full-time employee of Novo Nordisk Research and Development UK Limited. JMMH was a fulltime employee of Novo Nordisk Research Centre Oxford during this project and owns shares in Novo Nordisk A/S.

\item Assistance with manuscript preparation: During the preparation of this manuscript, ChatGPT-6 was used to assist with stylistic and grammatical refinement. All AI-generated content was critically reviewed and edited by the authors, who take full responsibility for its accuracy and integrity.
\item Ethics approval and consent to participate: Not applicable.
\item Consent for publication: Not applicable. This study did not include identifiable individual-level information, images or videos requiring consent for publication.
\item Data availability: 

\begin{itemize}
\item The Fluo-N2DH-GOWT1 dataset \cite{bartova2011recruitment} is provided by the Cell Tracking Challenge \cite{mavska2023cell} and is available at \url{https://data.celltrackingchallenge.net/training-datasets/Fluo-N2DH-GOWT1.zip}.

\item The ACDC dataset \cite{bernard2018deep} is publicly available at \url{https://www.creatis.insa-lyon.fr/Challenge/acdc/databases.html}.

\item The original OASIS-1 dataset \cite{marcus2007open} was provided by OASIS Cross-Sectional. Principal investigators: D. Marcus, R. Buckner, J. Csernansky and J. Morris. Funding support was provided by P50 AG05681, P01 AG03991, P01 AG026276, R01 AG021910, P20 MH071616 and U24 RR021382. In this work, we used the HyperMorph-preprocessed release \cite{hoopes2021hypermorph}, which is available at \url{https://github.com/adalca/medical-datasets/blob/master/neurite-oasis.md}.

\item The MindBoggle-101 dataset is available at \url{https://mindboggle.info/data} under the Creative Commons Attribution 4.0 International License.

\item The Lung-CT, BCV-CT, CHAOS-MR and Abdomen-CT--MR datasets are provided by the Learn2Reg Medical Image Registration Challenge at MICCAI 2021 and are available at \url{https://learn2reg.grand-challenge.org/Learn2Reg2021/}.

\item The Abdomen-CT dataset is provided by the Learn2Reg Medical Image Registration Challenge at MICCAI 2020 and is available at \url{https://learn2reg.grand-challenge.org/Learn2Reg2020/}.

\end{itemize}

\item Materials availability: This study did not generate any new biological materials or physical resources. All medical imaging datasets used in this work are publicly available and can be accessed through the repositories cited in the Methods and Data availability section.
\item Code availability: The source code for the proposed method, including training, inference, pre-processing and evaluation scripts, is available at \url{https://github.com/xi-jia/EasyPeeler}. The repository includes a minimal runnable example and configuration files for all datasets evaluated in this study. 

\item Author contributions: B.W.P., J.M.M.H., T.E.N., J.C.H., and K.R.B. are senior authors who conceived and supervised the overall project. B.W.P. and J.M.M.H. secured funding and provided strategic direction throughout the study. X.J., B.W.P., and F.W. designed the study with contributions from all authors. X.J. developed the methodology and workflow and performed the primary experiments and analyses, with input from F.W., R.M., and B.W.P. F.W., R.M., M.C.F., T.H.D., and K.R.B. contributed to experimental design, analyzed and interpreted the data with contributions from all authors. X.J. wrote the original manuscript draft and B.W.P. then critically revised the manuscript. T.E.N. and B.W.P. jointly supervised the research and critically reviewed the manuscript with M.C.F., T.H.D., J.M.M.H., and K.R.B. All authors discussed the results, contributed to manuscript revision, and approved the final version of the manuscript.

\end{itemize}

\begin{appendices}
\newpage
\section{Datasets and preprocessing}
\label{sec:supp:datasets}
\counterwithin{figure}{section}
\counterwithin{table}{section}
We evaluated ten deformable registration tasks spanning 2D and 3D, intra- and inter-subject, and mono- and cross-modality registration settings. Dataset characteristics, data splits, and evaluation annotations are summarized in Supplementary Table~\ref{supp:tab:datasets}, while dataset-specific preprocessing and pair construction are described below.

\begin{sidewaystable}[hbt!]
\centering
\caption{\textbf{Overview of datasets and registration tasks.}
The ten evaluation tasks span 2D and 3D registration, multiple imaging modalities, and intra-sample, intra-subject, inter-subject, and cross-modality settings.
Numbers under ``Data split'' indicate the number of images, volumes, or paired scans assigned to the training, validation, and test sets, respectively; ``Registration pairs'' gives the corresponding number of moving--fixed pairs used in each split.
The Abdomen-CT--MR task differs from the population-level experiments in that all eight paired scans were used for instance-level optimization initialized from the BCV--CHAOS model.}
\label{supp:tab:datasets}
\small
\setlength{\tabcolsep}{4pt}
\renewcommand{\arraystretch}{1.15}
\begin{tabular}{p{2.4cm} p{1.8cm} p{0.8cm} p{2.2cm} p{2.2cm} p{2.5cm} p{3.0cm}}
\hline
\textbf{Task} &
\textbf{Modality} &
\textbf{Dim.} &
\textbf{Registration setting} &
\textbf{Data split (train/val/test)} &
\textbf{Registration pairs (train/val/test)} &
\textbf{Evaluation annotations} \\
\hline

Fluo-N2DH-GOWT1 &
Fluorescence microscopy &
2D &
Intra-sample &
31/30/31 images &
930/870/930 &
Cell masks \\

ACDC &
Cardiac MRI &
2D &
Intra-subject, ES--ED &
40/10/50 paired slices &
40/10/50 &
LV, RV, myocardium \\

OASIS-1 &
Brain MRI &
2D &
Inter-subject &
201/12/201 images &
40,200/22/400 &
24 anatomical structures \\

MindBoggle-101 &
Brain MRI &
3D &
Inter-subject &
21/10/30 volumes &
420/90/870 &
50 anatomical structures \\

Lung-CT &
Respiratory CT &
3D &
Intra-subject, inhale--exhale &
8/4/8 paired volumes &
8/4/8 &
Lung masks \\

Abdomen-CT &
Abdominal CT &
3D &
Inter-subject &
15/5/10 volumes &
210/20/90 &
9 abdominal structures \\

BCV-CT &
Abdominal CT &
3D &
Inter-subject &
20/10/18 volumes &
380/90/306 &
Liver, spleen, left kidney, right kidney \\

CHAOS-MR &
Abdominal MRI &
3D &
Inter-subject &
16/5/15 volumes &
240/20/210 &
Liver, spleen, left kidney, right kidney \\

BCV--CHAOS &
CT--MR &
3D &
Inter-subject, cross-modality &
20/10/18 CT; 16/5/15 MR &
320/50/270 &
Liver, spleen, left kidney, right kidney \\

Abdomen-CT--MR &
CT--MR &
3D &
Intra-subject, cross-modality &
8 paired volumes &
8 instance-level pairs &
Liver, spleen, left kidney, right kidney \\

\hline
\end{tabular}
\end{sidewaystable}

\paragraph{Fluo-N2DH-GOWT1 (fluorescence microscopy; intra-sample; 2D).}
We used the Fluo-N2DH-GOWT1 dataset \cite{bartova2011recruitment} from the Cell Tracking Challenge, a public benchmark for evaluating cell segmentation and tracking in time-lapse microscopy \cite{mavska2023cell}. The dataset contains 2D+t fluorescence image sequences of GFP--GOWT1 mouse embryonic stem cells. The sequence used here comprised 92 two-dimensional frames, with approximately 27 cells per frame. Images were acquired on a Leica TCS SP5 confocal microscope using a Plan-Apochromat 63$\times$/1.4 NA oil-immersion objective, with a native pixel size of  $0.240 \times 0.240~\mu\mathrm{m}$ and a temporal sampling interval of 5~min. The original $1024 \times 1024$ images were resampled to $256 \times 256$ by bilinear interpolation, preserving the field of view and yielding an effective pixel size of $0.960 \times 0.960~\mu\mathrm{m}$. To enforce temporal separation between splits, images were partitioned sequentially with a minimum timestep gap of two into training, validation, and test sets containing 31, 30, and 31 images, respectively. Directed moving--fixed pairs were generated from temporally separated frames within each split, yielding 930, 870, and 930 ordered pairs for training, validation, and testing, respectively. Registration performance was evaluated using the annotated cell masks.

\paragraph{ACDC (cardiac MRI; intra-subject ES--ED; 2D).}
The ACDC dataset \cite{bernard2018deep} comprises cine MR images from 150 patients acquired during breath-hold, with short-axis stacks covering the left ventricle from base to apex. Manual annotations are provided for the left ventricular cavity (LV), right ventricular cavity (RV), and myocardium (Myo) at end-diastole (ED) and end-systole (ES). We used the 100 training subjects with ES and ED annotations and selected mid-ventricular slices for ES--ED registration. Following \cite{jia2025decoder}, slices were resampled to an in-plane resolution of $1.8 \times 1.8~\mathrm{mm}^2$ and center-cropped to $128 \times 128$ pixels. Intensities were clipped at the 0.01th and 99.99th percentiles and min--max normalized to $[0,1]$. The resulting image pairs were divided into training, validation, and test sets containing 40, 10, and 50 pairs, respectively. Registration performance was evaluated using the LV, RV, and Myo annotations.

\paragraph{OASIS-1 (brain MRI; inter-subject; 2D).}
OASIS-1 \cite{marcus2007open} contains 434 cross-sectional T1-weighted MRI scans from 416 subjects. We used the HyperMorph-preprocessed release \cite{hoopes2021hypermorph}, comprising 414 volumes after skull stripping, bias-field correction, and affine alignment to a common template, with a voxel spacing of $1 \times 1 \times 1~\mathrm{mm}^3$ and matrix size of $160 \times 192 \times 224$. We extracted the mid-coronal slice from each volume and divided the resulting images into training, validation, and test sets containing 201, 12, and 201 images, respectively. Following \cite{jia2022u},\footnote{\url{https://github.com/xi-jia/LKU-Net/tree/main/LKU-Net_2D_OASIS}} we constructed 40,200 training pairs, 22 validation pairs, and 400 test pairs. Registration performance was evaluated using 24 anatomical labels present in the extracted slices.

\paragraph{MindBoggle-101 (brain MRI; inter-subject; 3D).}
MindBoggle-101 \cite{klein2012101} provides 101 T1-weighted brain MRI volumes with manual anatomical labels, pre-aligned to MNI152 space with a voxel spacing of $1 \times 1 \times 1~\mathrm{mm}^3$ and matrix size of $182 \times 218 \times 182$. We used 62 volumes from the NKI-RS-22, NKI-TRT-20, and OASIS-TRT-20 subsets. Intensities were clipped at the 1st and 99th percentiles and min--max normalized to $[0,1]$, and volumes were center-cropped to $160 \times 192 \times 160$. One volume was excluded because of an image-direction error. The remaining 61 volumes were divided into training, validation, and test sets containing 21, 10, and 30 volumes, respectively, yielding 420, 90, and 870 ordered moving--fixed pairs. Registration performance was evaluated using 50 anatomical structures, comprising 25 structures per hemisphere, following \cite{kuang2019faim,liu2023contrastive}.

\paragraph{Lung-CT (respiratory CT; intra-subject; 3D).}
The Lung-CT registration dataset provided by the Learn2Reg challenge \cite{hering2020learn2reg} comprises 20 paired inhale--exhale CT scans with lung masks from Radboud University Medical Center, Nijmegen, the Netherlands. Volumes were affinely aligned and resampled to a common matrix size of $192 \times 192 \times 208$ with a voxel spacing of $1.75 \times 1.25 \times 1.75~\mathrm{mm}^3$. The 20 inhale--exhale pairs were divided into training, validation, and test sets containing 8, 4, and 8 pairs, respectively. This task captures large respiratory deformations and variation in the field of view. Registration performance was evaluated using the lung masks.

\paragraph{Abdomen-CT (abdominal CT; inter-subject; 3D).}
The Abdomen-CT cohort originates from \cite{xu2016evaluation} and was preprocessed for the Learn2Reg challenge \cite{hering2022learn2reg}. We used the 30 training CT volumes, which were affinely pre-aligned, cropped, padded, and resampled to $192 \times 160 \times 256$ at an isotropic voxel spacing of $2 \times 2 \times 2~\mathrm{mm}^3$. Volumes were subsequently center-cropped to $192 \times 160 \times 224$, and intensities were scaled to $[0,1]$. Thirteen anatomical labels are available. Following \cite{zheng2024residual}, evaluation used nine structures: spleen, right kidney, left kidney, oesophagus, liver, aorta, inferior vena cava, portal/splenic vein, and pancreas. The volumes were divided into training, validation, and test sets containing 15, 5, and 10 volumes, respectively, yielding 210, 20, and 90 ordered moving--fixed pairs.

\paragraph{BCV-CT (abdominal CT; inter-subject; 3D).}
BCV-CT \cite{xu2016evaluation} comprises 50 portal-venous-phase abdominal CT volumes with varying fields of view and spatial resolutions. We used the Learn2Reg-preprocessed version, in which volumes were affinely aligned using DEEDS-linear and resampled to a common matrix size of $192 \times 160 \times 192$. Intensities were clipped to $[-1000,500]$ HU and scaled to $[0,1]$. One scan was excluded because of an I/O error and one because of errors in the provided segmentation masks. The remaining 48 volumes were divided into training, validation, and test sets containing 20, 10, and 18 volumes, respectively, yielding 380, 90, and 306 ordered moving--fixed pairs. Registration performance was evaluated using four organ labels: liver, spleen, left kidney, and right kidney.

\paragraph{CHAOS-MR (abdominal MRI; inter-subject; 3D).}
The CHAOS dataset \cite{kavur2019chaos,kavur2021chaos,hering2022learn2reg} comprises 40 T2-SPIR abdominal MRI volumes acquired on a 1.5-T Philips scanner with a $256 \times 256$ acquisition matrix. In-plane voxel spacing ranges from 1.36 to 1.89~mm, slice spacing from 5.5 to 9.0~mm, and volumes contain 26--50 slices (mean, 36). All volumes were preprocessed to a common matrix size of $192 \times 160 \times 192$. Four scans were excluded because of errors in the provided segmentation masks. The remaining 36 volumes were divided into training, validation, and test sets containing 16, 5, and 15 volumes, respectively, yielding 240, 20, and 210 ordered moving--fixed pairs. Registration performance was evaluated using four organ labels: liver, spleen, left kidney, and right kidney.

\paragraph{BCV--CHAOS (CT--MR; inter-subject; 3D).}
For inter-subject cross-modality registration, BCV-CT and CHAOS-MR volumes were paired within the corresponding data splits. The BCV-CT training, validation, and test sets contained 20, 10, and 18 volumes, respectively, whereas the corresponding CHAOS-MR sets contained 16, 5, and 15 volumes. Pairing each CT volume with each MR volume within a split yielded 320, 50, and 270 CT--MR pairs for training, validation, and testing, respectively. Registration performance was evaluated using the four anatomical labels shared between the datasets: liver, spleen, left kidney, and right kidney.

\paragraph{Abdomen-CT--MR (CT--MR; intra-subject; 3D).}
The Abdomen-CT--MR dataset from TCIA \cite{akin_2016_tcga_kirc,linehan_2016_tcga_kirp,erickson_2016_tcga_lihc} contains eight curated CT scans with corresponding T1-weighted MR scans for intra-subject CT--MR registration \cite{hering2022learn2reg}. Image pairs were affinely pre-aligned using DEEDS-linear and resampled to $192 \times 160 \times 192$ at an isotropic voxel spacing of $2 \times 2 \times 2~\mathrm{mm}^3$. Unlike the population-level experiments above, all eight pairs were used for instance-level optimization initialized with weights learned from the inter-subject BCV--CHAOS task. Registration performance was evaluated using four organ labels: liver, spleen, left kidney, and right kidney.

\newpage
\section{Dataset-specific benchmark analysis}
\label{sec:supp:benchmark}
\counterwithin{figure}{section}
\counterwithin{table}{section}
We benchmarked \textit{Easy Peeler} against three widely used learning-based registration frameworks: VoxelMorph \cite{balakrishnan2019voxelmorph}, TransMorph \cite{chen2022transmorph}, and LKU-Net \cite{jia2022u}. These methods were selected because they are established reference models with publicly available implementations and can be applied consistently across both 2D and 3D registration tasks. This enabled a standardized comparison across the ten heterogeneous benchmarks.

All methods were evaluated under the same experimental protocol. Each method was trained with three independent random seeds. Unless otherwise stated, per-case predictions were averaged across seeds, and performance is reported as mean $\pm$ s.d. over test cases. Statistical significance was assessed using paired Wilcoxon signed-rank tests on per-case scores. We report Dice as the primary metric and include HD95, ASSD, NMI, $|J|_{<0}\%$, and std($J$) as complementary measures of alignment and deformation quality. Higher Dice and NMI values, together with lower HD95 and ASSD values, indicate better anatomical overlap and image similarity, whereas lower $|J|_{<0}\%$ and std($J$) indicate fewer local foldings and smoother deformation fields.

Across the ten datasets, the feature-enhanced \textit{Easy Peeler} variant achieved the highest averaged Dice in every task. NMI was generally comparable to the strongest baseline, while the Jacobian-based metrics showed that \textit{Easy Peeler} usually preserved or improved deformation regularity, with BCV--CHAOS representing the main accuracy--topology trade-off. Across the eight datasets with HD95 and ASSD results, EP w/ Feature achieved the lowest ASSD in all eight tasks and the lowest HD95 in seven of eight tasks. Fluo-N2DH-GOWT1 and OASIS-1 were not included in this surface-distance analysis because corresponding segmentation labels were unavailable for the paired images after registration, preventing reliable computation of HD95 and ASSD.
 Computational cost was measured using multiply--addition operations, reported in units of Giga-Multi-Adds (G, $10^9$) or Mega-Multi-Adds (M, $10^6$). We separately report train-time and inference-time costs. For experiments using frozen foundation features, the feature-extractor overhead is included; in the 2.5D setting, this cost is reported both per 3-slice block and per volume pair.

\begin{figure}[htb!]
    \centering
    \includegraphics[width=0.99\linewidth]{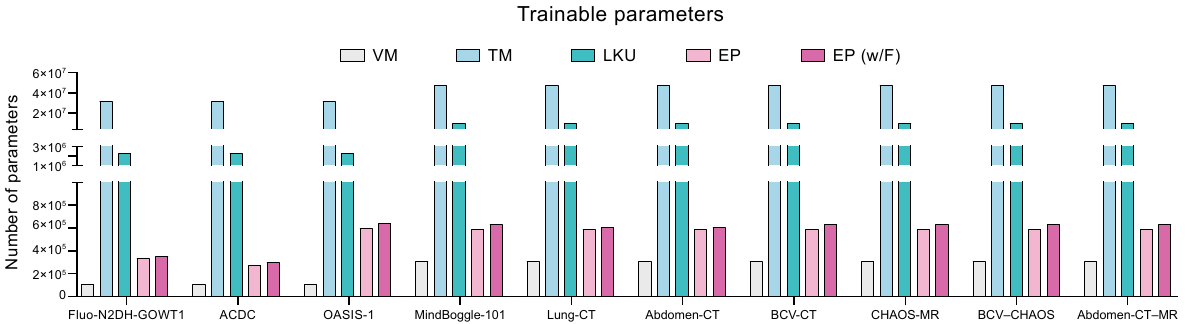}
    \caption{The number of trainable parameters  of compared methods. \textit{Easy Peeler} reduces trainable model complexity relative to TransMorph and LKU-Net.}
    \label{fig:paras_compute}
\end{figure}

Across the ten benchmarks, \textit{Easy Peeler} matched or exceeded the accuracy of VoxelMorph, TransMorph and LKU-Net (Fig.~\ref{fig:performance}; Tables~\ref{tab:supp_results_gfp},~\ref{tab:supp_results_acdc},~\ref{tab:supp_results_oasis},~\ref{tab:supp_results_mind101},~\ref{tab:supp_results_lungct},~\ref{tab:supp_results_Abdomen-CT},~\ref{tab:supp_results_BCV-CT},~\ref{tab:supp_results_CHAOS-MR},~\ref{tab:supp_results_BCV-CTCHAOS-MR},~\ref{tab:supp_results_l2rctmr}). In representative tasks, \textit{Easy Peeler} reduced training compute by approximately 30--87\%, depending on the task and selected feature representation (Table~\ref{tab:supp_costs} and Fig.~\ref{fig:paras_compute}). These results indicate that frozen foundation features can be reused effectively in a compact registration pipeline, providing robust performance across diverse image registration settings without requiring a large task-specific backbone.

\paragraph{Fluo-N2DH-GOWT1 (temporal 2D microscopy, mouse embryonic stem cells).}
Baseline \textit{Easy Peeler} (EP) achieves a Dice score of 93.12$\pm$2.86. Incorporating FCN-L1 (FCN Layer-1) features improves performance to 94.31$\pm$1.98, outperforming VoxelMorph (93.23$\pm$2.36; p$<$0.0001), TransMorph (93.75$\pm$2.54; p$<$0.0001), and LKU-Net (94.02$\pm$2.11; p$<$0.0001). NMI remained similar across methods: EP+FCN-L1 reached 1.21$\pm$0.02, close to VoxelMorph and TransMorph and slightly below LKU-Net (1.22$\pm$0.03). The deformation-quality metrics favoured \textit{Easy Peeler} over the reference baselines: EP+FCN-L1 reduced $|J|_{<0}\%$ to 1.62$\pm$0.95, compared with 3.36$\pm$1.24 for VoxelMorph, 2.14$\pm$0.74 for TransMorph and 2.41$\pm$0.78 for LKU-Net. Its std($J$) was 1.09$\pm$0.70, comparable to TransMorph (1.11$\pm$0.26) and below VoxelMorph and LKU-Net. Notably, EP without frozen features produced the smoothest deformations, with $|J|_{<0}\%=$1.15$\pm$0.52 and std($J$)=0.59$\pm$0.26, but at a lower Dice score.

In terms of computation, the training cost was 1.01 G and 1.03 G Multi-Adds for EP and EP+FCN-L1, respectively, lower than VoxelMorph (1.48 G), TransMorph (6.30 G) and LKU-Net (6.04 G). 
Relative to VoxelMorph, TransMorph, and LKU-Net, EP reduces training and inference costs by 31.76\%, 83.97\%, and 83.28\%, respectively. EP+FCN-L1 reduces training cost by 30.41\%, 83.65\%, and 82.95\%. At inference, FCN-L1 feature extraction adds 256.64 \textit{M} Multi-Adds; consequently, the infer cost of EP+FCN-L1 is 4.28\% higher than VoxelMorph but remains far lower than TransMorph (75.50\%) and LKU-Net (74.45\%).





\paragraph{ACDC (cardiac cine MRI, intra-subject, ES--ED).}
EP achieved a Dice score of 85.02$\pm$5.66. Incorporating SAM-D1 (SAM Depth-1) feature  significantly improved the performance to 86.82$\pm$4.30, outperforming VoxelMorph (85.71$\pm$5.01; p$=$0.0014), TransMorph (85.66$\pm$5.39; p$=$0.0018), and LKU-Net (84.94$\pm$5.24; p$<$0.0001). The Dice improvement was accompanied by better surface distances: EP w/ Feature reduced HD95 to 4.38 mm and ASSD to 1.44 mm, compared with 4.68--4.93 mm HD95 and 1.56--1.62 mm ASSD for the three baselines. NMI for EP+SAM-D1 was 1.29$\pm$0.04, similar to TransMorph and LKU-Net and only slightly below VoxelMorph (1.30$\pm$0.04). EP+SAM-D1 also gave the most regular deformation fields among the compared methods, with the lowest $|J|_{<0}\%$ (0.29$\pm$0.33) and the lowest std($J$) (0.21$\pm$0.05). Thus, the Dice improvement on ACDC was accompanied by a small but consistent reduction in deformation folding and Jacobian variability. In terms of computation, EP required 800.49\,$\mathrm{M}$ Multi-Adds for training, increasing to 858.61\,$\mathrm{M}$ Multi-Adds with the SAM-D1 feature. This training cost is higher than VoxelMorph (369.41\,$\mathrm{M}$) but lower than TransMorph (1.60\,$\mathrm{G}$) and LKU-Net (1.51\,$\mathrm{G}$) with reductions of  46.34\% and 43.14\%, respectively. At inference, SAM-D1 feature extraction adds 630.75 \textit{M} Multi-Adds per image, increasing the total inference cost to above both TransMorph and LKU-Net.





\paragraph{OASIS-1 (T1w brain MRI, inter-subject).}
EP achieves a a Dice score of 77.25$\pm$3.89; adding MedSAM-D6 feature increased Dice to 77.70$\pm$3.79, exceeding VoxelMorph (76.72$\pm$3.87; p$<$0.0001) and performing comparably to TransMorph (77.62$\pm$3.80; p=0.0513) and LKU-Net (77.67$\pm$3.85; p=0.6008). NMI was also comparable to the strongest baselines: EP+MedSAM-D6 achieved 1.33$\pm$0.02, matching TransMorph and LKU-Net at the reported precision. The $|J|_{<0}\%$ was slightly lower for EP+MedSAM-D6 (1.26$\pm$0.46) than for VoxelMorph (1.33$\pm$0.50), TransMorph (1.28$\pm$0.47) and LKU-Net (1.30$\pm$0.48). std($J$) was 0.54$\pm$0.08, similar to the baseline methods, with LKU-Net showing the lowest value at 0.53$\pm$0.08. These results indicate that on OASIS-1, \textit{Easy Peeler} provides a small Dice gain while maintaining essentially comparable image similarity and deformation regularity. For training compute, EP required 1.78 G Multi-Adds and EP+MedSAM-D6 required 1.86 G, both higher than VoxelMorph (692.65 M) but lower than TransMorph (2.97 G) and LKU-Net (2.83 G). At inference, MedSAM-D6 feature extraction adds 722.89 \textit{M} Multi-Adds per image, making the total inference cost of EP+MedSAM-D6 higher than those of TransMorph and LKU-Net.





\paragraph{MindBoggle-101 (T1w brain MRI, inter-subject).}
EP achieved a Dice score of 57.34$\pm$3.33. Incorporating SAM-D12 features improved performance to 58.09$\pm$3.25, higher than VoxelMorph (55.69$\pm$3.20; p$<$0.0001), TransMorph (57.69$\pm$3.17; p$<$0.0001), and LKU-Net (57.97$\pm$3.45; p$<$0.0001). EP w/ SAM-D12 Feature achieved 6.53 mm HD95 and 1.22 mm ASSD, slightly superior to the strongest baseline values. NMI for EP+SAM-D12 was 1.28$\pm$0.01, comparable to TransMorph and slightly below LKU-Net (1.29$\pm$0.02). The deformation-quality metrics were favourable: EP+SAM-D12 achieved the lowest $|J|_{<0}\%$ (0.87$\pm$0.23) and the lowest std($J$) (0.54$\pm$0.06) among all methods. These results suggest that the Dice gain on MindBoggle-101 was accompanied by fewer foldings and smoother local volume changes. Training compute for EP and EP+SAM was 154.48\,$\mathrm{G}$ and 159.10\,$\mathrm{G}$, respectively, lower than VoxelMorph (285.23\,$\mathrm{G}$), TransMorph (469.80\,$\mathrm{G}$), and LKU-Net (761.23\,$\mathrm{G}$), corresponding to training-cost reductions of 44.22–79.71\% across these approaches. At inference, SAM-D12 feature extraction adds 841.01\,$\mathrm{M}$ per 3-slice block (107.65\,$\mathrm{G}$ per volume pair), yielding a total inference cost of 266.75\,$\mathrm{G}$, which is still lower than VoxelMorph (by 6.48\%), TransMorph (by 43.22\%) and LKU-Net (by 64.96\%).



\paragraph{Lung-CT (3D respiratory).}
EP achieved a Dice score of 95.33$\pm$1.92, higher than VoxelMorph (93.82$\pm$1.77), TransMorph (93.31$\pm$3.40), and LKU-Net (95.03$\pm$2.08). Incorporating DINOv2-S-D1 feature (DINOv2-Small Depth-1) increased the mean Dice score to 96.18$\pm$1.08, although this improvement over the baseline EP without foundation features was not statistically significant (p=0.6406). The resulting performance was significantly higher than VoxelMorph (p=0.0078) and TransMorph (p=0.0078), and comparable to LKU-Net (p=0.1484). EP+DINOv2-S-D1 reduced HD95 to 6.25 mm and ASSD to 1.16 mm, compared with the best baseline values of 8.19 mm and 1.47 mm. EP+DINOv2-S-D1 also produced the lowest std($J$), with std($J$)=3.79$\pm$0.53 compared with 4.71$\pm$0.97 for VoxelMorph, 6.68$\pm$1.89 for TransMorph and 5.89$\pm$0.77 for LKU-Net. However, $|J|_{<0}\%$ was not the lowest: EP+DINOv2-S-D1 yielded 0.72$\pm$0.77, higher than VoxelMorph (0.15$\pm$0.07), TransMorph (0.65$\pm$0.98) and LKU-Net (0.21$\pm$0.08). NMI was 1.23$\pm$0.03, similar to VoxelMorph and LKU-Net and slightly below TransMorph (1.24$\pm$0.04). Thus, on Lung-CT, \textit{Easy Peeler} achieved the strongest Dice and surface-distance performance and lower global Jacobian variability, but with a modest increase in local folding relative to the baselines. Train-time Multi-Adds were 168.96\,$\mathrm{G}$ (EP) and 156.79\,$\mathrm{G}$ (EP w/ F), lower than VoxelMorph (444.95\,$\mathrm{G}$), TransMorph (789.19\,$\mathrm{G}$), and LKU-Net (1.19\,$\mathrm{T}$), corresponding to training-cost reductions of 62.03–86.82\% across these approaches. At inference, DINOv2-S-D1 feature extraction adds 233.38\,$\mathrm{M}$ per 3-slice block (29.87\,$\mathrm{G}$ per volume pair), yielding a total inference cost of 186.66\,$\mathrm{G}$, which is lower than VoxelMorph (by 58.05\%), TransMorph (by 76.35\%), and LKU-Net (by 84.31\%).




\paragraph{Abdomen-CT (3D abdominal CT, inter-subject).}
EP achieved a Dice score of 49.99$\pm$12.44, higher than VoxelMorph (46.05$\pm$11.65), TransMorph (48.52$\pm$12.22), and LKU-Net (48.21$\pm$12.06). Adding FCN-L1 features further increased Dice to 51.93$\pm$12.69 (all p$<$0.0001). EP+FCN-L1 achieved the lowest HD95 (33.63$mm$) and ASSD (8.67$mm$). EP+FCN-L1 also achieved the highest NMI among the compared methods at 1.17$\pm$0.03. The Jacobian metrics showed a clear deformation-quality advantage: EP+FCN-L1 reduced $|J|_{<0}\%$ to 3.22$\pm$1.71, compared with 8.58$\pm$2.40 for VoxelMorph, 13.62$\pm$2.06 for TransMorph and 10.00$\pm$2.17 for LKU-Net. std($J$) was also lower for EP+FCN-L1 (0.64$\pm$0.08) than for the three baselines. These results indicate that on Abdomen-CT, the accuracy gain was accompanied by substantially fewer foldings and smoother deformation fields. Training compute was 154.48\,$\mathrm{G}$ and 156.01\,$\mathrm{G}$ for EP and EP+FCN, respectively, lower than VoxelMorph (399.32\,$\mathrm{G}$), TransMorph (657.68\,$\mathrm{G}$), and LKU-Net (1.07\,T), corresponding to training-cost reductions of 60.93–85.56\% across these approaches. At inference, FCN-L1 feature extraction adds 256.64\,$\mathrm{M}$ per 3-slice block (32.85\,$\mathrm{G}$ per volume pair), yielding a total inference cost of 188.86\,$\mathrm{G}$, which still reduces inference compute by 52.70\%, 71.28\%, and 82.35\% versus VoxelMorph, TransMorph, and LKU-Net, respectively.




\paragraph{BCV-CT (3D abdominal CT, inter-subject).}
EP achieved 60.98$\pm$14.81 Dice, surpassing VoxelMorph (56.20$\pm$13.93), TransMorph (60.48$\pm$14.32), and LKU-Net (59.26$\pm$14.09). Adding  SAM-D9 feature increased Dice to 64.47$\pm$14.51 (p$<$0.0001) and achieved the lowest HD95 (17.07$\pm$5.45 voxels) and ASSD (4.30$\pm$2.24 voxels). NMI for EP+SAM-D9 was 1.16$\pm$0.03, similar to but slightly below the 1.17 reported for the strongest baselines. Despite this small NMI difference, EP+SAM-D9 substantially improved deformation quality: $|J|_{<0}\%$ was reduced to 3.44$\pm$1.68, compared with 9.37$\pm$2.78 for VoxelMorph, 10.63$\pm$3.06 for TransMorph and 12.01$\pm$2.64 for LKU-Net. std($J$) was also lowest for EP+SAM-D9 (0.67$\pm$0.08). Thus, on BCV-CT, \textit{Easy Peeler} improved anatomical overlap and surface distances while producing fewer foldings and smoother Jacobian fields. Train-time Multi-Adds are 130.34\,$\mathrm{G}$ and 134.24\,$\mathrm{G}$ for EP and EP+SAM-D9, below VoxelMorph (342.27\,$\mathrm{G}$), TransMorph (563.74\,$\mathrm{G}$), and LKU-Net (913.47\,$\mathrm{G}$), corresponding to training-cost reductions of 60.78–85.73\% across these approaches. At inference, extracting SAM-D9 feature adds 781.95\,$\mathrm{M}$ per 3-slice block (84.45\,$\mathrm{G}$ per volume), yielding a total inference cost of 218.69\,$\mathrm{G}$, reducing cost by 36.11\%, 61.21\%, and 76.06\% relative to the three baselines.




\paragraph{CHAOS-MR (multi-modality CT--MR, inter-subject).}
EP reached 66.73$\pm$9.98 Dice, exceeding VoxelMorph (65.34$\pm$10.03). Adding MedSAM-D12 feature increased Dice to 68.51$\pm$9.36, higher than TransMorph (67.93$\pm$10.04, p$<$ 0.0001) and LKU-Net (68.27$\pm$9.66, p= 0.0204). EP w/ Feature achieved the  lowest ASSD (4.47$\pm$1.83 voxels) and second-best HD95 (17.84$\pm$5.40 voxels). NMI for EP+MedSAM-D12 was 1.16$\pm$0.02, matching the baseline methods at the reported precision. The deformation-quality metrics were comparable but not uniformly better: EP+MedSAM-D12 had $|J|_{<0}\%$=0.49$\pm$0.28, close to VoxelMorph (0.46$\pm$0.27) and LKU-Net (0.42$\pm$0.25), and std($J$)=0.55$\pm$0.15, close to LKU-Net (0.53$\pm$0.14). These results indicate that on CHAOS-MR, \textit{Easy Peeler} provides a small accuracy gain while maintaining deformation regularity similar to the strongest baselines. Train-time Multi-Adds are 154.48\,$\mathrm{G}$ and 159.10\,$\mathrm{G}$ for EP and EP+MedSAM-D12, both below VoxelMorph (342.27\,$\mathrm{G}$), TransMorph (563.74\,$\mathrm{G}$), and LKU-Net (913.47\,$\mathrm{G}$), corresponding to training-cost reductions of 53.52–83.09\% across these approaches. At inference, extracting MedSAM-D12 feature adds 841.01\,$\mathrm{M}$  per 3-slice block (107.65\,$\mathrm{G}$ per volume pair), yielding a total inference cost of 266.75\,$\mathrm{G}$, which is lower by 22.06\%, 52.68\%, and 70.80\% than the three baselines, respectively.



\paragraph{BCV--CHAOS (multi-modality, CT--MR, inter-subject).}
The baseline methods yielded limited Dice improvement over the initial alignment (23.96$\pm$11.39): VoxelMorph (25.81$\pm$13.33), TransMorph (35.15$\pm$15.47), LKU-Net (28.98$\pm$13.36).  \textit{Easy Peeler} substantially improved Dice, reaching 51.02$\pm$13.92 without frozen features and 52.96$\pm$12.90 with DINOv3-B-D9 features. EP w/ Feature reduced HD95 and ASSD to 21.74 and 6.61 voxels, respectively, compared with the strongest baseline values of 30.85 and 11.21 voxels. NMI did not show the same pattern: EP+DINOv3-B-D9 achieved 1.13$\pm$0.01, below VoxelMorph (1.15$\pm$0.02), TransMorph (1.16$\pm$0.01) and LKU-Net (1.16$\pm$0.02). The Jacobian metrics also revealed a trade-off. EP+DINOv3-B-D9 produced $|J|_{<0}\%$=9.06$\pm$1.12 and std($J$)=1.45$\pm$0.07, both higher than the best baseline values. Therefore, BCV--CHAOS should be interpreted as the clearest example of an accuracy--topology trade-off: \textit{Easy Peeler} greatly improves anatomical overlap in a difficult cross-modality setting, but the resulting deformation field contains more local foldings than the most regular baseline.
Train-time cost was 154.48\,$\mathrm{G}$ and 134.24\,$\mathrm{G}$, both below VoxelMorph (342.27\,$\mathrm{G}$), TransMorph (563.74\,$\mathrm{G}$), and LKU-Net (913.47\,$\mathrm{G}$), corresponding to training-cost reductions of 54.87–85.30\% across these approaches. Inference adds 668.56\,$\mathrm{M}$  per 3-slice block (72.20\,$\mathrm{G}$ per volume pair), yielding a total inference cost of 206.44\,$\mathrm{G}$, which is lower than VoxelMorph (by 39.68\%), TransMorph (by 63.38\%), and LKU-Net (by 77.40\%).




\paragraph{Abdomen-CT--MR (multi-modality, CT--MR, intra-subject).}
EP achieved 85.94$\pm$5.18 Dice, exceeding VoxelMorph (81.19$\pm$10.23), TransMorph (79.18$\pm$18.04), and LKU-Net (83.82$\pm$8.80). DINOv3-B-D9 (DINOv3-Base Depth-9) further improved Dice to 87.26$\pm$4.61, with p-values to the compared methods being 0.0078, 0.0078, 0.0078, and 0.0156. EP w/ Feature reduced HD95 to 11.51 $mm$ and ASSD to 2.64 $mm$, outperforming VoxelMorph, TransMorph and LKU-Net. These results indicate that the Dice gains in difficult abdominal and CT--MR settings are accompanied by improved boundary alignment, not only increased volumetric overlap. NMI for EP+DINOv3-B-D9 was 1.24$\pm$0.02, below TransMorph and LKU-Net (both 1.26 at the reported precision), indicating that the Dice gain was not driven by higher global intensity similarity. In contrast, the deformation-quality metrics favored Easy Peeler: EP+DINOv3-B-D9 reduced $|J|_{<0}\%$ to 1.42$\pm$0.90, compared with 2.60$\pm$1.32 for VoxelMorph, 3.46$\pm$2.53 for TransMorph and 2.81$\pm$1.53 for LKU-Net. std($J$) was also lowest for EP+DINOv3-B-D9 at 0.51$\pm$0.08. Thus, in this intra-subject CT--MR task, \textit{Easy Peeler} improved Dice and surface distances while also producing fewer foldings and smoother deformation fields. Train-time Multi-Adds are 130.34\,$\mathrm{G}$ and 159.10\,$\mathrm{G}$ for EP and EP+DINOv3-B-D9, below VoxelMorph (342.27\,$\mathrm{G}$), TransMorph (563.74\,$\mathrm{G}$), and LKU-Net (913.47\,$\mathrm{G}$), corresponding to training-cost reductions of 53.52–85.73\% across these approaches. At inference, extracting DINOv3-Base-Depth-9 feature adds 668.56\,$\mathrm{M}$ per 3-slice block (85.58\,$\mathrm{G}$ per volume pair), yielding a total inference cost of 244.68\,$\mathrm{G}$, representing reductions of 28.51\%, 56.60\%, and 73.21\% relative to VoxelMorph, TransMorph, and LKU-Net, respectively.



\clearpage
\begin{table}[hbt!]
\centering
\setlength{\tabcolsep}{2.0pt}
\caption{Registration performance on Fluo-N2DH-GOWT1.
Comparison of five methods on the Fluo-N2DH-GOWT1 test set. Values are mean$\pm$s.d. over test cases. Metrics: Dice (\%), Normalized Mutual Information (NMI) between warped moving and fixed images, $|J|_{ < 0}\%$, and std($J$).}
\label{tab:supp_results_gfp}
\begin{tabular}{lcccc}
\toprule
Method & Seed 1 & Seed 2 & Seed 3 & Averaged \\
\midrule
\multicolumn{5}{l}{\textbf{Dice (\%)}} \\
\midrule
Initial & - & - & - & 56.64 $\pm$ 20.18\\
VM        & 93.18 $\pm$ 2.49 & 93.22 $\pm$ 2.37 & 93.29 $\pm$ 2.29 & 93.23 $\pm$ 2.36 \\
TM        & 93.34 $\pm$ 3.17 & 93.92 $\pm$ 2.56 & 94.01 $\pm$ 2.24 & 93.75 $\pm$ 2.54 \\
LKU       & 94.12 $\pm$ 2.25 & 93.95 $\pm$ 2.09 & 93.98 $\pm$ 2.10 & 94.02 $\pm$ 2.11 \\
EP(w/o F) & 93.23 $\pm$ 2.84 & 93.09 $\pm$ 2.89 & 93.05 $\pm$ 2.98 & 93.12 $\pm$ 2.86 \\
EP(w/ F) & 94.25 $\pm$ 2.02 & 94.57 $\pm$ 1.74 & 94.11 $\pm$ 2.65 & 94.31 $\pm$ 1.98 \\
\bottomrule
\end{tabular}
\end{table}

\clearpage
\begin{table}[hbt!]
\centering
\setlength{\tabcolsep}{2.0pt}
\caption{Registration performance on ACDC.
Comparison of five methods on the ACDC test set. Values are mean$\pm$s.d. over test cases. Metrics: Dice (\%), HD95 (mm), ASSD (mm), NMI, $|J|_{ < 0}\%$, and std($J$).}
\label{tab:supp_results_acdc}
\begin{tabular}{lcccc}
\toprule
Method & Seed 1 & Seed 2 & Seed 3 & Averaged \\
\midrule
\multicolumn{5}{l}{\textbf{Dice (\%)}} \\
\midrule
Initial & - & - & - & 64.38 $\pm$ 11.17\\
VM        & 86.04 $\pm$ 5.11 & 85.77 $\pm$ 5.02 & 85.32 $\pm$ 5.34 & 85.71 $\pm$ 5.01 \\
TM        & 85.95 $\pm$ 5.22 & 85.32 $\pm$ 5.35 & 85.73 $\pm$ 5.91 & 85.66 $\pm$ 5.39 \\
LKU       & 85.43 $\pm$ 5.41 & 85.26 $\pm$ 5.06 & 84.14 $\pm$ 6.05 & 84.94 $\pm$ 5.24 \\
EP(w/o F) & 84.54 $\pm$ 6.25 & 85.61 $\pm$ 5.22 & 84.92 $\pm$ 6.21 & 85.02 $\pm$ 5.66 \\
EP(w/ F) & 86.88 $\pm$ 4.47 & 86.73 $\pm$ 4.72 & 86.86 $\pm$ 4.43 & 86.82 $\pm$ 4.30 \\
\bottomrule
\end{tabular}
\end{table}

\clearpage
\begin{table}[hbt!]
\centering
\setlength{\tabcolsep}{2.0pt}
\caption{Registration performance on OASIS-1.
Comparison of five methods on the OASIS-1 test set. Values are mean$\pm$s.d. over test cases. Metrics: Dice (\%), NMI, $|J|_{ < 0}\%$, and std($J$).}
\label{tab:supp_results_oasis}
\begin{tabular}{lcccc}
\toprule
Method & Seed 1 & Seed 2 & Seed 3 & Averaged \\
\midrule
\multicolumn{5}{l}{\textbf{Dice (\%)}} \\
\midrule
Initial & - & - & - & 54.40 $\pm$ 8.85 \\
VM        & 76.71 $\pm$ 4.01 & 76.76 $\pm$ 3.91 & 76.68 $\pm$ 3.85 & 76.72 $\pm$ 3.87 \\
TM        & 77.64 $\pm$ 3.85 & 77.67 $\pm$ 3.91 & 77.54 $\pm$ 3.78 & 77.62 $\pm$ 3.80 \\
LKU       & 77.67 $\pm$ 3.89 & 77.68 $\pm$ 3.89 & 77.66 $\pm$ 3.92 & 77.67 $\pm$ 3.85 \\
EP(w/o F) & 77.13 $\pm$ 3.93 & 77.23 $\pm$ 3.92 & 77.38 $\pm$ 3.93 & 77.25 $\pm$ 3.89 \\
EP(w/ F)  & 77.68 $\pm$ 3.84 & 77.78 $\pm$ 3.90 & 77.65 $\pm$ 3.80 & 77.70 $\pm$ 3.79 \\
\bottomrule
\end{tabular}
\end{table}

\begin{table}[hbt!]
\centering
\setlength{\tabcolsep}{2.0pt}
\caption{Registration performance on MindBoggle-101. Comparison of five methods on the MindBoggle-101 test set. Values are mean$\pm$s.d. over test cases. Metrics: Dice (\%), HD95 (mm), ASSD (mm), NMI, $|J|_{ < 0}\%$, and std($J$).}
\label{tab:supp_results_mind101}
\begin{tabular}{lcccc}
\toprule
Method & Seed 1 & Seed 2 & Seed 3 & Averaged \\
\midrule
\multicolumn{5}{l}{\textbf{Dice (\%)}} \\
\midrule
Initial & - & - & - & 34.54 $\pm$ 1.88 \\
VM        & 55.72 $\pm$ 3.13 & 55.63 $\pm$ 3.24 & 55.72 $\pm$ 3.24 & 55.69 $\pm$ 3.20 \\
TM        & 57.71 $\pm$ 3.14 & 57.72 $\pm$ 3.15 & 57.64 $\pm$ 3.25 & 57.69 $\pm$ 3.17 \\
LKU       & 57.92 $\pm$ 3.46 & 57.98 $\pm$ 3.43 & 58.00 $\pm$ 3.52 & 57.97 $\pm$ 3.45 \\
EP(w/o F) & 57.47 $\pm$ 3.29 & 57.16 $\pm$ 3.35 & 57.39 $\pm$ 3.35 & 57.34 $\pm$ 3.33 \\
EP(w/ F)  & 58.10 $\pm$ 3.25 & 58.07 $\pm$ 3.30 & 58.10 $\pm$ 3.22 & 58.09 $\pm$ 3.25 \\
\bottomrule
\end{tabular}
\end{table}

\begin{table}[hbt!]
\centering
\setlength{\tabcolsep}{2.0pt}
\caption{Registration performance on Lung-CT. Comparison of five methods on the Lung-CT test set. Values are mean$\pm$s.d. over test cases. Metrics: Dice (\%), HD95 (mm), ASSD (mm), NMI, $|J|_{ < 0}\%$, and std($J$).}
\label{tab:supp_results_lungct}
\begin{tabular}{lcccc}
\toprule
Method & Seed 1 & Seed 2 & Seed 3 & Averaged \\
\midrule
\multicolumn{5}{l}{\textbf{Dice (\%)}} \\
\midrule
Initial          & - & - & - & 84.64 $\pm$ 5.94 \\
VM        & 92.19 $\pm$ 3.54 & 94.33 $\pm$ 1.39 & 94.92 $\pm$ 1.60 & 93.82 $\pm$ 1.77 \\
TM        & 93.38 $\pm$ 3.48 & 91.75 $\pm$ 5.26 & 94.80 $\pm$ 2.37 & 93.31 $\pm$ 3.40 \\
LKU       & 96.08 $\pm$ 0.82 & 93.51 $\pm$ 4.12 & 95.49 $\pm$ 1.55 & 95.03 $\pm$ 2.08 \\
EP(w/o F) & 96.15 $\pm$ 1.23 & 94.02 $\pm$ 4.40 & 95.80 $\pm$ 1.79 & 95.33 $\pm$ 1.92 \\
EP(w/ F)  & 96.18 $\pm$ 1.11 & 95.99 $\pm$ 1.14 & 96.37 $\pm$ 1.01 & 96.18 $\pm$ 1.08 \\
\bottomrule
\end{tabular}
\end{table}

\begin{table}[hbt!]
\centering
\setlength{\tabcolsep}{2.0pt}
\caption{Registration performance on Abdomen-CT. Comparison of five methods on the Abdomen-CT test set. Values are mean$\pm$s.d. over test cases. Metrics: Dice (\%), HD95 (mm), ASSD (mm), NMI, $|J|_{ < 0}\%$, and std($J$).}
\label{tab:supp_results_Abdomen-CT}
\begin{tabular}{lcccc}
\toprule
Method & Seed 1 & Seed 2 & Seed 3 & Averaged \\
\midrule
\multicolumn{5}{l}{\textbf{Dice (\%)}} \\
\midrule
Initial & - & -& -& 35.13 $\pm$ 0.10 \\
VM        & 46.43 $\pm$ 11.71 & 46.05 $\pm$ 11.52 & 45.67 $\pm$ 11.75 & 46.05 $\pm$ 11.65 \\
TM        & 48.39 $\pm$ 12.32 & 48.93 $\pm$ 12.42 & 48.23 $\pm$ 12.00 & 48.52 $\pm$ 12.22 \\
LKU       & 48.04 $\pm$ 12.21 & 48.08 $\pm$ 12.04 & 48.51 $\pm$ 12.02 & 48.21 $\pm$ 12.06 \\
EP(w/o F) & 50.41 $\pm$ 12.39 & 49.68 $\pm$ 12.37 & 49.89 $\pm$ 12.60 & 49.99 $\pm$ 12.44 \\
EP(w/ F)  & 52.46 $\pm$ 12.70 & 51.89 $\pm$ 12.77 & 51.44 $\pm$ 12.66 & 51.93 $\pm$ 12.69 \\
\bottomrule
\end{tabular}
\end{table}

\begin{table}[hbt!]
\centering
\setlength{\tabcolsep}{2.0pt}
\caption{Registration performance on BCV-CT. Comparison of five methods on the BCV-CT test set. Values are mean$\pm$s.d. over test cases. Metrics: Dice (\%), HD95 (voxels), ASSD (voxels), NMI, $|J|_{ < 0}\%$, and std($J$).}
\label{tab:supp_results_BCV-CT}
\begin{tabular}{lcccc}
\toprule
Method & Seed 1 & Seed 2 & Seed 3 & Averaged \\
\midrule
\multicolumn{5}{l}{\textbf{Dice (\%)}} \\
\midrule
Initial & - & - & - & 40.85 $\pm$ 11.82 \\
VM          & 56.23 $\pm$ 13.99 & 56.61 $\pm$ 13.95 & 55.75 $\pm$ 13.90 & 56.20 $\pm$ 13.93 \\
TM          & 60.40 $\pm$ 14.47 & 60.54 $\pm$ 14.28 & 60.49 $\pm$ 14.28 & 60.48 $\pm$ 14.32 \\
LKU         & 59.29 $\pm$ 14.25 & 59.79 $\pm$ 14.00 & 58.69 $\pm$ 14.13 & 59.26 $\pm$ 14.09 \\
EP(w/o F)   & 61.10 $\pm$ 14.77 & 60.82 $\pm$ 14.90 & 61.02 $\pm$ 14.80 & 60.98 $\pm$ 14.81 \\
EP(w/ F)    & 64.13 $\pm$ 14.67 & 64.79 $\pm$ 14.42 & 64.49 $\pm$ 14.55 & 64.47 $\pm$ 14.51 \\
\bottomrule
\end{tabular}
\end{table}

\clearpage
\begin{table}[hbt!]
\centering
\setlength{\tabcolsep}{2.0pt}
\caption{Registration performance on CHAOS-MR. Comparison of five methods on the CHAOS-MR test set. Values are mean$\pm$s.d. over test cases. Metrics: Dice (\%), HD95 (voxels), ASSD (voxels), NMI, $|J|_{ < 0}\%$, and std($J$).}
\label{tab:supp_results_CHAOS-MR}
\begin{tabular}{lcccc}
\toprule
Method & Seed 1 & Seed 2 & Seed 3 & Averaged \\
\midrule
\multicolumn{5}{l}{\textbf{Dice (\%)}} \\
\midrule
Initial & - & - & - & 44.79 $\pm$ 9.54 \\
VM          & 65.21 $\pm$ 10.12 & 65.37 $\pm$ 10.05 & 65.44 $\pm$ 10.00 & 65.34 $\pm$ 10.03 \\
TM          & 68.15 $\pm$ 10.13 & 67.87 $\pm$ 10.11 & 67.78 $\pm$ 9.97 & 67.93 $\pm$ 10.04 \\
LKU         & 68.01 $\pm$ 9.82 & 68.37 $\pm$ 9.63 & 68.45 $\pm$ 9.63 & 68.27 $\pm$ 9.66 \\
EP(w/o F)   & 66.67 $\pm$ 10.05 & 66.89 $\pm$ 9.98 & 66.62 $\pm$ 9.96 & 66.73 $\pm$ 9.98 \\
EP(w/ F)    & 68.18 $\pm$ 9.57 & 69.06 $\pm$ 9.12 & 68.29 $\pm$ 9.52 & 68.51 $\pm$ 9.36 \\
\bottomrule
\end{tabular}
\end{table}

\begin{table}[hbt!]
\centering
\setlength{\tabcolsep}{2.0pt}
\caption{Registration performance on BCV--CHAOS. Comparison of five methods on the BCV--CHAOS test set. Values are mean$\pm$s.d. over test cases. Metrics: Dice (\%), HD95 (voxels), ASSD (voxels), NMI, $|J|_{ < 0}\%$, and std($J$).}
\label{tab:supp_results_BCV-CTCHAOS-MR}
\begin{tabular}{lcccc}
\toprule
Method & Seed 1 & Seed 2 & Seed 3 & Averaged \\
\midrule
\multicolumn{5}{l}{\textbf{Dice (\%)}} \\
\midrule
Initial & - & - & - & 23.96 $\pm$ 11.39 \\
VM          & 25.96 $\pm$ 13.56 & 25.23 $\pm$ 13.19 & 26.23 $\pm$ 13.26 & 25.81 $\pm$ 13.33 \\
TM          & 35.60 $\pm$ 15.51 & 32.10 $\pm$ 15.43 & 37.76 $\pm$ 15.64 & 35.15 $\pm$ 15.47 \\
LKU         & 26.52 $\pm$ 13.24 & 34.28 $\pm$ 14.31 & 26.14 $\pm$ 12.78 & 28.98 $\pm$ 13.36 \\
EP(w/o F)   & 51.09 $\pm$ 13.94 & 49.72 $\pm$ 14.27 & 52.26 $\pm$ 13.76 & 51.02 $\pm$ 13.92 \\
EP(w/ F)    & 51.88 $\pm$ 13.15 & 52.40 $\pm$ 13.51 & 54.59 $\pm$ 12.29 & 52.96 $\pm$ 12.90 \\
\bottomrule
 \end{tabular}
\end{table}

\clearpage
\begin{table}[hbt!]
\centering
\setlength{\tabcolsep}{2.0pt}
\caption{Registration performance on Abdomen-CT--MR. Comparison of five methods on the Abdomen-CT--MR test set. Values are mean$\pm$s.d. over test cases. Metrics: Dice (\%), HD95 (mm), ASSD (mm), NMI, $|J|_{ < 0}\%$, and std($J$).}
\label{tab:supp_results_l2rctmr}
\begin{tabular}{lcccc}
\toprule
Method & Seed 1 & Seed 2 & Seed 3 & Averaged \\
\midrule
\multicolumn{5}{l}{\textbf{Dice (\%)}} \\
\midrule
Initial     & - & - & - & 37.32 $\pm$ 16.12 \\
VM          & 78.52 $\pm$ 15.83 & 81.06 $\pm$ 9.29 & 84.00 $\pm$ 7.11 & 81.19 $\pm$ 10.23 \\
TM          & 77.81 $\pm$ 19.23 & 76.32 $\pm$ 24.61 & 83.42 $\pm$ 10.53 & 79.18 $\pm$ 18.04 \\
LKU         & 80.43 $\pm$ 16.02 & 86.11 $\pm$ 4.61 & 84.93 $\pm$ 7.02 & 83.82 $\pm$ 8.80 \\
EP(w/o F)   & 86.20 $\pm$ 5.26 & 85.27 $\pm$ 5.37 & 86.36 $\pm$ 5.29 & 85.94 $\pm$ 5.18 \\
EP(w/ F)    & 87.23 $\pm$ 4.45 & 87.14 $\pm$ 4.79 & 87.41 $\pm$ 4.60 & 87.26 $\pm$ 4.61 \\
\bottomrule
\end{tabular}
\end{table}

\clearpage
\begin{table}[hbt!]
\centering
\setlength{\tabcolsep}{2.0pt}
\caption{\textbf{Training and inference compute and relative reductions versus baselines.} Values report training compute and inference compute (Giga-Multi-Adds).
Reductions are computed as $100\times(1-\text{Cost}_{\text{ours}}/\text{Cost}_{\text{comparator}})$.}
\label{tab:supp_costs}
\begin{tabular}{lcccc}
\toprule
\multicolumn{5}{c}{\textbf{EP$^\star$ cost and reduction (\%) vs VM/TM/LKU}} \\
\midrule
& \multicolumn{2}{c}{Cost} & \multicolumn{2}{c}{Reduction(\%)} \\
\cmidrule(lr){2-3}\cmidrule(lr){4-5}
Task & \multicolumn{2}{c}{Train\&Infer} & \multicolumn{2}{c}{Train\&Infer} \\
\midrule
Fluo-N2DH-GOWT1 & \multicolumn{2}{c}{1.01} & \multicolumn{2}{c}{31.76/83.97/83.28} \\
ACDC & \multicolumn{2}{c}{0.80} & \multicolumn{2}{c}{-116.69/49.97/46.99} \\
OASIS-1 & \multicolumn{2}{c}{1.78} & \multicolumn{2}{c}{-156.98/40.07/37.10} \\
MindBoggle-101 & \multicolumn{2}{c}{154.48} & \multicolumn{2}{c}{45.84/67.12/79.71} \\
Lung-CT & \multicolumn{2}{c}{168.96} & \multicolumn{2}{c}{62.03/78.59/85.80} \\
\bottomrule
\end{tabular}
\end{table}
\vspace{-20pt}
\noindent\textsuperscript{$\star$} In the first EP block, the columns are labelled “Train\&Infer,” because the train and inference costs are identical for EP without features.

\newpage
\section{Feature-level analysis of foundation encoders}
\label{sec:supp:model_choice}
\counterwithin{figure}{section}
\counterwithin{table}{section}

We systematically evaluated the effect of 47 encoding features on registration accuracy. (Details of the features are provided in Section~\ref{sec:method}.) The experimental results revealed markedly heterogeneous effects of different features on registration performance. When performance changes relative to baseline EP were averaged across all 24 subtasks (Table~\ref{tab:supp_averged_47_rate}), 34 of the 47 evaluated features led to an average improvement, whereas 13 resulted in performance degradation. The most substantial performance gain was observed with the feature from DINOv2-S-D6, which achieved an average improvement of 0.94 and a positive improvement rate of 75.00\%. This was closely followed by DINOv3-S-D3 (average improvement: 0.85; positive improvement rate: 83.33\%), DINOv3-S-D6 (average improvement: 0.76; positive improvement rate: 75.00\%), and DINOv2-S-D3 (average improvement: 0.74; positive improvement rate: 79.17\%). Other notable features include DINOv3-S-D9 (improvement: 0.73; positive improvement rate: 79.17\%), FCN-L2 (improvement: 0.71; positive improvement rate: 87.50\%), and SAM-D9 (improvement: 0.69; positive improvement rate: 70.83\%).

Conversely, several features significantly impaired registration accuracy. The most detrimental was SAM2-Large, which caused an average performance decrease of -1.85. Other features from the SAM2 series also led to considerable declines: SAM2-Small (-1.01), SAM2-Base (-0.95), and SAM2-Tiny (-0.83). Additionally, the FCN-Cls1 resulted in a notable performance drop of 1.36, and DeepLabv3-ASPP dropped 1.14. Different with the early layers of DINOv3, both DINOv3-S-D12  and DINOv3-B-D12 lead to a decrement in registration performance.

Regarding the rate of positive impact, DINOv3-S-D1 and SAM-D3 demonstrated the highest consistency, both had the positive improvement rate of 91.67\%. Features from FCN-L2 and FCN-L1 also showed high reliability, both achieving a 87.5\% positive improvement rate. In contrast, the embedding features of both SAM and MedSAM, along with several classifier-based FCN and DeepLabv3 features, exhibited a higher likelihood of negative effects, with positive probabilities falling around or below 40\%.

These findings demonstrate that feature selection is critical for learning-based registration, where intermediate neural layers (e.g., from FCN, DINOv2, DINOv3, and specific depths of SAM) generally provide more robust and beneficial representations, whereas deeper layers such as certain embedding outputs, classifier-based features, and features from the SAM2 model family tend to degrade performance.

\begin{table}[]
\caption{The effect of 47 encoding features on registration accuracy averaged over all testing images from 24 sub-registration tasks.}
\label{tab:supp_averged_47_rate}
\begin{tabular}{lrr}\hline
Features       & Average Improvement in Dice (\%) & Positive improvement rate (\%) \\ \hline
ResNet18       & 0.40                                                 & 83.33                                              \\
ResNet152      & -0.11                                                & 62.50                                              \\
FCN-MP         & 0.20                                                 & 66.67                      \\
FCN-L1         & 0.65                                                 & 87.50                                              \\
FCN-L2         & 0.71                                                 & 87.50                                              \\
FCN-Cls1       & -1.36                                                & 37.50                                              \\
FCN-ClsF       & -0.87                                                & 29.17                                              \\
DeepLabv3-MP   & 0.28                                                 & 75.00                                              \\
DeepLabv3-L1   & 0.25                                                 & 79.17                                              \\
DeepLabv3-ASPP & -1.14                                                & 16.67                                              \\
DeepLabv3-ClsF & -0.57                                                & 29.17                                              \\
DINOv2-S-D1    & 0.51                                                 & 79.17                                              \\
DINOv2-S-D3    & 0.74                                                 & 79.17                                              \\
DINOv2-S-D6    & 0.94                                                 & 75.00                                              \\
DINOv2-S-D9    & 0.27                                                 & 66.67                                              \\
DINOv2-S-D12   & 0.11                                                 & 70.83                                              \\
DINOv2-B-D1    & 0.37                                                 & 70.83                                              \\
DINOv2-B-D3    & 0.69                                                 & 75.00                                              \\
DINOv2-B-D6    & 0.58                                                 & 70.83                                              \\
DINOv2-B-D9    & 0.30                                                 & 79.17                                              \\
DINOv2-B-D12   & 0.06                                                 & 79.17                                              \\
DINOv3-S-D1    & 0.63                                                 & 91.67                                              \\
DINOv3-S-D3    & 0.85                                                 & 83.33                                              \\
DINOv3-S-D6    & 0.76                                                 & 75.00                                              \\
DINOv3-S-D9    & 0.73                                                 & 79.17                                              \\
DINOv3-S-D12   & -0.05                                                & 66.67                                              \\
DINOv3-B-D1    & 0.58                                                 & 79.17                                              \\
DINOv3-B-D3    & 0.68                                                 & 79.17                                              \\
DINOv3-B-D6    & 0.67                                                 & 75.00                                              \\
DINOv3-B-D9    & 0.49                                                 & 75.00                                              \\
DINOv3-B-D12   & -0.27                                                & 75.00                                              \\
SAM-D1         & 0.47                                                 & 75.00                                              \\
SAM-D3         & 0.57                                                 & 91.67                                              \\
SAM-D6         & 0.47                                                 & 70.83                                              \\
SAM-D9         & 0.69                                                 & 70.83                                              \\
SAM-D12        & 0.64                                                 & 70.83                                              \\
SAM-Emb        & -0.67                                                & 37.50                                              \\
MedSAM-D1      & 0.14                                                 & 75.00                                              \\
MedSAM-D3      & 0.22                                                 & 66.67                                              \\
MedSAM-D6      & 0.57                                                 & 70.83                                              \\
MedSAM-D9      & 0.21                                                 & 75.00                                              \\
MedSAM-D12     & 0.40                                                 & 75.00                                              \\
MedSAM-Emb     & -0.56                                                & 41.67                                              \\
SAM2-T         & -0.83                                                & 25.00                                              \\
SAM2-S         & -1.01                                                & 25.00                                              \\
SAM2-B         & -0.95                                                & 33.33                                              \\
SAM2-L         & -1.85                                                & 25.00    \\\hline                                         
\end{tabular}
\end{table}

\newpage
\section{Effect of feature-extraction orientation on volumetric registration}
\label{sec:supp:orientation}
\counterwithin{figure}{section}
\counterwithin{table}{section}

This Supplementary Note provides the dataset-specific Dice, HD95, ASSD and deformation-quality analysis supporting the main-text section on ``Does orientation matter in 2.5D extraction?''. The reported values are provided in Table~\ref{tab:supp_3views_orientation}. In addition to anatomical overlap and surface-distance metrics, we report the percentage of negative Jacobian determinants, $|J|_{<0}\%$, and the standard deviation of the Jacobian determinant, std($J$).


\paragraph{MindBoggle-101.}
Feature-extraction orientation had negligible practical effect on MindBoggle-101. Dice scores were nearly identical across sagittal, coronal and axial extraction, with values of 58.07\%, 58.09\% and 57.91\%, respectively. Surface-distance metrics were also highly similar, with HD95 ranging from 6.49 to 6.53 mm and ASSD from 1.21 to 1.22 mm. Deformation-quality metrics followed the same pattern: $|J|_{<0}\%$ ranged only from 0.87\% to 0.89\%, and std($J$) from 0.54 to 0.55. Thus, although pairwise tests may identify statistically significant differences, the absolute effect size was very small for both registration accuracy and deformation regularity.

\paragraph{Lung-CT.}
All three orientations achieved very similar Dice scores on Lung-CT, ranging from 96.15\% to 96.18\%. However, boundary-distance metrics showed larger orientation-dependent variation. Coronal extraction achieved the lowest HD95 and ASSD, with values of 6.25 mm and 1.16 mm, compared with 7.54 mm and 1.36 mm for sagittal extraction and 9.92 mm and 1.48 mm for axial extraction. The Jacobian metrics showed a more nuanced pattern. Coronal extraction produced the lowest std($J$), 3.79, indicating the smoothest overall distribution of local volume changes, whereas sagittal extraction produced the lowest folding rate, with $|J|_{<0}=0.32\%$ compared with 0.72\% for coronal and 0.76\% for axial extraction. Thus, Dice was insensitive to orientation, surface distances and std($J$) favoured coronal extraction, and $|J|_{<0}\%$ favoured sagittal extraction.

\paragraph{Abdomen-CT.}
In Abdomen-CT, sagittal extraction produced the best overall performance, achieving 51.93\% Dice, 33.63 mm HD95 and 8.87 mm ASSD. Coronal and axial extraction were slightly lower in Dice, with values of 51.06\% and 50.78\%, respectively. The corresponding HD95 and ASSD differences were small relative to the large inter-case variability in this dataset. Deformation quality was also slightly better with sagittal extraction, which achieved the lowest $|J|_{<0}\%$ and std($J$), with values of 3.22\% and 0.64, respectively. Coronal and axial extraction produced similar but marginally higher folding rates and Jacobian variability. Overall, the orientation effect was modest, with sagittal extraction providing a small but consistent advantage across accuracy, surface-distance and deformation-quality metrics.

\paragraph{BCV-CT.}
Orientation had a limited effect on BCV-CT performance. Coronal extraction achieved the highest Dice score, 64.47\%, and the lowest HD95 and ASSD, 17.07 voxels and 4.30 voxels, respectively. Sagittal and axial extraction were close, with Dice scores of 63.77\% and 63.61\%. The maximum Dice difference across orientations was 0.86 points, and the surface-distance differences were similarly small. The Jacobian metrics were also similar across orientations: $|J|_{<0}\%$ ranged from 3.39\% to 3.55\%, and std($J$) ranged from 0.67 to 0.70. Coronal extraction had the lowest std($J$), whereas sagittal extraction had the lowest folding rate. These results suggest that BCV-CT registration is largely robust to the choice of extraction plane, with coronal extraction providing a slight advantage in overlap, surface distance and overall Jacobian variability.

\paragraph{CHAOS-MR.}
CHAOS-MR also showed only minor orientation-dependent differences. Sagittal extraction achieved the highest Dice score, 68.51\%, and the lowest HD95 and ASSD, 17.84 voxels and 4.47 voxels. Coronal extraction was nearly identical, with 67.99\% Dice, 18.04 voxels HD95 and 4.59 voxels ASSD. Axial extraction was also close, with 68.00\% Dice, 18.11 voxels HD95 and 4.58 voxels ASSD. Deformation-quality differences were small: axial extraction produced the lowest $|J|_{<0}\%$ and std($J$), with values of 0.47\% and 0.54, whereas sagittal extraction gave 0.49\% and 0.55. Therefore, sagittal extraction was slightly favoured by Dice and surface-distance metrics, while axial extraction was marginally favoured by Jacobian-based regularity metrics. The absolute differences were small, indicating limited practical sensitivity to orientation.

\paragraph{BCV--CHAOS.}
BCV--CHAOS showed the strongest orientation dependence among the evaluated datasets. Coronal extraction achieved the highest Dice score, 52.96\%, closely followed by sagittal extraction at 52.54\%. However, sagittal extraction yielded slightly better surface distances, with 21.28 voxels HD95 and 6.50 voxels ASSD compared with 21.74 voxels and 6.61 voxels for coronal extraction. Axial extraction performed substantially worse in terms of overlap and boundary accuracy, reducing Dice to 46.04\% and increasing HD95 and ASSD to 24.25 voxels and 7.79 voxels. In contrast, the Jacobian metrics favoured axial extraction, which produced the lowest $|J|_{<0}\%$ and std($J$), with values of 8.57\% and 1.35. Coronal extraction provided the best compromise among the high-accuracy orientations, with lower $|J|_{<0}\%$ and std($J$) than sagittal extraction. These results suggest that orientation choice is most important in difficult inter-subject cross-modality CT--MR registration, where improving anatomical overlap can trade off against deformation regularity.

\paragraph{Abdomen-CT--MR.}
Abdomen-CT--MR showed small but consistent improvements with axial extraction. Dice increased from 86.97\% with sagittal extraction and 87.10\% with coronal extraction to 87.26\% with axial extraction. Surface-distance metrics followed the same trend, with axial extraction achieving the lowest HD95 and ASSD, 11.51 mm and 2.64 mm, compared with 12.61 mm and 2.79 mm for sagittal extraction. The deformation-quality metrics were also best for axial extraction, with the lowest $|J|_{<0}\%$ and std($J$), 1.42\% and 0.51, respectively. Coronal extraction was intermediate, and sagittal extraction was slightly less favourable across all five metrics. Although the absolute differences were modest, axial extraction was consistently favoured across overlap, surface-distance and deformation-regularity measures.

\clearpage
\begin{table}[hbt!]
\centering
\setlength{\tabcolsep}{2.0pt}
\caption{Effect of feature-extraction orientation (Sagittal /Coronal/Axial) on volumetric registration. Values are mean$\pm$s.d. over test cases. Metrics: Dice (\%), HD95, ASSD, $|J|_{ < 0}\%$, and std($J$). HD95 and ASSD are reported in voxels for BCV-CT, CHAOS-MR, and BCV--CHAOS and in millimeters for all other datasets.}
\label{tab:supp_3views_orientation}
\begin{tabular}{lcccccc}
\toprule
View & Dice & HD95 & ASSD  & $|J|_{ < 0}\%$ & std($J$) \\
\midrule
\multicolumn{6}{l}{\textbf{1. MindBoggle-101}} \\
\midrule
Sagittal  (SAM-D9) & 58.07 $\pm$ 3.24 & 6.49 $\pm$ 0.59 & 1.21 $\pm$ 0.15 & 0.89 $\pm$ 0.24 & 0.55 $\pm$ 0.06 \\
Coronal (SAM-D12) & 58.09 $\pm$ 3.25 & 6.53 $\pm$ 0.59 & 1.22 $\pm$ 0.15 & 0.87 $\pm$ 0.23 & 0.54 $\pm$ 0.06 \\
Axial (SAM-D9) & 57.91 $\pm$ 3.14 & 6.53 $\pm$ 0.59 & 1.22 $\pm$ 0.16 & 0.87 $\pm$ 0.23 & 0.54 $\pm$ 0.06 \\
\addlinespace[0.3em]
\multicolumn{6}{l}{\textbf{2. Lung-CT}} \\
\midrule
Sagittal  (DINOv2-B-D3) & 96.17 $\pm$ 1.20 & 7.54 $\pm$ 4.05 & 1.36 $\pm$ 0.60 & 0.32 $\pm$ 0.20 & 5.98 $\pm$ 1.77 \\
Coronal (DINOv2-S-D1) & 96.18 $\pm$ 1.08 & 6.25 $\pm$ 2.55 & 1.16 $\pm$ 0.39 & 0.72 $\pm$ 0.77 & 3.79 $\pm$ 0.53 \\
Axial (FCN-Cls1) & 96.15 $\pm$ 1.13 & 9.92 $\pm$ 3.72 & 1.48 $\pm$ 0.53 & 0.76 $\pm$ 0.13 & 5.05 $\pm$ 0.57 \\
\addlinespace[0.3em]
\multicolumn{6}{l}{\textbf{3. Abdomen-CT}} \\
\midrule
Sagittal (FCN-L1) & 51.93 $\pm$ 12.69 & 33.63 $\pm$ 19.32 & 8.87 $\pm$ 6.51 & 3.22 $\pm$ 1.71 & 0.64 $\pm$ 0.08 \\
Coronal (DINOv3-S-D3) & 51.06 $\pm$ 12.36 & 34.14 $\pm$ 19.34 & 9.17 $\pm$ 6.61 & 3.44 $\pm$ 1.68 & 0.65 $\pm$ 0.08 \\
Axial (FCN-L1) & 50.78 $\pm$ 12.66 & 33.96 $\pm$ 19.37 & 9.12 $\pm$ 6.51 & 3.40 $\pm$ 1.77 & 0.65 $\pm$ 0.08 \\
\addlinespace[0.3em]
\multicolumn{6}{l}{\textbf{4. BCV-CT}} \\
\midrule
Sagittal  (DINOv2-S-D6)  & 63.77 $\pm$ 14.39 & 17.18 $\pm$ 5.31 & 4.36 $\pm$ 2.21 & 3.39 $\pm$ 1.50 & 0.68 $\pm$ 0.08 \\
Coronal  (SAM-D9)  & 64.47 $\pm$ 14.51 & 17.07 $\pm$ 5.45 & 4.30 $\pm$ 2.24 & 3.44 $\pm$ 1.68 & 0.67 $\pm$ 0.08 \\
Axial  (DINOv2-B-D6) & 63.61 $\pm$ 13.72 & 17.48 $\pm$ 5.26 & 4.39 $\pm$ 2.14 & 3.55 $\pm$ 1.52 & 0.70 $\pm$ 0.08 \\
\addlinespace[0.3em]
\multicolumn{6}{l}{\textbf{5. CHAOS-MR}} \\
\midrule
Sagittal   (MedSAM-D12) & 68.51 $\pm$ 9.36 & 17.84 $\pm$ 5.40 & 4.47 $\pm$ 1.83 & 0.49 $\pm$ 0.28 & 0.55 $\pm$ 0.15 \\
Coronal (DINOv2-S-D6) & 67.99 $\pm$ 9.57 & 18.04 $\pm$ 5.56 & 4.59 $\pm$ 1.93 & 0.53 $\pm$ 0.29 & 0.55 $\pm$ 0.15 \\
Axial (SAM-D9) & 68.00 $\pm$ 9.57 & 18.11 $\pm$ 5.51 & 4.58 $\pm$ 1.90 & 0.47 $\pm$ 0.28 & 0.54 $\pm$ 0.14 \\
\addlinespace[0.3em]
\multicolumn{6}{l}{\textbf{6. BCV--CHAOS}} \\
\midrule
Sagittal  (MedSAM-D12) & 52.54 $\pm$ 13.06 & 21.28 $\pm$ 6.44 & 6.50 $\pm$ 2.64 & 9.39 $\pm$ 1.43 & 1.50 $\pm$ 0.08 \\
Coronal  (DINOv3-B-D9) & 52.96 $\pm$ 12.90 & 21.74 $\pm$ 6.62 & 6.61 $\pm$ 2.75 & 9.06 $\pm$ 1.12 & 1.45 $\pm$ 0.07 \\
Axial (DINOv3-B-D1)    & 46.04 $\pm$ 14.72 & 24.25 $\pm$ 7.65 & 7.79 $\pm$ 3.40 & 8.57 $\pm$ 1.32 & 1.35 $\pm$ 0.08 \\
\addlinespace[0.3em]
\multicolumn{6}{l}{\textbf{7. Abdomen-CT--MR}} \\
\midrule
Sagittal (MedSAM-D9)    & 86.97 $\pm$ 4.75 & 12.61 $\pm$ 6.18 & 2.79 $\pm$ 1.25 & 1.59 $\pm$ 0.85 & 0.53 $\pm$ 0.08 \\
Coronal (MedSAM-D6)     & 87.10 $\pm$ 4.58 & 12.05 $\pm$ 6.05 & 2.72 $\pm$ 1.19 & 1.53 $\pm$ 0.83 & 0.53 $\pm$ 0.07 \\
Axial (DINOv3-B-D9)  & 87.26 $\pm$ 4.61 & 11.51 $\pm$ 6.20 & 2.64 $\pm$ 1.22 & 1.42 $\pm$ 0.90 & 0.51 $\pm$ 0.08 \\
\bottomrule
\end{tabular}
\end{table}

\newpage
\section{Comparative analysis of feature families}
\label{sec:supp:fature_families}
\counterwithin{figure}{section}
\counterwithin{table}{section}

This Supplementary Note provides dataset-specific analyses supporting the main-text comparison of CNN versus ViT features, general versus medical features, model scale, and newer versus older foundation models. The values reported in this section can be assessed from Tables~\ref{tab:supp_cnn_vit}, \ref{tab:supp_sam_medsam}, \ref{tab:supp:resnet18_vs_152}, \ref{tab:dinov2_s_vs_b}, \ref{tab:supp_dinov3_s_b}, and Fig.~\ref{supp:fig:depth}. 


\subsection{CNN features vs. ViT features}

We compared the best-performing CNN-based and ViT-based feature representations for each of the ten registration tasks. Overall, ViT features achieved higher Dice scores in eight of ten tasks, whereas CNN features achieved higher Dice scores in two tasks. However, the magnitude and statistical significance of these differences varied substantially across datasets. For datasets with surface-distance results, ViT features also more often improved HD95 and ASSD, although the advantage was task-dependent. The Jacobian metrics showed a broadly similar but not identical pattern: ViT features reduced $|J|_{<0}\%$ relative to the best CNN feature in seven of ten tasks and produced comparable or lower std($J$) in seven of ten tasks. Thus, ViT features were more frequently favoured by both overlap accuracy and deformation quality, but CNN features remained preferable in selected datasets.
\begin{itemize}
    \item Fluo-N2DH-GOWT1. CNN features performed best on Fluo-N2DH-GOWT1. FCN-L1 achieved a Dice score of 94.31\%, compared with 94.07\% for the best ViT feature, DINOv3-S-D12. Although the absolute difference was small, the paired comparison was statistically significant (p$<$0.0001). The deformation metrics also favoured FCN-L1, which produced lower $|J|_{<0}\%$ and std($J$), 1.62\% and 1.09, compared with 1.83\% and 1.60 for DINOv3-S-D12.
    \item ACDC. SAM-D1 achieved 86.82\% Dice, whereas DeepLabv3-L1 achieved 86.70\%; the difference was not statistically significant (p=0.3725). Surface distances were also nearly identical. SAM-D1 had slightly lower HD95, 4.38 mm versus 4.42 mm, whereas DeepLabv3-L1 had slightly lower ASSD, 1.43 mm versus 1.44 mm. The Jacobian metrics slightly favoured SAM-D1, with $|J|_{<0}=0.29\%$ and std($J$)=0.21 compared with 0.31\% and 0.22 for DeepLabv3-L1.
    \item OASIS-1. The best ViT feature, MedSAM-D6, achieved 77.70\% Dice, compared with 77.60\% for the best CNN feature, FCN-L2. The difference was statistically significant (p=0.0030), although the absolute gain was modest. MedSAM-D6 also slightly reduced $|J|_{<0}\%$, 1.26\% versus 1.28\%, while std($J$) was identical at the reported precision.
    \item MindBoggle-101. SAM-D12 achieved 58.09\% Dice, compared with 57.97\% for FCN-L2, with a statistically significant difference (p$<$0.0001). Surface distances were effectively unchanged: both methods achieved 6.53 mm HD95 and 1.22 mm ASSD at the reported precision. The Jacobian metrics were also identical, with $|J|_{<0}=0.87\%$ and std($J$)=0.54. Thus, the ViT feature improved Dice without materially altering boundary accuracy or deformation quality.
    \item Lung-CT. DINOv2-S-D1 and FCN-L2 were nearly equivalent in Dice, 96.18\% and 96.17\%, respectively, and the difference was not statistically significant (p=0.8438). However, the surface-distance and deformation-quality metrics strongly favoured DINOv2-S-D1. It reduced HD95 from 8.55 mm to 6.25 mm, ASSD from 1.49 mm to 1.16 mm, $|J|_{<0}\%$ from 1.86\% to 0.72\%, and std($J$) from 7.95 to 3.79. Thus, although Dice did not separate the two feature families, the ViT feature produced better boundary alignment and a more regular deformation field.
    \item Abdomen-CT. CNN features achieved the highest Dice score on Abdomen-CT. FCN-L1 reached 51.93\%, compared with 51.75\% for DINOv3-S-D3; this difference was not statistically significant (p=0.1919). FCN-L1 also achieved better surface distances, with 33.63 mm HD95 and 8.87 mm ASSD compared with 34.20 mm and 9.09 mm for DINOv3-S-D3. The Jacobian metrics further favoured FCN-L1, with $|J|_{<0}=3.22\%$ and std($J$)=0.64 compared with 3.89\% and 0.67.
    \item BCV-CT. SAM-D9 achieved 64.47\% Dice, compared with 63.60\% for FCN-L2, with p$<$0.0001. The ViT feature was also better in surface-distance metrics, reducing HD95 from 17.46 to 17.07 voxels and ASSD from 4.49 to 4.30 voxels. Deformation quality also improved, with $|J|_{<0}\%$ decreasing from 4.50\% to 3.44\% and std($J$) from 0.74 to 0.67. These results support a ViT advantage for both anatomical overlap and deformation regularity in this abdominal CT benchmark.
    \item CHAOS-MR. MedSAM-D12 achieved 68.51\% Dice, compared with 67.61\% for FCN-L1, with p$<$0.0001. It also improved surface-distance metrics, reducing HD95 from 18.18 to 17.84 voxels and ASSD from 4.63 to 4.47 voxels. The folding rate was slightly lower with MedSAM-D12, 0.49\% versus 0.52\%, while std($J$) was unchanged at 0.55. Thus, the ViT feature improved overlap and boundary alignment without increasing deformation irregularity.
    \item BCV--CHAOS. DINOv3-B-D9 achieved 52.96\% Dice, compared with 51.07\% for FCN-L2, with p$<$0.0001. It also improved surface distances, reducing HD95 from 22.00 to 21.74 voxels and ASSD from 6.77 to 6.61 voxels. The folding rate decreased from 10.13\% to 9.06\%, but std($J$) increased slightly from 1.43 to 1.45. Thus, the ViT feature improved Dice, boundary alignment and folding rate, but with slightly greater variability in local volume change.
    \item Abdomen-CT--MR. DINOv3-B-D9 achieved 87.26\% Dice, compared with 86.46\% for ResNet152, with p=0.0156. It also improved HD95 from 13.29 mm to 11.51 mm and ASSD from 2.90 mm to 2.64 mm. The Jacobian metrics were also better, with $|J|_{<0}\%$ decreasing from 1.62\% to 1.42\% and std($J$) from 0.53 to 0.51. This indicates a consistent ViT advantage across overlap, boundary accuracy and deformation quality.
\end{itemize}

Taken together, ViT features were more frequently optimal than CNN features, especially in brain MRI, abdominal imaging and cross-modality registration. Where surface distances were available, ViT features usually improved or matched HD95 and ASSD, but CNN features remained preferable for Fluo-N2DH-GOWT1 and Abdomen-CT.


\subsection{General features vs. medical features}

We compared SAM and MedSAM features to determine whether medical-domain pre-training consistently improves registration performance. Across the ten benchmarks, MedSAM achieved higher Dice scores in five tasks, whereas SAM achieved higher Dice scores in the other five tasks. The surface-distance and Jacobian metrics were similarly mixed. MedSAM often reduced HD95, ASSD or $|J|_{<0}\%$ in respiratory, abdominal MR and CT--MR tasks, but did not consistently improve Dice or std($J$).


\begin{itemize}
    \item Fluo-N2DH-GOWT1. MedSAM achieved a slightly higher Dice score than SAM, but the difference was not statistically significant (p=0.9516). The deformation metrics slightly favoured SAM, with $|J|_{<0}=2.24\%$ and std($J$)=1.51, compared with 2.25\% and 1.59 for MedSAM.
    \item ACDC. SAM (86.82\%) achieved a higher Dice score than MedSAM (86.37\%), but the difference was not statistically significant (p=0.2491). SAM also had better surface distances, with 4.38 mm HD95 and 1.44 mm ASSD compared with 4.59 mm and 1.51 mm for MedSAM. MedSAM produced a marginally lower folding rate, 0.28\% versus 0.29\%, while std($J$) was identical at 0.21.
    \item OASIS-1. SAM and MedSAM performed almost identically, with Dice scores of 77.69\% and 77.70\%, respectively, and no significant difference (p=0.6917). MedSAM produced a marginally lower $|J|_{<0}\%$, 1.26\% versus 1.28\%, while std($J$) was identical.
    \item MindBoggle-101. SAM achieved a slightly higher Dice score than MedSAM, 58.09\% versus 58.04\%, with p$<$0.0001. SAM also slightly improved HD95, 6.53 mm versus 6.55 mm, while ASSD was identical at 1.22 mm. The deformation metrics also slightly favoured SAM, with $|J|_{<0}=0.87\%$ compared with 0.90\% for MedSAM and identical std($J$).
    \item Lung-CT. SAM (96.09\%) achieved a slightly higher Dice score than MedSAM (96.00\%), but the difference was not statistically significant (p=0.3125). In contrast, MedSAM was better in surface-distance and deformation-quality metrics. MedSAM reduced HD95 from 7.78 mm to 6.76 mm, ASSD from 1.30 mm to 1.27 mm, $|J|_{<0}\%$ from 1.89\% to 0.69\%, and std($J$) from 4.17 to 3.85. Thus, although SAM had a small Dice advantage, MedSAM produced more regular deformation fields and better boundary alignment.
    \item Abdomen-CT. SAM significantly outperformed MedSAM in Dice, 51.11\% versus 50.41\% (p$<$0.0001), and was also slightly better in surface distances, with 34.39 mm HD95 and 9.22 mm ASSD compared with 34.41 mm and 9.29 mm for MedSAM. However, the Jacobian metrics favoured MedSAM, which reduced $|J|_{<0}\%$ from 3.94\% to 3.36\% and std($J$) from 0.67 to 0.64. This indicates an accuracy--regularity trade-off in Abdomen-CT.
    \item BCV-CT. SAM significantly outperformed MedSAM, achieving 64.47\% Dice compared with 63.87\% (p$<$0.0001). SAM also improved HD95 from 17.16 to 17.07 voxels and ASSD from 4.36 to 4.30 voxels. The Jacobian metrics slightly favoured SAM as well, with $|J|_{<0}=3.44\%$ and std($J$)=0.67 compared with 3.51\% and 0.69 for MedSAM.
    \item CHAOS-MR. MedSAM achieved a higher Dice score than SAM, 68.51\% versus 68.05\% with p$<$0.0001. MedSAM also improved surface distances, reducing HD95 from 18.12 to 17.84 voxels and ASSD from 4.57 to 4.47 voxels. The Jacobian metrics were essentially identical, with both features yielding $|J|_{<0}=0.49\%$ and std($J$)=0.55.
    \item BCV--CHAOS. MedSAM showed its clearest Dice advantage on BCV--CHAOS, achieving 52.54\% compared with 51.52\% for SAM (p$<$0.0001). It also improved surface distances, reducing HD95 from 21.94 to 21.28 voxels and ASSD from 6.74 to 6.50 voxels. MedSAM slightly reduced $|J|_{<0}\%$, 9.39\% versus 9.53\%, but increased std($J$), 1.50 versus 1.42. Thus, MedSAM improved overlap, boundary alignment and folding rate, but produced greater variability in local volume change.
    \item Abdomen-CT--MR. MedSAM achieved a higher Dice score than SAM, but the difference was not statistically significant (p=0.0781). MedSAM nevertheless improved surface distances, reducing HD95 from 12.93 mm to 10.82 mm and ASSD from 2.80 mm to 2.58 mm. It also slightly reduced $|J|_{<0}\%$, 1.28\% versus 1.34\%, while std($J$) was identical at 0.50.
\end{itemize}
Overall, MedSAM did not consistently outperform SAM in Dice. However, in several 3D medical-imaging tasks, particularly Lung-CT, CHAOS-MR, BCV--CHAOS and Abdomen-CT--MR, MedSAM improved HD95 and ASSD even when Dice differences were small. The Jacobian metrics also show that medical-domain pre-training can reduce folding in selected datasets, but does not uniformly produce smoother deformation fields.

\subsection{Small model features vs. large model features}

We examined whether increasing model scale improves registration performance by comparing features extracted from the Small and Base variants of DINOv2 and DINOv3, and from the first layer of the shallow ResNet18 and the deeper ResNet152. Across these comparisons, larger models did not consistently produce better registration features. Surface-distance and Jacobian metrics further show that larger models can improve deformation regularity in some settings, but this does not necessarily coincide with higher Dice or better boundary alignment.

\paragraph{ResNet18 versus ResNet152.}
ResNet152 feature achieved higher Dice on Fluo-N2DH-GOWT1 (p$<$0.0001), MindBoggle-101 (p$<$0.0001), CHAOS-MR (p=0.0004) and Abdomen-CT--MR (p=0.1953), whereas ResNet18 achieved higher Dice on ACDC (p$<$0.0001), OASIS-1 (p=0.9871), Lung-CT (p=0.0078), Abdomen-CT (p$<$0.0001), BCV-CT (p$<$0.0001), and BCV--CHAOS (p$<$0.0001). Where surface distances were available, ResNet18 often achieved better HD95 and ASSD, particularly on Lung-CT, Abdomen-CT, BCV-CT and BCV--CHAOS. For example, on Lung-CT, ResNet18 achieved 12.22 mm HD95 and 1.86 mm ASSD, compared with 22.06 mm and 2.78 mm for ResNet152. In contrast, ResNet152 more often produced more regular deformation fields: it reduced $|J|_{<0}\%$ in seven of ten tasks and reduced or matched std($J$) in eight of ten tasks. This was most apparent in Abdomen-CT, BCV-CT and BCV--CHAOS, where ResNet152 had lower folding and lower std($J$) despite lower Dice and worse surface distances. Thus, ResNet18 was frequently stronger for overlap and boundary accuracy, whereas ResNet152 was more often favoured by Jacobian-based regularity.

\paragraph{DINOv2-Small vs. DINOv2-Base.}
DINOv2-Base outperformed DINOv2-Small on Fluo-N2DH-GOWT1 (p$<$0.0001), ACDC (p=0.4043), MindBoggle-101 (p$<$0.0001), CHAOS-MR (p=0.7166), BCV--CHAOS (p=0.0014), and Abdomen-CT--MR (p=0.6406), whereas DINOv2-Small was favoured on OASIS-1 (p=0.4821), Lung-CT (p=0.2500), Abdomen-CT (p$<$0.0001), BCV-CT (p=0.1021). Surface-distance metrics showed no uniform advantage for either scale. DINOv2-Base improved HD95 and ASSD on MindBoggle-101, CHAOS-MR, BCV--CHAOS and Abdomen-CT--MR, whereas DINOv2-Small was better on Lung-CT, Abdomen-CT and BCV-CT. The Jacobian metrics were also task-dependent. DINOv2-Base reduced folding in ACDC, OASIS-1, Lung-CT, Abdomen-CT, BCV--CHAOS and Abdomen-CT--MR, with the largest regularity advantage on BCV--CHAOS, where $|J|_{<0}\%$ decreased from 10.32\% to 7.97\% and std($J$) from 1.51 to 1.36. In contrast, DINOv2-Small produced better Dice, HD95, ASSD and std($J$) on Lung-CT, although DINOv2-Base had the lower folding rate. These results indicate that increasing DINOv2 width can improve deformation regularity in some settings, but does not confer a consistent accuracy or boundary-distance benefit.

\paragraph{DINOv3-Small versus DINOv3-Base.}
The DINOv3 comparison also showed a task-dependent pattern. DINOv3-Base outperformed DINOv3-Small on OASIS-1 (p=0.9822), MindBoggle-101 (p$<$0.0001), CHAOS-MR (p-=0.4473),  BCV--CHAOS (p$<$0.0001), and Abdomen-CT--MR (p=0.1953), whereas DINOv3-Small outperformed DINOv3-Base on Fluo-N2DH-GOWT1 (p$<$0.0001), ACDC (p=0.0858), Lung-CT (p=0.4609), Abdomen-CT (p$<$0.0001), and BCV-CT (p$<$0.0001). The effects of scaling from DINOv3-Small to DINOv3-Base were inconsistent across datasets. DINOv3-Base reduced both HD95 and ASSD on Lung-CT, BCV–CHAOS, and Abdomen-CT--MR. However, it increased both metrics on ACDC. Differences were negligible on MindBoggle-101, BCV-CT, and CHAOS-MR. On Abdomen-CT, HD95 improved, whereas ASSD increased. The Jacobian metrics more often favoured DINOv3-Base: it reduced $|J|_{<0}\%$ in eight of ten tasks and reduced or matched std($J$) in nine of ten tasks. This regularity advantage was especially clear on Lung-CT, where DINOv3-Base reduced std($J$) from 8.14 to 3.89, and on BCV--CHAOS, where it reduced $|J|_{<0}\%$ from 10.05\% to 9.06\% and std($J$) from 1.52 to 1.45. However, on Abdomen-CT and BCV-CT, DINOv3-Small achieved higher Dice despite less favourable Jacobian metrics. Thus, DINOv3-Base more often improved deformation regularity, but was not universally preferable for registration accuracy.

Together, the DINO and ResNet comparisons show that model scale alone does not determine registration performance. Larger or deeper models may improve deformation regularity in selected datasets, but they do not consistently improve Dice, HD95 or ASSD. This supports task-specific feature selection rather than defaulting to the largest available encoder.

\subsection{Newer foundation models vs. earlier foundation models}

We compared DINOv3 with DINOv2 to assess whether newer foundation models consistently provide better features for registration. The results did not support a universal advantage for DINOv3. Dice, surface-distance and Jacobian metrics were all dataset-dependent, and higher Dice sometimes coincided with worse HD95, ASSD or deformation regularity.

\paragraph{Small models.}
For the Small variants, DINOv3-S outperformed DINOv2-S in Dice on Fluo-N2DH-GOWT1 (p$<$0.0001), ACDC (p=0.5025), Abdomen-CT  (p$<$0.0001),  BCV--CHAOS (p=0.0067), and Abdomen-CT--MR (p=0.7422), whereas DINOv2-S was favoured on OASIS-1 (p=0.4375), MindBoggle-101 (p$<$0.0001), Lung-CT  (p=0.1953), BCV-CT  (p=0.8998) and CHAOS-MR  (p=0.0007). Where surface distances were available, DINOv3-S improved HD95 and ASSD on ACDC and Abdomen-CT, but DINOv2-S was better on Lung-CT and BCV-CT. On BCV--CHAOS and Abdomen-CT--MR, DINOv3-S produced slightly higher Dice, but DINOv2-S achieved better HD95 and ASSD. The Jacobian metrics were similarly mixed. DINOv3-S reduced folding or std($J$) in ACDC, CHAOS-MR and BCV--CHAOS, but on Fluo-N2DH-GOWT1 it improved Dice while producing substantially higher $|J|_{<0}\%$ and std($J$), 1.83\% and 1.60 compared with 0.50\% and 0.40 for DINOv2-S. Thus, newer Small-model features improved selected accuracy metrics, but did not consistently improve boundary alignment or deformation quality.

\paragraph{Base models.}
For the Base variants, DINOv3-B outperformed DINOv2-B in Dice on Fluo-N2DH-GOWT1 (p$<$0.0001), OASIS-1 (p=0.2895), BCV--CHAOS (p$<$0.0001) and Abdomen-CT--MR (p=0.1953), whereas DINOv2-B was favoured on ACDC (p=0.2695), MindBoggle-101 (p$<$0.0001), Lung-CT (p=0.7422), Abdomen-CT (p=0.4832), BCV-CT (p$<$0.0001) and CHAOS-MR (p=0.0115). Surface-distance trends were again mixed. DINOv3-B improved HD95 and ASSD on Lung-CT, Abdomen-CT and Abdomen-CT--MR, but DINOv2-B was better on MindBoggle-101, CHAOS-MR and BCV--CHAOS. On BCV--CHAOS, DINOv3-B achieved higher Dice, but DINOv2-B produced better HD95, ASSD, $|J|_{<0}\%$ and std($J$). The Jacobian metrics showed no uniform advantage for either generation: DINOv3-B improved regularity on MindBoggle-101 and BCV-CT, whereas DINOv2-B was more regular on Fluo-N2DH-GOWT1, Abdomen-CT, and BCV--CHAOS. These findings show that newer foundation models do not uniformly improve either registration accuracy, boundary alignment or deformation quality.

Overall, DINOv3 does not uniformly outperform DINOv2. Instead, the preferred model depends on the dataset, feature scale and evaluation endpoint. In several settings, the model with higher Dice did not have the best HD95, ASSD, $|J|_{<0}\%$ or std($J$), reinforcing the need to evaluate both anatomical accuracy and deformation quality when selecting foundation-model features.

\subsection{Comparisons on different depths}
\begin{figure}
    \centering
    \includegraphics[width=0.82\linewidth]{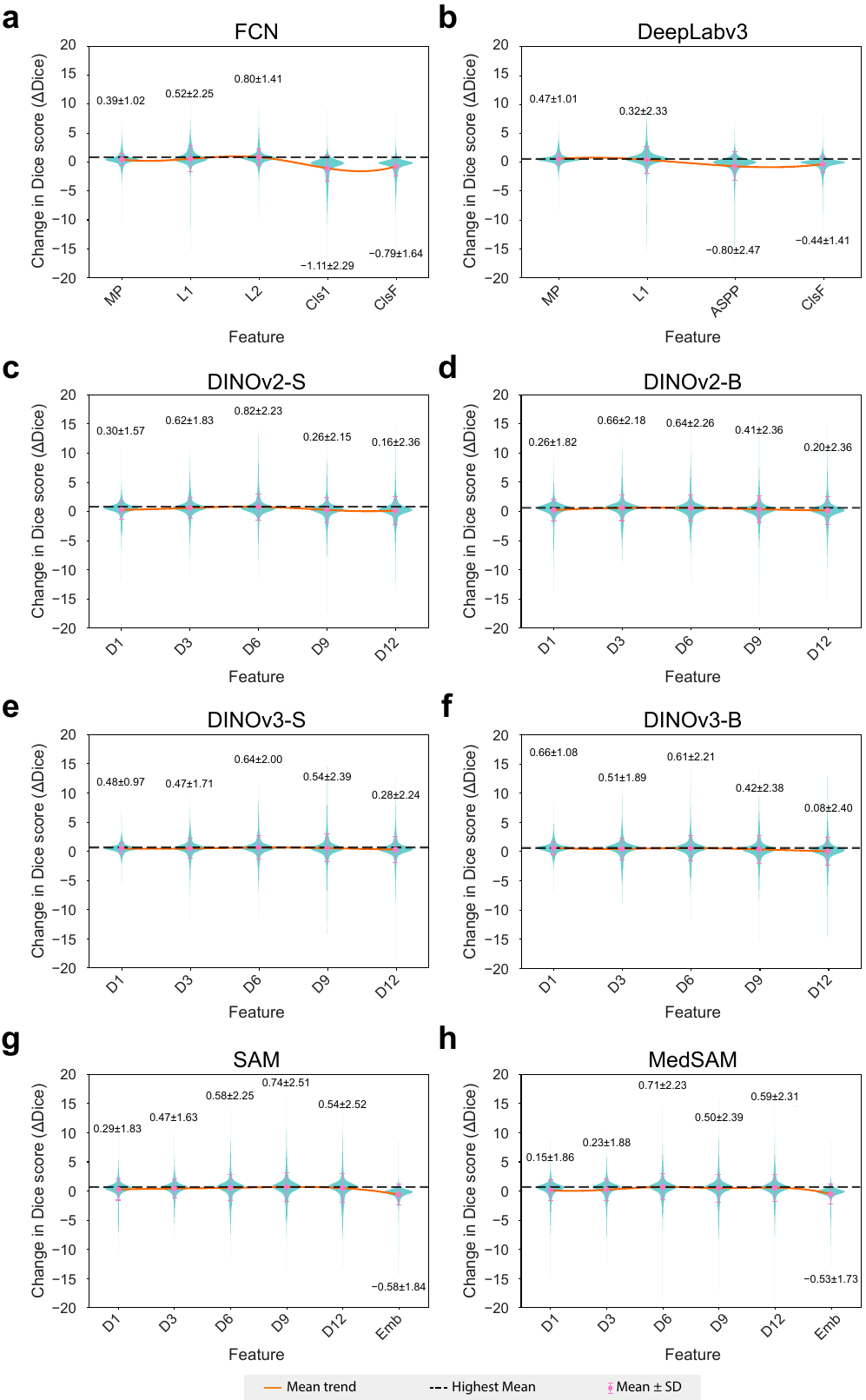}
    \caption{\textbf{Intermediate features are more effective than the deepest features.} For each violin plot, the number of data points is \textbf{n = 6,666}, corresponding to the total number of test images across 24 sub-tasks: 930 from Fluo-N2DH-GOWT1, 50 from ACDC, 400 from OASIS-1, and 3 orientations × (870 from MindBoggle-101, 8 from Lung-CT, 90 from Abdomen-CT, 306 from BCV-CT, 210 from CHAOS-MR, 270 from BCV–CHAOS, and 8 from Abdomen-CT--MR).}
    \label{supp:fig:depth}
\end{figure}

We examined how feature depth affects registration performance by measuring per test-pair \(\Delta\)Dice, defined as
\[
\Delta\mathrm{Dice}
=
\mathrm{Dice}_{\text{EP w/ F}}
-
\mathrm{Dice}_{\text{EP w/o F}},
\]
across eight encoder families and 6,666 test-pairs (Fig.~\ref{supp:fig:depth}). Specifically, we evaluated five FCN features, four DeepLabv3 features, five DINO-family features, and six SAM or MedSAM features. Across the ten registration tasks and 24 subtasks, the largest Dice gains were generally obtained from shallow or intermediate features, whereas the deepest features were rarely optimal and, in several cases, reduced performance.

For example, FCN features peaked at L2
($\Delta\mathrm{Dice}$=0.80$\pm$1.41) but became detrimental at Cls1 and ClsF
((-1.11$\pm$2.29) and (-0.79$\pm$1.64), respectively). A similar pattern was observed for MedSAM and SAM: intermediate features produced positive gains, whereas the deepest features reduced accuracy
(MedSAM-Emb: (-0.53$\pm$1.73); SAM-Emb: (-0.58$\pm$1.84)). DINOv2 and DINOv3 also favoured early-to-middle depths rather than their deepest representations. DINOv2-Small peaked at D6 (0.82$\pm$2.23) and  DINOv2-Base peaked at D3  (0.66$\pm$2.18), respectively, while DINOv3-Small peaked at D6  (0.64$\pm$2.00) and DINOv3-Base peaked at D1 (0.66$\pm$1.08). DeepLabv3 further supported this trend: shallow features yielded positive gains, whereas deeper features decreased performance
(ASPP: (-0.80$\pm$2.47); ClsF: (-0.44$\pm$1.41)).

These findings indicate that high-level semantic or classifier-proximal representations are not necessarily optimal for dense correspondence. Instead, mid-level features that preserve local structure and spatial detail appear better suited for medical image registration.

\begin{table}[hbt!]
\centering
\setlength{\tabcolsep}{1.0pt}
\caption{Best-performing \textbf{CNN} vs \textbf{ViT} feature across 10 tasks. Values are mean$\pm$s.d. over test cases. Metrics: Dice (\%), HD95, ASSD, $|J|_{ < 0}\%$, and std($J$). HD95 and ASSD are reported in voxels for BCV-CT, CHAOS-MR, and BCV--CHAOS and in millimeters for all other datasets.}
\label{tab:supp_cnn_vit}
\begin{tabular}{lcccccc}
\toprule
Feature & Dice & HD95 & ASSD  & $|J|_{ < 0}\%$ & std($J$) \\
\midrule
\multicolumn{6}{l}{\textbf{1. Fluo-N2DH-GOWT1}} \\
\midrule
FCN-L1 & 94.31 $\pm$ 1.98 & - & - & 1.62 $\pm$ 0.95 & 1.09 $\pm$ 0.70 \\
DINOv3-S-D12 & 94.07 $\pm$ 1.96 & - & - & 1.83 $\pm$ 1.30 & 1.60 $\pm$ 1.26 \\
\addlinespace[0.3em]
\multicolumn{6}{l}{\textbf{2. ACDC}} \\
\midrule
DeepLabv3-L1 & 86.70 $\pm$ 4.37 & 4.42 $\pm$ 1.76 & 1.43 $\pm$ 0.57 & 0.31 $\pm$ 0.35 & 0.22 $\pm$ 0.05 \\
SAM-D1 & 86.82 $\pm$ 4.30 & 4.38 $\pm$ 1.84 & 1.44 $\pm$ 0.60 & 0.29 $\pm$ 0.33 & 0.21 $\pm$ 0.05 \\
\addlinespace[0.3em]
\multicolumn{6}{l}{\textbf{3. OASIS-1}} \\
\bottomrule
\end{tabular}
\end{table}

\begin{table}[hbt!]
\centering
\setlength{\tabcolsep}{1.0pt}
\caption{Best-performing \textbf{SAM} vs \textbf{MedSAM} feature across 10 tasks.  Values are mean$\pm$s.d. over test cases. Metrics: Dice (\%), HD95, ASSD, $|J|_{ < 0}\%$, and std($J$). HD95 and ASSD are reported in voxels for BCV-CT, CHAOS-MR, and BCV--CHAOS and in millimeters for all other datasets.}
\label{tab:supp_sam_medsam}
\begin{tabular}{lcccccc}
\toprule
Feature & Dice & HD95 & ASSD  & $|J|_{ < 0}\%$ & std($J$) \\
\midrule
\multicolumn{6}{l}{\textbf{1. Fluo-N2DH-GOWT1}} \\
\midrule
SAM-D3 & 93.35 $\pm$ 2.99 & - & - & 2.24 $\pm$ 1.70 & 1.51 $\pm$ 0.99 \\
MedSAM-D9 & 93.40 $\pm$ 2.68 & - & - & 2.25 $\pm$ 1.78 & 1.59 $\pm$ 0.97 \\
\addlinespace[0.3em]
\multicolumn{6}{l}{\textbf{2. ACDC}} \\
\midrule
SAM-D1 & 86.82 $\pm$ 4.30 & 4.38 $\pm$ 1.84 & 1.44 $\pm$ 0.60 & 0.29 $\pm$ 0.33 & 0.21 $\pm$ 0.05 \\
MedSAM-Emb & 86.37 $\pm$ 4.87 & 4.59 $\pm$ 1.80 & 1.51 $\pm$ 0.68 & 0.28 $\pm$ 0.29 & 0.21 $\pm$ 0.05 \\
\addlinespace[0.3em]
\multicolumn{6}{l}{\textbf{3. OASIS-1}} \\
\bottomrule
\end{tabular}
\end{table}

\begin{table}[hbt!]
\centering
\setlength{\tabcolsep}{2.0pt}
\caption{ \textbf{ResNet18} vs \textbf{ResNet152} feature across 10 tasks.  Values are mean$\pm$s.d. over test cases. Metrics: Dice (\%), HD95, ASSD, $|J|_{ < 0}\%$, and std($J$). HD95 and ASSD are reported in voxels for BCV-CT, CHAOS-MR, and BCV--CHAOS and in millimeters for all other datasets.}
\label{tab:supp:resnet18_vs_152}
\begin{tabular}{lcccccc}
\toprule
Feature & Dice & HD95 & ASSD  & $|J|_{ < 0}\%$ & std($J$) \\
\midrule
\multicolumn{6}{l}{\textbf{1. Fluo-N2DH-GOWT1}} \\
\midrule
ResNet18 & 93.48 $\pm$ 2.64 & - & - & 1.69 $\pm$ 0.88 & 1.06 $\pm$ 0.51 \\
ResNet152 & 93.86 $\pm$ 2.24 & - & - & 1.00 $\pm$ 0.59 & 0.60 $\pm$ 0.33 \\
\addlinespace[0.3em]
\multicolumn{6}{l}{\textbf{2. ACDC}} \\
\midrule
ResNet18 & 84.85 $\pm$ 6.09 & 5.04 $\pm$ 2.25 & 1.67 $\pm$ 0.85 & 0.37 $\pm$ 0.37 & 0.24 $\pm$ 0.06 \\
ResNet152 & 83.99 $\pm$ 6.18 & 5.27 $\pm$ 2.29 & 1.77 $\pm$ 0.87 & 0.39 $\pm$ 0.38 & 0.24 $\pm$ 0.06 \\
\addlinespace[0.3em]
\multicolumn{6}{l}{\textbf{3. OASIS-1}} \\
\bottomrule
\end{tabular}
\end{table}

\begin{table}[hbt!]
\centering
\setlength{\tabcolsep}{2.0pt}
\caption{Best-performing \textbf{DINOv2-Small (DINOv2-S)} vs \textbf{DINOv2-Base (DINOv2-B)} feature across 10 tasks.  Values are mean$\pm$s.d. over test cases. Metrics: Dice (\%), HD95, ASSD, $|J|_{ < 0}\%$, and std($J$). HD95 and ASSD are reported in voxels for BCV-CT, CHAOS-MR, and BCV--CHAOS and in millimeters for all other datasets.}
\label{tab:dinov2_s_vs_b}
\begin{tabular}{lcccccc}
\toprule
Feature & Dice & HD95 & ASSD  & $|J|_{ < 0}\%$ & std($J$) \\
\midrule
\multicolumn{6}{l}{\textbf{1. Fluo-N2DH-GOWT1}} \\
\midrule
DINOv2-S-D6  & 93.06 $\pm$ 3.35 & - & - & 0.50 $\pm$ 0.40 & 0.40 $\pm$ 0.21 \\
DINOv2-B-D12 & 93.40 $\pm$ 2.74 & - & - & 0.69 $\pm$ 0.53 & 0.60 $\pm$ 0.40 \\
\addlinespace[0.3em]
\multicolumn{6}{l}{\textbf{2. ACDC}} \\
\midrule
DINOv2-S-D1 & 85.94 $\pm$ 4.73 & 4.97 $\pm$ 2.44 & 1.57 $\pm$ 0.67 & 0.34 $\pm$ 0.36 & 0.22 $\pm$ 0.05 \\
DINOv2-B-D1 & 85.96 $\pm$ 4.24 & 4.94 $\pm$ 2.13 & 1.59 $\pm$ 0.67 & 0.31 $\pm$ 0.31 & 0.21 $\pm$ 0.05 \\
\addlinespace[0.3em]
\multicolumn{6}{l}{\textbf{3. OASIS-1}} \\
\bottomrule
\end{tabular}
\end{table}

\clearpage
\begin{table}[hbt!]
\centering
\setlength{\tabcolsep}{2.0pt}
\caption{Best-performing \textbf{DINOv3-Small (DINOv3-S)} vs \textbf{DINOv3-Base (DINOv3-B)} feature across 10 tasks.  Values are mean$\pm$s.d. over test cases. Metrics: Dice (\%), HD95, ASSD, $|J|_{ < 0}\%$, and std($J$). HD95 and ASSD are reported in voxels for BCV-CT, CHAOS-MR, and BCV--CHAOS and in millimeters for all other datasets.}
\label{tab:supp_dinov3_s_b}
\begin{tabular}{lcccccc}
\toprule
Feature & Dice & HD95 & ASSD & $|J|_{ < 0}\%$ & std($J$) \\
\midrule
\multicolumn{6}{l}{\textbf{1. Fluo-N2DH-GOWT1}} \\
\midrule
DINOv3-S-D12 & 94.07 $\pm$ 1.96 & - & - & 1.83 $\pm$ 1.30 & 1.60 $\pm$ 1.26 \\
DINOv3-B-D1  & 93.58 $\pm$ 2.89 & - & - & 1.70 $\pm$ 1.25 & 1.38 $\pm$ 0.98 \\
\addlinespace[0.3em]
\multicolumn{6}{l}{\textbf{2. ACDC}} \\
\midrule
DINOv3-S-D1 & 86.07 $\pm$ 4.64 & 4.61 $\pm$ 1.94 & 1.52 $\pm$ 0.67 & 0.30 $\pm$ 0.36 & 0.22 $\pm$ 0.05 \\
DINOv3-B-D1 & 85.61 $\pm$ 5.03 & 4.85 $\pm$ 2.05 & 1.59 $\pm$ 0.72 & 0.34 $\pm$ 0.34 & 0.22 $\pm$ 0.05 \\
\addlinespace[0.3em]
\multicolumn{6}{l}{\textbf{3. OASIS-1}} \\
\bottomrule
\end{tabular}
\end{table}

\newpage
\begin{table}[hbt!]
\centering
\setlength{\tabcolsep}{2.0pt}
\caption{Best-performing \textbf{DINOv2-Small (DINOv2-S)} vs \textbf{DINOv3-Small (DINOv3-S)} feature across 10 tasks. Values are mean$\pm$s.d. over test cases. Metrics: Dice (\%), HD95, ASSD, $|J|_{ < 0}\%$, and std($J$). HD95 and ASSD are reported in voxels for BCV-CT, CHAOS-MR, and BCV--CHAOS and in millimeters for all other datasets.}
\label{tab:dinov2s_vs_dinov3s}
\begin{tabular}{lcccccc}
\toprule
Feature & Dice & HD95 & ASSD  & $|J|_{ < 0}\%$ & std($J$) \\
\midrule
\multicolumn{6}{l}{\textbf{1. Fluo-N2DH-GOWT1}} \\
\midrule
DINOv2-S-D6  & 93.06 $\pm$ 3.35 & - & - & 0.50 $\pm$ 0.40 & 0.40 $\pm$ 0.21 \\
DINOv3-S-D12 & 94.07 $\pm$ 1.96 & - & - & 1.83 $\pm$ 1.30 & 1.60 $\pm$ 1.26 \\
\addlinespace[0.3em]
\multicolumn{6}{l}{\textbf{2. ACDC}} \\
\midrule
DINOv2-S-D1 & 85.94 $\pm$ 4.73 & 4.97 $\pm$ 2.44 & 1.57 $\pm$ 0.67 & 0.34 $\pm$ 0.36 & 0.22 $\pm$ 0.05 \\
DINOv3-S-D1 & 86.07 $\pm$ 4.64 & 4.61 $\pm$ 1.94 & 1.52 $\pm$ 0.67 & 0.30 $\pm$ 0.36 & 0.22 $\pm$ 0.05 \\
\addlinespace[0.3em]
\multicolumn{6}{l}{\textbf{3. OASIS-1}} \\
\bottomrule
\end{tabular}
\end{table}

\clearpage
\begin{table}[hbt!]
\centering
\setlength{\tabcolsep}{2.0pt}
\caption{Best-performing \textbf{DINOv2-Base (DINOv2-B)} vs \textbf{DINOv3-Base (DINOv3-B)} feature across 10 tasks. Values are mean$\pm$s.d. over test cases. Metrics: Dice (\%), HD95, ASSD, $|J|_{ < 0}\%$, and std($J$). HD95 and ASSD are reported in voxels for BCV-CT, CHAOS-MR, and BCV--CHAOS and in millimeters for all other datasets.}
\label{tab:dinov2b_vs_dinov3b}
\begin{tabular}{lcccccc}
\toprule
Feature & Dice & HD95 & ASSD & $|J|_{ < 0}\%$ & std($J$) \\
\midrule
\multicolumn{6}{l}{\textbf{1. Fluo-N2DH-GOWT1}} \\
\midrule
DINOv2-B-D12 & 93.40 $\pm$ 2.74 & - & - & 0.69 $\pm$ 0.53 & 0.60 $\pm$ 0.40 \\
DINOv3-B-D1  & 93.58 $\pm$ 2.89 & - & - & 1.70 $\pm$ 1.25 & 1.38 $\pm$ 0.98 \\
\addlinespace[0.3em]
\multicolumn{6}{l}{\textbf{2. ACDC}} \\
\midrule
DINOv2-B-D1 & 85.96 $\pm$ 4.24 & 4.94 $\pm$ 2.13 & 1.59 $\pm$ 0.67 & 0.31 $\pm$ 0.31 & 0.21 $\pm$ 0.05 \\
DINOv3-B-D1 & 85.61 $\pm$ 5.03 & 4.85 $\pm$ 2.05 & 1.59 $\pm$ 0.72 & 0.34 $\pm$ 0.34 & 0.22 $\pm$ 0.05 \\
\addlinespace[0.3em]
\multicolumn{6}{l}{\textbf{3. OASIS-1}} \\
\bottomrule
\end{tabular}
\end{table}

\newpage
\section{Relationship between mutual information and registration performance}
\label{sec:supp:correlation}
We investigated whether mutual information (MI) between the original image and candidate feature representations could predict registration performance. The analysis included 24 sub-task combinations derived from three 2D registration tasks and seven 3D registration tasks evaluated using sagittal, coronal and axial feature extraction. For each task-view combination, we computed MI between each feature representation and the corresponding input image, and assessed the relationship between MI and Dice using simple linear regression across 47 candidate features.

When all 47 features were included, the association between MI and Dice was generally weak. Only Abdomen-CT showed high correlations across all three views, with $R=0.6803$ for sagittal extraction, $R=0.6982$ for coronal extraction, and $R=0.7127$ for axial extraction. In most other task-view combinations, MI explained only a small proportion of the variation in registration accuracy, indicating that preserving information about the input image is not, by itself, sufficient to identify the most useful feature for registration.

We next repeated the analysis after restricting each task-view combination to the five features with the highest MI and the five features with the lowest MI. This contrastive analysis strengthened the association between MI and Dice in several settings. 21 of 24 task-view combinations showed absolute $R$ values above 0.5, and 10 reached $R$ values larger than 0.8. 

Lung-CT was a notable exception to the overall trend. In this task, lower-MI features were associated with higher Dice, suggesting that features closely preserving the original image appearance are not always optimal for registration. One possible explanation is that, for large and high-contrast anatomical structures such as the lungs, features that suppress appearance-level information while preserving shape-relevant cues may provide a more useful registration signal. This observation further supports the conclusion that MI should not be interpreted as a universal proxy for feature quality.

Overall, MI provides a useful descriptive measure of feature informativeness, and may help identify clearly uninformative representations in some settings. However, it does not reliably predict the best-performing feature across tasks. Feature selection in \textit{Easy Peeler} should therefore remain validation driven, with MI used only as a secondary diagnostic rather than as a standalone criterion.

\newpage
\section{The Architecture Design of Easy Peeler}
\label{sec:supp:method}

\subsection{Detailed architecture of the learnable transition-decoder}

\paragraph{Feature-map resolution.}
Let \(\Omega_{\mathrm{feat}}\) denote the spatial resolution of the moving and fixed feature maps at the coarsest scale of the transition-decoder. For 2D registration,
\[
\Omega_{\mathrm{feat}} = 32\times32.
\]

For a 3D orientation branch
\[
o \in \{\mathrm{sagittal},\mathrm{coronal},\mathrm{axial}\},
\]
let \(N_o\) denote the input-volume size along the corresponding orientation and define
\[
n_o = \lceil N_o/3\rceil.
\]
The feature-map resolution is
\[
\Omega_{\mathrm{feat}}
=
\begin{cases}
n_o\times32\times32,
& o=\mathrm{sagittal},\\[2pt]
32\times n_o\times32,
& o=\mathrm{coronal},\\[2pt]
32\times32\times n_o,
& o=\mathrm{axial}.
\end{cases}
\]

\paragraph{Multi-scale intensity and structure cues.}
Resolution-matched image-intensity and structure cues are constructed at three scales, indexed by \(\ell=0,1,2\) from coarse to fine:
\[
I_M^{(\ell)}
=
\Pi_o^{(\ell)}(I_M),
\qquad
I_F^{(\ell)}
=
\Pi_o^{(\ell)}(I_F),
\]
where \(\Pi_o^{(\ell)}(\cdot)\) denotes adaptive average pooling to the target resolution associated with scale \(\ell\) and orientation \(o\).

For 3D registration, the pooling targets are
\[
\begin{aligned}
\Pi_o^{(0)} &:\
\begin{cases}
(n_o,32,32), & \mathrm{sagittal},\\
(32,n_o,32), & \mathrm{coronal},\\
(32,32,n_o), & \mathrm{axial},
\end{cases}\\[6pt]
\Pi_o^{(1)} &:\
\begin{cases}
(n_o,64,64), & \mathrm{sagittal},\\
(64,n_o,64), & \mathrm{coronal},\\
(64,64,n_o), & \mathrm{axial},
\end{cases}\\[6pt]
\Pi_o^{(2)} &:\
\begin{cases}
(2n_o,128,128), & \mathrm{sagittal},\\
(128,2n_o,128), & \mathrm{coronal},\\
(128,128,2n_o), & \mathrm{axial}.
\end{cases}
\end{aligned}
\]

The corresponding structural representations are
\[
S_M^{(\ell)}
=
\operatorname{Sobel}\!\left(I_M^{(\ell)}\right),
\qquad
S_F^{(\ell)}
=
\operatorname{Sobel}\!\left(I_F^{(\ell)}\right).
\]
For 3D registration, fixed Sobel kernels are applied independently along the three spatial axes. The responses are squared and summed, followed by a square root, to produce a single-channel gradient-magnitude representation. For 2D registration, the corresponding two-dimensional Sobel operation is used.

\paragraph{Prompting-module architecture.}
After feature filtering, each filtered representation is independently augmented with an intensity or structure cue. Each prompting module is implemented as a bottleneck consisting of
\[
1\times1\times1
\rightarrow
\mathrm{PReLU}
\rightarrow
3\times3\times3
\rightarrow
\mathrm{PReLU}
\rightarrow
1\times1\times1
\rightarrow
\mathrm{PReLU}
\]
for 3D registration, with the corresponding \(1\times1\), \(3\times3\), and \(1\times1\) operations used for 2D registration.


\paragraph{Gated feature fusion.}
For both the intensity- and structure-prompted representations, moving and fixed features are concatenated along the channel dimension and modulated by a learned depthwise gate. The intensity branch is
\[
\begin{aligned}
\tilde{f}_{\mathrm{ins}}
&=
[g_M^{\mathrm{ins}},g_F^{\mathrm{ins}}],\\
a_{\mathrm{ins}}
&=
\sigma\!\left(
\mathcal{A}_{\mathrm{ins}}
(\tilde{f}_{\mathrm{ins}})
\right),\\
f_{\mathrm{ins}}
&=
\tilde{f}_{\mathrm{ins}}\odot a_{\mathrm{ins}},
\end{aligned}
\]
and the structure branch is
\[
\begin{aligned}
\tilde{f}_{\mathrm{str}}
&=
[g_M^{\mathrm{str}},g_F^{\mathrm{str}}],\\
a_{\mathrm{str}}
&=
\sigma\!\left(
\mathcal{A}_{\mathrm{str}}
(\tilde{f}_{\mathrm{str}})
\right),\\
f_{\mathrm{str}}
&=
\tilde{f}_{\mathrm{str}}\odot a_{\mathrm{str}}.
\end{aligned}
\]

The two cue-specific representations are concatenated and gated once more:
\[
\begin{aligned}
\tilde{f}
&=
[f_{\mathrm{ins}},f_{\mathrm{str}}],\\
a
&=
\sigma\!\left(
\mathcal{A}_{\mathrm{cue}}(\tilde{f})
\right),\\
f
&=
\tilde{f}\odot a.
\end{aligned}
\]
Each \(\mathcal{A}\) consists of a depthwise convolution, with the number of groups equal to the number of input channels, followed by sigmoid activation.

\paragraph{Multi-scale decoder architecture.}
The decoder contains three scales (\(N=3\)), with three convolutional blocks per scale (\(K=3\)). A residual encoder block is defined as
\[
\operatorname{ResEnc}(x)
=
B(x)+x,
\]
and a skip-add encoder block as
\[
\operatorname{SkipEnc}(x,s)
=
B(x)+s,
\]
where \(B(\cdot)\) is a bottleneck transformation comprising
\[
1\times1\times1
\rightarrow
3\times3\times3
\rightarrow
1\times1\times1
\]
convolutions with PReLU activations for 3D registration and the corresponding 2D operations for 2D registration.

At the coarsest scale,
\[
\begin{aligned}
h_0^{(1)}
&=
\operatorname{ResEnc}_{0,1}(f),\\
h_0^{(2)}
&=
\operatorname{ResEnc}_{0,2}(h_0^{(1)}),\\
h_0^{(3)}
&=
\operatorname{ResEnc}_{0,3}(h_0^{(2)}).
\end{aligned}
\]

\paragraph{Orientation-specific upsampling.}
Transition from scale \(\ell=0\) to \(\ell=1\) is performed using a transposed convolution \(\mathcal{U}_{0\rightarrow1}^{(o)}\). For the 3D orientation branches, the kernel and stride are selected such that the two in-plane dimensions are doubled while the orientation dimension is preserved:
\[
\begin{array}{c|c|c}
\text{Branch} & \text{Kernel} & \text{Stride}\\
\hline
\mathrm{axial}
& (2,2,1)
& (2,2,1)\\
\mathrm{coronal}
& (2,1,2)
& (2,1,2)\\
\mathrm{sagittal}
& (1,2,2)
& (1,2,2)
\end{array}
\]

Thus,
\[
\tilde{h}_1
=
\mathcal{U}_{0\rightarrow1}^{(o)}
\!\left(h_0^{(3)}\right).
\]

At scale \(\ell=1\), the upsampled decoder representation is concatenated with the moving and fixed intensity and structure cues:
\[
c_1
=
\left[
\tilde{h}_1,
I_M^{(1)},
I_F^{(1)},
S_M^{(1)},
S_F^{(1)}
\right].
\]
The concatenated representation is modulated by a depthwise gating mask,
\[
\hat{c}_1
=
c_1\odot
\sigma\!\left(\mathcal{G}_1(c_1)\right).
\]
The first block incorporates the upsampled representation through a skip-add connection, followed by two residual blocks:
\[
\begin{aligned}
h_1^{(1)}
&=
\operatorname{SkipEnc}_{1,1}
(\hat{c}_1,\tilde{h}_1),\\
h_1^{(2)}
&=
\operatorname{ResEnc}_{1,2}(h_1^{(1)}),\\
h_1^{(3)}
&=
\operatorname{ResEnc}_{1,3}(h_1^{(2)}).
\end{aligned}
\]

The same procedure is used to reach the finest decoder scale:
\[
\tilde{h}_2
=
\mathcal{U}_{1\rightarrow2}^{(o)}
\!\left(h_1^{(3)}\right),
\]
followed by cue concatenation,
\[
c_2
=
\left[
\tilde{h}_2,
I_M^{(2)},
I_F^{(2)},
S_M^{(2)},
S_F^{(2)}
\right],
\]
depthwise gating,
\[
\hat{c}_2
=
c_2\odot
\sigma\!\left(\mathcal{G}_2(c_2)\right),
\]
and residual processing,
\[
\begin{aligned}
h_2^{(1)}
&=
\operatorname{SkipEnc}_{2,1}
(\hat{c}_2,\tilde{h}_2),\\
h_2^{(2)}
&=
\operatorname{ResEnc}_{2,2}(h_2^{(1)}),\\
h_2^{(3)}
&=
\operatorname{ResEnc}_{2,3}(h_2^{(2)}).
\end{aligned}
\]

With \(N=3\) scales, two transposed-convolution upsampling operations are therefore used, corresponding to the transitions \(0\rightarrow1\) and \(1\rightarrow2\). More generally, an \(N\)-scale decoder contains \(N-1\) inter-scale upsampling operations.

The final decoder representation is projected using
\[
z
=
\mathcal{O}_{\omega}(h_2^{(3)}),
\]
where \(\mathcal{O}_{\omega}\) is a \(3\times3\times3\) convolution for 3D registration and a \(3\times3\) convolution for 2D registration. The output has two channels in 2D and three channels in 3D:
\[
z \in \mathbb{R}^{2\times H'\times W'}
\]
or
\[
z \in \mathbb{R}^{3\times H'\times W'\times D'},
\]
respectively.

\paragraph{Fourier decoding.}
The representation \(z\) is produced at the finest decoder scale, corresponding to \(128\times128\) for 2D registration and to the orientation-dependent \(\Pi_o^{(2)}\) resolution for 3D registration. It is mapped to the target image resolution using resolution-agnostic Fourier upsampling~\cite{jia2023fourier}:
\[
\begin{aligned}
\text{2D:}\quad
u
&= \mathcal{F}_{\uparrow}
\!\left(
z,
\Omega_{\mathrm{img}}
\right)=
\mathcal{F}^{-1}
\left[
\operatorname{ZeroPad}
\left(
\mathcal{F}(z),H\times W
\right)
\right],\\
\text{3D:}\quad
u
&= \mathcal{F}_{\uparrow}
\!\left(
z,
\Omega_{\mathrm{img}}
\right)=
\mathcal{F}^{-1}
\left[
\operatorname{ZeroPad}
\left(
\mathcal{F}(z),H\times W\times D
\right)
\right].
\end{aligned}
\]
The resulting full-resolution displacement fields have dimensions
\[
u \in \mathbb{R}^{2\times H\times W}
\]
and
\[
u \in \mathbb{R}^{3\times H\times W\times D}
\]
for 2D and 3D registration, respectively. This decoding step contains no learned parameters.

Unless otherwise stated, all experiments used three decoder scales (\(N=3\)) and three convolutional blocks per scale (\(K=3\)).

\newpage
\section{Extending Easy Peeler to Pyramid Registration}
\label{sec:supp:pyramid_extension}

We extend Easy Peeler to pyramid registration by introducing explicit coarse-to-fine displacement refinement across the three decoder scales.
The feature representations, intensity and structure cues, prompting and fusion modules, residual decoder blocks, and orientation-specific feature upsampling follow the architecture defined in Section~\ref{sec:supp:method}. The key difference is that the pyramidal variant predicts a displacement field at each decoder scale. The coarsest-scale prediction is propagated to the next finer scale and used to warp the moving intensity and structure cues; the decoder then predicts a residual displacement that is spatially composed with the propagated coarse estimate. This procedure is repeated until the finest decoder scale, after which the composed displacement is Fourier-upsampled to the original
image resolution.

\subsection{Learnable pyramidal transition-decoder}

\paragraph{Multi-scale representations.}
We use the same three-scale intensity and structure representations
defined in Section~\ref{sec:supp:method},
\[
I_M^{(\ell)},\quad I_F^{(\ell)},\quad
S_M^{(\ell)},\quad S_F^{(\ell)},
\qquad \ell\in\{0,1,2\},
\]
with scales indexed from coarse to fine. Their resolutions are determined
by the orientation-dependent pooling operators
$\Pi_o^{(\ell)}$ defined above.

\paragraph{Coarsest-scale displacement prediction.}
At the coarsest scale, the input moving and fixed maps are processed by the feature-filtering, prompting, fusion, and residual-decoding modules defined in Section~\ref{sec:supp:method},
yielding the coarsest decoder representation $h_0^{(3)}$.

Unlike the standard Easy Peeler decoder, which produces a displacement
only after the finest decoder scale, the pyramidal variant attaches a
displacement head to the coarsest representation:
\[
\psi_0
=
\mathcal{O}_{\omega}^{(0)}
\!\left(h_0^{(3)}\right),
\]
where $\mathcal{O}_{\omega}^{(0)}$ is a
$3\times3\times3$ convolution for 3D registration and the corresponding
$3\times3$ convolution for 2D registration. The initial displacement
estimate is therefore
\[
\phi_0 = \psi_0.
\]

\paragraph{Cross-scale displacement propagation and warping.}
To refine the displacement at the next finer scale, the current composed
field $\phi_{\ell-1}$ is first upsampled to the resolution of scale
$\ell$ using the parameter-free Fourier zero-padding operation introduced
above:
\[
\tilde{\phi}_{\ell-1}
=
\mathcal{F}_{\uparrow}
\!\left(
\phi_{\ell-1},
\Pi_o^{(\ell)}
\right),
\qquad \ell\in\{1,2\}.
\]

The propagated displacement is then used to warp the moving intensity
and structure representations at the current scale:
\[
\begin{aligned}
I_M^{(\ell),\mathrm{w}}
&=
\operatorname{Warp}
\!\left(
I_M^{(\ell)},
\tilde{\phi}_{\ell-1}
\right),\\
S_M^{(\ell),\mathrm{w}}
&=
\operatorname{Warp}
\!\left(
S_M^{(\ell)},
\tilde{\phi}_{\ell-1}
\right).
\end{aligned}
\]
The corresponding fixed representations
$I_F^{(\ell)}$ and $S_F^{(\ell)}$ remain unchanged.

\paragraph{Finer-scale residual prediction.}
At each finer scale $\ell\in\{1,2\}$, decoder features are propagated
from the preceding scale using the orientation-specific transposed
convolution defined in Section~\ref{sec:supp:method}:
\[
\tilde{h}_{\ell}
=
\mathcal{U}_{\ell-1\rightarrow\ell}^{(o)}
\!\left(
h_{\ell-1}^{(3)}
\right).
\]

The decoder input follows the standard transition-decoder construction,
except that the moving intensity and structure cues are replaced by their
warped counterparts:
\[
c_{\ell}
=
\left[
\tilde{h}_{\ell},
I_M^{(\ell),\mathrm{w}},
I_F^{(\ell)},
S_M^{(\ell),\mathrm{w}},
S_F^{(\ell)}
\right].
\]
As in the standard decoder, this representation is modulated by a
depthwise gating mask,
\[
\hat{c}_{\ell}
=
c_{\ell}
\odot
\sigma\!\left(
\mathcal{G}_{\ell}(c_{\ell})
\right),
\]
and processed by one skip-add encoder block followed by two residual
blocks:
\[
\begin{aligned}
h_{\ell}^{(1)}
&=
\operatorname{SkipEnc}_{\ell,1}
\!\left(
\hat{c}_{\ell},
\tilde{h}_{\ell}
\right),\\
h_{\ell}^{(2)}
&=
\operatorname{ResEnc}_{\ell,2}
\!\left(
h_{\ell}^{(1)}
\right),\\
h_{\ell}^{(3)}
&=
\operatorname{ResEnc}_{\ell,3}
\!\left(
h_{\ell}^{(2)}
\right).
\end{aligned}
\]

A scale-specific output head then predicts a residual displacement field,
\[
\psi_{\ell}
=
\mathcal{O}_{\omega}^{(\ell)}
\!\left(
h_{\ell}^{(3)}
\right),
\qquad \ell\in\{1,2\}.
\]

\paragraph{Displacement composition.}
The residual displacement predicted at scale $\ell$ is spatially
composed with the propagated displacement from the preceding scale:
\[
\phi_{\ell}
=
\tilde{\phi}_{\ell-1}
\circ
\psi_{\ell},
\qquad \ell\in\{1,2\},
\]
where composition is defined as
\[
(\phi\circ\psi)(\mathbf{p})
=
\psi(\mathbf{p})
+
\phi\!\left(
\mathbf{p}+\psi(\mathbf{p})
\right),
\]
with $\mathbf{p}$ denoting a spatial coordinate. Thus, finer-scale
residual updates are spatially composed with, rather than directly added
to, the propagated coarse displacement.

For the three-scale architecture used here, the complete refinement is
therefore
\[
\phi_0=\psi_0,
\qquad
\phi_1=
\mathcal{F}_{\uparrow}
\!\left(
\phi_0,\Pi_o^{(1)}
\right)
\circ\psi_1,
\qquad
\phi_2=
\mathcal{F}_{\uparrow}
\!\left(
\phi_1,\Pi_o^{(2)}
\right)
\circ\psi_2.
\]

\paragraph{Final displacement output.}
After refinement at the finest decoder scale, the composed displacement
$\phi_2$ is mapped to the original image resolution using the same
parameter-free Fourier upsampling operation:
\[
u
=
\mathcal{F}_{\uparrow}
\!\left(
\phi_2,
\Omega_{\mathrm{img}}
\right),
\]
where $\Omega_{\mathrm{img}}=H\times W$ for 2D registration and
$\Omega_{\mathrm{img}}=H\times W\times D$ for 3D registration.
Accordingly,
\[
u\in\mathbb{R}^{2\times H\times W}
\]
for 2D registration and
\[
u\in\mathbb{R}^{3\times H\times W\times D}
\]
for 3D registration.

\begin{figure}
    \centering
    \includegraphics[width=0.99\linewidth]{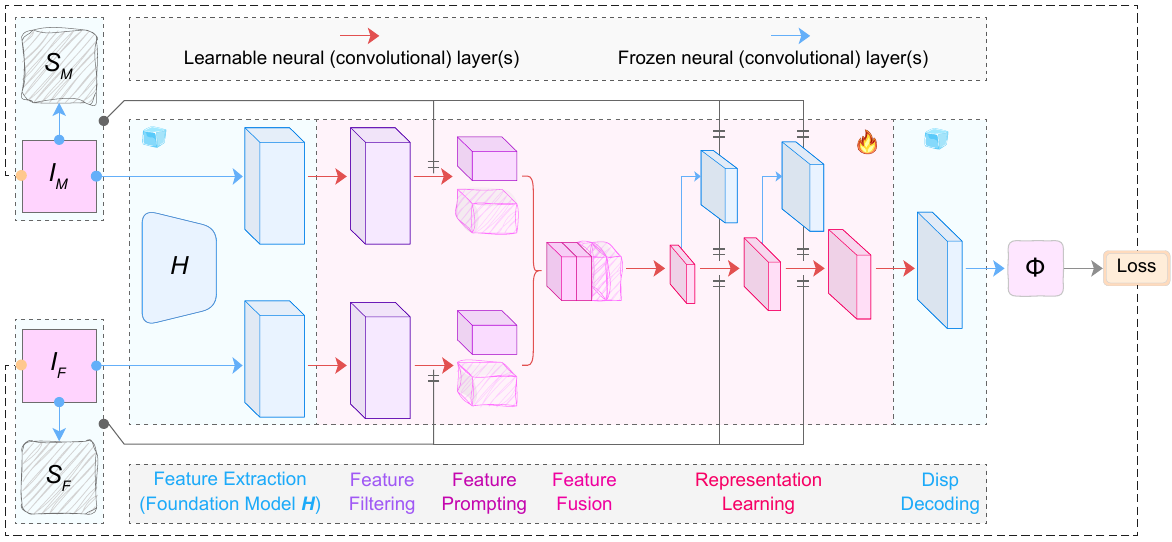}
    \caption{\textbf{Schematic Overview of Pyramid Easy Peeler.} Pyramid Easy Peeler follows the same overall design as the standard Easy Peeler architecture, while introducing coarse-to-fine displacement estimation and refinement across multiple scales. \label{supp:fig:pyramid_easypeeler}}
\end{figure}

All pyramid experiments use the same three decoder scales ($N=3$) and three convolutional blocks per scale ($K=3$) as the standard Easy Peeler architecture. A schematic overview of Pyramid Easy Peeler is shown in Fig.~\ref{supp:fig:pyramid_easypeeler}.

\subsection{Comparisons to Single-scale and Pyramid ANTs SyN}

All experiments were performed using the ANTs SyN implementation via the `ants.registration' function with the `SyNOnly' transformation model. To identify the best-performing ANTs SyN configuration~\cite{avants2008symmetric}, we conducted a systematic hyperparameter search over the number of registration iterations and the choice of similarity metric. We evaluated three similarity metrics commonly used in the ANTsPy package\footnote{\url{https://antspyx.readthedocs.io/en/latest/registration.html}}: `meansquares' (MSE), `local cross-correlation' (LCC), and `mattes'. For each metric, we explored a comprehensive range of iteration settings, organized into single-scale and multi-scale schemes. In the single-scale setting, each image pair was optimized at a single resolution level using 10, 20, 40, 80, 160, 320, 640, 1280, or 2560 iterations. In the multi-scale setting, we systematically evaluated two-, three-, four-, and five-level configurations, with each iteration sequence specifying the number of optimization iterations at successive resolution levels. For two levels, we tested $[20, 10]$, $[40, 20]$, $[80, 40]$, $[160, 80]$, $[320, 160]$, $[640, 320]$, $[640, 640]$, and $[1280, 1280]$; for three levels, $[40, 20, 10]$, $[80, 40, 20]$, $[160, 80, 40]$, $[320, 160, 80]$, $[640, 320, 160]$, $[640, 640, 640]$, and $[1280, 640, 640]$; for four levels, $[80, 40, 20, 10]$, $[160, 80, 40, 20]$, $[320, 160, 80, 40]$, $[640, 320, 160, 80]$, and $[640, 640, 640, 640]$; and for five levels, $[160, 80, 40, 20, 10]$, $[320, 160, 80, 40, 20]$, $[640, 320, 160, 80, 40]$, and $[1280, 640, 320, 160, 80]$. This grid search was performed independently for each similarity metric, and the best-performing configuration was selected based on the validation Dice score to ensure a consistent comparison across methods. 

Table~\ref{tab:supp_syn} provides dataset-wise comparisons of our methods with ANTs SyN under single- and multi-scale configurations.
\begin{enumerate}
    \item Fluo-N2DH-GOWT1. The best-performing single-scale ANTs SyN configuration uses MSE with 2,560 iterations, whereas the best-performing multi-scale configuration uses LCC with five resolution levels and an iteration schedule of $[320, 160, 80, 40, 20]$. Our methods achieve higher Dice scores than ANTs SyN in both settings. ANTs SyN, however, yields substantially lower $|J|_{<0}\%$ and $\mathrm{std}(J)$, indicating a trade-off between registration accuracy and deformation regularity. Comparing EP w/F with Pyramid EP w/F (highlighted in green), the latter achieves a lower Dice score but produces a more regular deformation field. In particular, single-scale EP w/F produces irregular deformations for some image pairs, leading to misaligned cells and undefined (NaN) HD95 and ASSD values. By contrast, Pyramid EP w/F successfully aligns all cells and yields valid HD95 and ASSD measurements, despite its lower overall Dice score.
    \item ACDC. The best-performing single-scale ANTs SyN configuration uses the MSE metric with 640 iterations, whereas the best multi-scale configuration uses a two-level MSE objective with an iteration schedule of $[320, 160]$. Our method outperforms ANTs SyN in the single-scale setting, whereas multi-scale ANTs SyN achieves a higher Dice score. To assess the potential benefit of test-time adaptation, we additionally perform instance-specific fine-tuning independently for each test pair while retaining the loss function and $\lambda$ selected on the validation set; these results are denoted by $^*$. With instance-specific fine-tuning, our method outperforms ANTs SyN across all three registration-accuracy metrics, achieving a higher Dice score together with lower HD95 and ASSD.
    \item OASIS-1. The best-performing single-scale ANTs SyN configuration uses the MSE metric with 1,280 iterations, whereas the best multi-scale configuration uses a three-level LCC objective with an iteration schedule of $[320, 160, 80]$. Our methods achieve higher Dice scores than ANTs SyN under both settings. ANTs SyN, however, yields substantially lower $|J|*{<0}\%$ and $\mathrm{std}(J)$, indicating fewer negative Jacobian determinants and lower spatial variability of the deformation field. This again reflects a trade-off between registration accuracy and deformation regularity. Similar to the results on Fluo-N2DH-GOWT1, Pyramid EP w/F achieves a lower Dice score than its single-scale counterpart but produces a more regular deformation field, as reflected by its lower $|J|*{<0}\%$ and $\mathrm{std}(J)$.
    \item MindBoggle-101. The best-performing single-scale ANTs SyN configuration uses the LCC metric with 640 iterations, whereas the best multi-scale configuration uses a four-level LCC objective with an iteration schedule of $[640, 640, 640, 640]$. Our methods achieve higher Dice scores than ANTs SyN under both single- and multi-scale settings. Pyramid ANTs SyN, however, yields substantially lower $|J|_{<0}\%$ and $\mathrm{std}(J)$, indicating fewer negative Jacobian determinants and lower spatial variability of the deformation field, and also achieves a lower HD95. Thus, although our methods provide higher overlap accuracy, Pyramid ANTs SyN produces more regular deformations and performs better in terms of HD95 on this dataset.
    \item Lung-CT. The best-performing single-scale ANTs SyN configuration uses the LCC metric with 2,560 iterations, whereas the best multi-scale configuration uses a three-level LCC objective with an iteration schedule of $[640, 320, 160]$. Our methods achieve higher Dice scores than ANTs SyN under both settings. Pyramid ANTs SyN, however, yields substantially lower $|J|_{<0}\%$ and $\mathrm{std}(J)$, indicating fewer negative Jacobian determinants and lower spatial variability of the deformation field, and also achieves a lower HD95. These results indicate a trade-off between overlap accuracy, boundary alignment, and deformation regularity on this dataset.
    \item Abdomen-CT. The best-performing single-scale ANTs SyN configuration uses the LCC metric with 2,560 iterations, whereas the best multi-scale configuration uses LCC with a five-level iteration schedule of $[1280, 640, 320, 160, 80]$. In the single-scale setting, EP w/FCN-L1 outperforms ANTs SyN by 9.92 percentage points in Dice. In the multi-scale setting, however, Pyramid ANTs SyN exceeds Pyramid EP by 5.44 percentage points in Dice and yields substantially lower $|J|_{<0}\%$ and $\mathrm{std}(J)$, indicating a more regular deformation field. With instance-specific optimization, Pyramid EP improves to a Dice score of 61.87\% while also reducing HD95 and ASSD, demonstrating the benefit of adapting the registration to individual test pairs.
    \item BCV-CT. The best-performing single-scale ANTs SyN configuration uses the LCC metric with 2,560 iterations, whereas the best multi-scale configuration uses a five-level LCC objective with an iteration schedule of $[1280, 640, 320, 160, 80]$. In the single-scale setting, EP w/SAM-D9 outperforms ANTs SyN by 7.15 percentage points in Dice. In the multi-scale setting, Pyramid ANTs SyN exceeds Pyramid EP by 2.85 percentage points in Dice and yields substantially lower $|J|_{<0}\%$ and $\mathrm{std}(J)$. With instance-specific optimization, Pyramid EP improves to a Dice score of 77.17\%, together with lower HD95 and ASSD.
    \item CHAOS-MR. The best-performing single-scale ANTs SyN configuration uses the MSE metric with 2,560 iterations, whereas the best multi-scale configuration uses a four-level LCC objective with an iteration schedule of $[640, 640, 640, 640]$. In the single-scale setting, EP w/MedSAM-D12 outperforms ANTs SyN by 11.18 percentage points in Dice. In the multi-scale setting, Pyramid ANTs SyN achieves a marginally higher Dice score than Pyramid EP, by 0.95 percentage points, while also yielding substantially lower $|J|_{<0}\%$ and $\mathrm{std}(J)$. With instance-specific optimization, Pyramid EP improves to a Dice score of 73.61\% and further reduces both HD95 and ASSD.
    \item BCV--CHAOS. The best-performing single-scale ANTs SyN configuration uses the Mattes mutual information metric with 2,560 iterations, whereas the best multi-scale configuration uses a three-level LCC objective with an iteration schedule of $[320, 160, 80]$. Our methods achieve higher Dice scores than ANTs SyN under both single- and multi-scale settings. Pyramid ANTs SyN, however, yields substantially lower $|J|_{<0}\%$ and $\mathrm{std}(J)$ and also achieves a lower HD95. Thus, the higher Dice overlap achieved by our methods is accompanied by reduced deformation regularity and, on this dataset, less favorable boundary alignment as measured by HD95.
    \item Abdomen-CT--MR. We use the ANTs SyN configurations selected on BCV--CHAOS: the single-scale configuration uses the Mattes mutual information metric with 2,560 iterations, whereas the multi-scale configuration uses a three-level LCC objective with an iteration schedule of $[320, 160, 80]$. Our methods achieve higher Dice scores than ANTs SyN under both settings. Pyramid ANTs SyN, however, yields substantially lower $|J|_{<0}\%$ and $\mathrm{std}(J)$ and also achieves a lower HD95. These results further demonstrate the recurring trade-off between overlap accuracy, boundary alignment, and deformation regularity across datasets.
\end{enumerate}

\begingroup
\footnotesize
\setlength{\tabcolsep}{2.5pt}
\renewcommand{\arraystretch}{1.05}
\setlength{\LTcapwidth}{\textwidth}

\begin{longtable}{@{}lccccc@{}}
\caption{\small Dataset-wise comparisons of our methods with ANTs SyN under
single- and pyramid configurations. Pyramid results are highlighted in
green. For selected datasets, we additionally report instance-specific
fine-tuning performed independently for each test pair, while retaining
the loss function and $\lambda$ selected on the validation set; these
results are denoted by $*$. Values are mean$\pm$s.d. over test cases.
Metrics: Dice (\%), HD95, ASSD, $|J|_{<0}$ (\%), and std($J$). HD95 and
ASSD are reported in voxels for BCV-CT, CHAOS-MR, and BCV--CHAOS, and in
millimeters for all other datasets.}
\label{tab:supp_syn}\\

\toprule
Method & Dice & HD95 & ASSD & $|J|_{<0}$ (\%) & std($J$) \\
\midrule
\endfirsthead

\multicolumn{6}{@{}l}{\textbf{Table \thetable\ continued}}\\
\toprule
Method & Dice & HD95 & ASSD & $|J|_{<0}$ (\%) & std($J$) \\
\midrule
\endhead

\midrule
\multicolumn{6}{r@{}}{Continued on next page}\\
\endfoot

\bottomrule
\endlastfoot

\multicolumn{6}{@{}l}{\textbf{1. Fluo-N2DH-GOWT1}} \\
\midrule

ANTs SyN
& 64.46 $\pm$ 21.94
& 6.43 $\pm$ 4.55
& 3.39 $\pm$ 2.90
& 0.02 $\pm$ 0.04
& 0.15 $\pm$ 0.05 \\

EP w/ FCN-L1
& 94.31 $\pm$ 1.98
& --
& --
& 1.62 $\pm$ 0.95
& 1.09 $\pm$ 0.70 \\

\rowcolor{SeaGreen3!30!}
ANTs SyN
& 84.34 $\pm$ 12.47
& 3.65 $\pm$ 3.04
& 1.74 $\pm$ 1.91
& 0.0
& 0.20 $\pm$ 0.07 \\

\rowcolor{SeaGreen3!30!}
EP w/ FCN-L1
& 93.28 $\pm$ 3.21
& 2.94 $\pm$ 1.15
& 0.59 $\pm$ 0.38
& 1.10 $\pm$ 0.78
& 0.60 $\pm$ 0.31 \\

\addlinespace[0.4em]
\multicolumn{6}{@{}l}{\textbf{2. ACDC}} \\
\midrule

ANTs SyN
& 86.68 $\pm$ 4.77
& 4.94 $\pm$ 2.29
& 1.49 $\pm$ 0.69
& 0.01 $\pm$ 0.03
& 0.20 $\pm$ 0.05 \\

EP w/ SAM-D1
& 86.82 $\pm$ 4.30
& 4.38 $\pm$ 1.84
& 1.44 $\pm$ 0.60
& 0.29 $\pm$ 0.33
& 0.21 $\pm$ 0.05 \\

\rowcolor{SeaGreen3!30!}
ANTs SyN
& 88.77 $\pm$ 3.68
& 3.92 $\pm$ 1.63
& 1.22 $\pm$ 0.49
& 0.00 $\pm$ 0.01
& 0.21 $\pm$ 0.05 \\

\rowcolor{SeaGreen3!30!}
EP w/ SAM-D1
& 87.48 $\pm$ 4.33
& 4.35 $\pm$ 2.08
& 1.37 $\pm$ 0.60
& 0.35 $\pm$ 0.44
& 0.25 $\pm$ 0.06 \\

\rowcolor{SeaGreen3!30!}
EP w/ SAM-D1 $^*$
& 90.20 $\pm$ 2.83
& 3.30 $\pm$ 1.33
& 1.04 $\pm$ 0.39
& 0.35 $\pm$ 0.41
& 0.26 $\pm$ 0.08 \\

\addlinespace[0.4em]
\multicolumn{6}{@{}l}{\textbf{3. OASIS-1}} \\
\midrule

ANTs SyN
& 72.30 $\pm$ 5.82
& --
& --
& 0.02 $\pm$ 0.03
& 0.33 $\pm$ 0.07 \\

EP w/ MedSAM-D6
& 77.70 $\pm$ 3.79
& --
& --
& 1.26 $\pm$ 0.46
& 0.54 $\pm$ 0.08 \\

\rowcolor{SeaGreen3!30!}
ANTs SyN
& 76.03 $\pm$ 4.54
& --
& --
& 0.00 $\pm$ 0.01
& 0.46 $\pm$ 0.06 \\

\rowcolor{SeaGreen3!30!}
EP w/ MedSAM-D6
& 77.54 $\pm$ 3.91
& --
& --
& 0.92 $\pm$ 0.38
& 0.53 $\pm$ 0.08 \\

\addlinespace[0.4em]
\multicolumn{6}{@{}l}{\textbf{4. MindBoggle-101}} \\
\midrule

ANTs SyN
& 52.35 $\pm$ 3.76
& 7.13 $\pm$ 0.81
& 1.45 $\pm$ 0.26
& $<0.00001$
& 0.26 $\pm$ 0.03 \\

EP w/ SAM-D12 (Coronal)
& 58.09 $\pm$ 3.25
& 6.53 $\pm$ 0.59
& 1.22 $\pm$ 0.15
& 0.87 $\pm$ 0.23
& 0.54 $\pm$ 0.06 \\

\rowcolor{SeaGreen3!30!}
ANTs SyN
& 58.19 $\pm$ 2.26
& 6.31 $\pm$ 0.61
& 1.16 $\pm$ 0.15
& $<0.00001$
& 0.27 $\pm$ 0.03 \\

\rowcolor{SeaGreen3!30!}
EP w/ SAM-D12 (Coronal)
& 59.24 $\pm$ 3.44
& 6.42 $\pm$ 0.60
& 1.17 $\pm$ 0.15
& 0.40 $\pm$ 0.14
& 0.50 $\pm$ 0.06 \\

\addlinespace[0.4em]
\multicolumn{6}{@{}l}{\textbf{5. Lung-CT}} \\
\midrule

ANTs SyN
& 88.64 $\pm$ 6.14
& 25.03 $\pm$ 12.78
& 3.96 $\pm$ 2.35
& $\approx 0.0005$
& 0.31 $\pm$ 0.06 \\

EP w/ DINOv2-S-D1 (Coronal)
& 96.18 $\pm$ 1.08
& 6.25 $\pm$ 2.55
& 1.16 $\pm$ 0.39
& 0.72 $\pm$ 0.77
& 3.79 $\pm$ 0.53 \\

\rowcolor{SeaGreen3!30!}
ANTs SyN
& 94.93 $\pm$ 2.32
& 17.35 $\pm$ 14.23
& 1.80 $\pm$ 1.09
& 0.0
& 0.37 $\pm$ 0.08 \\

\rowcolor{SeaGreen3!30!}
EP w/ DINOv2-S-D1 (Coronal)
& 95.56 $\pm$ 1.66
& 7.17 $\pm$ 5.80
& 1.41 $\pm$ 0.81
& 0.42 $\pm$ 0.84
& 5.93 $\pm$ 4.58 \\

\addlinespace[0.4em]
\multicolumn{6}{@{}l}{\textbf{6. Abdomen-CT}} \\
\midrule

ANTs SyN
& 42.01 $\pm$ 18.42
& 40.50 $\pm$ 27.88
& 13.23 $\pm$ 12.14
& 0.03 $\pm$ 0.09
& 0.36 $\pm$ 0.06 \\

EP w/ FCN-L1 (Sagittal)
& 51.93 $\pm$ 12.69
& 33.63 $\pm$ 19.32
& 8.87 $\pm$ 6.51
& 3.22 $\pm$ 1.71
& 0.64 $\pm$ 0.08 \\

\rowcolor{SeaGreen3!30!}
ANTs SyN
& 58.26 $\pm$ 13.82
& 32.99 $\pm$ 25.78
& 7.97 $\pm$ 7.69
& 0.11 $\pm$ 0.36
& 0.50 $\pm$ 0.09 \\

\rowcolor{SeaGreen3!30!}
EP w/ FCN-L1 (Sagittal)
& 52.82 $\pm$ 12.65
& 33.75 $\pm$ 19.62
& 8.83 $\pm$ 6.54
& 1.39 $\pm$ 1.04
& 0.63 $\pm$ 0.08 \\

\rowcolor{SeaGreen3!30!}
EP w/ FCN-L1 (Sagittal) $^*$
& 61.87 $\pm$ 12.64
& 28.56 $\pm$ 16.93
& 6.68 $\pm$ 6.78
& 1.92 $\pm$ 1.49
& 0.68 $\pm$ 0.11 \\

\addlinespace[0.4em]
\multicolumn{6}{@{}l}{\textbf{7. BCV-CT}} \\
\midrule

ANTs SyN
& 57.32 $\pm$ 17.98
& 17.15 $\pm$ 5.57
& 4.86 $\pm$ 2.46
& 0.03 $\pm$ 0.15
& 0.33 $\pm$ 0.04 \\

EP w/ SAM-D9 (Coronal)
& 64.47 $\pm$ 14.51
& 17.07 $\pm$ 5.45
& 4.30 $\pm$ 2.24
& 3.44 $\pm$ 1.68
& 0.67 $\pm$ 0.08 \\

\rowcolor{SeaGreen3!30!}
ANTs SyN
& 74.01 $\pm$ 11.33
& 13.36 $\pm$ 5.26
& 2.89 $\pm$ 1.53
& 0.02 $\pm$ 0.06
& 0.44 $\pm$ 0.06 \\

\rowcolor{SeaGreen3!30!}
EP w/ SAM-D9 (Coronal)
& 71.16 $\pm$ 11.46
& 15.27 $\pm$ 4.95
& 3.48 $\pm$ 1.72
& 2.41 $\pm$ 0.99
& 0.78 $\pm$ 0.07 \\

\rowcolor{SeaGreen3!30!}
EP w/ SAM-D9 (Coronal) $^*$
& 77.17 $\pm$ 11.88
& 13.82 $\pm$ 5.60
& 2.94 $\pm$ 1.96
& 3.08 $\pm$ 1.31
& 0.80 $\pm$ 0.08 \\

\addlinespace[0.4em]
\multicolumn{6}{@{}l}{\textbf{8. CHAOS-MR}} \\
\midrule

ANTs SyN
& 57.33 $\pm$ 12.20
& 20.03 $\pm$ 5.57
& 6.02 $\pm$ 2.44
& 0.01 $\pm$ 0.01
& 0.20 $\pm$ 0.04 \\

EP w/ MedSAM-D12 (Sagittal)
& 68.51 $\pm$ 9.36
& 17.84 $\pm$ 5.40
& 4.47 $\pm$ 1.83
& 0.49 $\pm$ 0.28
& 0.55 $\pm$ 0.15 \\

\rowcolor{SeaGreen3!30!}
ANTs SyN
& 71.22 $\pm$ 10.87
& 16.07 $\pm$ 6.41
& 4.15 $\pm$ 2.57
& 0.00 $\pm$ 0.01
& 0.27 $\pm$ 0.05 \\

\rowcolor{SeaGreen3!30!}
EP w/ MedSAM-D12 (Sagittal)
& 70.27 $\pm$ 9.05
& 17.19 $\pm$ 5.33
& 4.21 $\pm$ 1.73
& 0.14 $\pm$ 0.10
& 0.49 $\pm$ 0.13 \\

\rowcolor{SeaGreen3!30!}
EP w/ MedSAM-D12 (Sagittal) $^*$
& 73.61 $\pm$ 9.78
& 15.74 $\pm$ 5.80
& 3.76 $\pm$ 1.93
& 0.17 $\pm$ 0.14
& 0.47 $\pm$ 0.13 \\

\addlinespace[0.4em]
\multicolumn{6}{@{}l}{\textbf{9. BCV--CHAOS}} \\
\midrule

ANTs SyN
& 34.68 $\pm$ 14.60
& 29.21 $\pm$ 8.56
& 10.65 $\pm$ 4.60
& 0.00 $\pm$ 0.02
& 0.29 $\pm$ 0.07 \\

EP w/ DINOv3-B-D9 (Coronal)
& 52.96 $\pm$ 12.90
& 21.74 $\pm$ 6.62
& 6.61 $\pm$ 2.75
& 9.06 $\pm$ 1.12
& 1.45 $\pm$ 0.07 \\

\rowcolor{SeaGreen3!30!}
ANTs SyN
& 39.10 $\pm$ 14.84
& 27.99 $\pm$ 8.06
& 9.67 $\pm$ 4.33
& 0.00 $\pm$ 0.01
& 0.44 $\pm$ 0.05 \\

\rowcolor{SeaGreen3!30!}
EP w/ DINOv3-B-D9 (Coronal)
& 50.81 $\pm$ 12.64
& 21.88 $\pm$ 6.81
& 6.77 $\pm$ 2.77
& 7.20 $\pm$ 1.06
& 1.36 $\pm$ 0.10 \\

\addlinespace[0.4em]
\multicolumn{6}{@{}l}{\textbf{10. Abdomen-CT--MR}} \\
\midrule

ANTs SyN
& 63.71 $\pm$ 17.73
& 11.47 $\pm$ 5.43
& 3.64 $\pm$ 2.10
& $\approx 0.0005$
& 0.26 $\pm$ 0.03 \\

EP w/ DINOv3-B-D9 (Axial)
& 87.26 $\pm$ 4.61
& 11.51 $\pm$ 6.20
& 2.64 $\pm$ 1.22
& 1.42 $\pm$ 0.90
& 0.51 $\pm$ 0.08 \\

\rowcolor{SeaGreen3!30!}
ANTs SyN
& 75.41 $\pm$ 9.09
& 8.38 $\pm$ 3.00
& 2.24 $\pm$ 0.87
& $\approx 0.00004$
& 0.32 $\pm$ 0.06 \\

\rowcolor{SeaGreen3!30!}
EP w/ DINOv3-B-D9 (Axial)
& 86.67 $\pm$ 4.66
& 11.65 $\pm$ 6.25
& 2.73 $\pm$ 1.27
& 0.94 $\pm$ 0.65
& 0.54 $\pm$ 0.07 \\

\end{longtable}

\endgroup

\newpage
\section{Practical feature-selection details}
\label{sec:supp:pratical_advice}
In our experiments, a small shortlist of candidate features was usually sufficient to identify a strong configuration. Intermediate features such as FCN-L1, SAM-D3, DINOv2 mid-level features and DINOv3 mid-level features were among the more reliable starting points, whereas the deepest features and SAM2 features were often less effective. Because no single feature family was universally optimal, we recommend using \textit{Easy Peeler} as a search-and-validate framework: choose a compact candidate set, validate on a small development split, and select the feature map that improves Dice without increasing boundary errors or foldings.
\newpage

\newpage
\section{Additional Methodological Details}
\label{sec:supp:trainingdetails}
\begin{table}[h]
\centering
\setlength{\tabcolsep}{2.5pt}
\caption{Dataset-specific training configurations. All models used diffusion regularization with the specified $\lambda$ values. $C_g$ is the channel dimensions of $g^{\mathrm{ins}}_M$, $g^{\mathrm{str}}_M$, $g^{\mathrm{ins}}_F$, and $g^{\mathrm{str}}_F$.}
\label{tab:training_config}
\begin{tabular}{lcccc}
\toprule
Dataset & $C_g$ & Similarity Metric & $\lambda$ & Epochs \\
\midrule
Fluo-N2DH-GOWT1 & 36 & MSE & Model-specific$^*$ & 500 \\
ACDC & 32& MSE & 0.01 & 500 \\
OASIS-1 & 48 & MSE & 0.005 & 10$^\dagger$ \\
MindBoggle-101 & 32& MSE & 0.05 & 500 \\
Lung-CT & 32& MSE & 0.001 & 500$^\star$ \\
Abdomen-CT & 32& LNCC & 1.0 & 500 \\
BCV-CT & 32& LNCC & 1.0 & 500 \\
CHAOS-MR & 32& MSE & 0.05 & 500 \\
BCV--CHAOS & 32& MI & 0.5 & 500 \\
Abdomen-CT--MR & 32 & MI + MIND & 0.105 & 500 \\
\bottomrule
\end{tabular}
\end{table}
\noindent\textsuperscript{$^*$}For Fluo-N2DH-GOWT1, $\lambda$ was individually tuned for each model from $\{0.001, 0.0005, 0.0001, 0.00005, 0.00001\}$ based on validation performance.\\
\textsuperscript{$\dagger$}OASIS-1 training was limited to 10 epochs due to the large number of available training pairs.\\
\textsuperscript{$\star$}Lung-CT employed SVF diffeomorphic registration with scaling-and-squaring integration.\\
\textsuperscript{$\ddagger$}For the Abdomen-CT--MR dataset, the combined similarity loss was defined as ${\mathcal L}_{S} = {\mathcal L}_{MI} + 0.05 \cdot {\mathcal L}_{MIND}$, with weights determined through empirical validation to balance the contributions of global statistical alignment and local feature matching.



\end{appendices}
\newpage

\bibliography{sn-bibliography}

@dataset{akin_2016_tcga_kirc,
  author    = {Akin, O. and Elnajjar, P. and Heller, M. and Jarosz, R. and Erickson, B. J. and Kirk, S. and Lee, Y. and Linehan, M. W. and Gautam, R. and Vikram, R. and Garcia, K. M. and Roche, C. and Bonaccio, E. and Filippini, J.},
  title     = {The Cancer Genome Atlas Kidney Renal Clear Cell Carcinoma Collection (TCGA-KIRC)},
  year      = {2016},
  version   = {3},
  publisher = {The Cancer Imaging Archive},
  doi       = {10.7937/K9/TCIA.2016.V6PBVTDR},
  url       = {https://doi.org/10.7937/K9/TCIA.2016.V6PBVTDR},
  note      = {Data set}
}

@dataset{linehan_2016_tcga_kirp,
  author    = {Linehan, M. and Gautam, R. and Kirk, S. and Lee, Y. and Roche, C. and Bonaccio, E. and Filippini, J. and Rieger-Christ, K. and Lemmerman, J. and Jarosz, R.},
  title     = {The Cancer Genome Atlas Cervical Kidney Renal Papillary Cell Carcinoma Collection (TCGA-KIRP)},
  year      = {2016},
  version   = {4},
  publisher = {The Cancer Imaging Archive},
  doi       = {10.7937/K9/TCIA.2016.ACWOGBEF},
  url       = {https://doi.org/10.7937/K9/TCIA.2016.ACWOGBEF},
  note      = {Data set}
}

@dataset{erickson_2016_tcga_lihc,
  author    = {Erickson, B. J. and Kirk, S. and Lee, Y. and Bathe, O. and Kearns, M. and Gerdes, C. and Rieger-Christ, K. and Lemmerman, J.},
  title     = {The Cancer Genome Atlas Liver Hepatocellular Carcinoma Collection (TCGA-LIHC)},
  year      = {2016},
  version   = {5},
  publisher = {The Cancer Imaging Archive},
  doi       = {10.7937/K9/TCIA.2016.IMMQW8UQ},
  url       = {https://doi.org/10.7937/K9/TCIA.2016.IMMQW8UQ},
  note      = {Data set}
}

@inproceedings{kuang2019faim,
  title={Faim--a convnet method for unsupervised 3d medical image registration},
  author={Kuang, Dongyang and Schmah, Tanya},
  booktitle={Machine Learning in Medical Imaging: 10th International Workshop, MLMI 2019, Held in Conjunction with MICCAI 2019, Shenzhen, China, October 13, 2019, Proceedings 10},
  pages={646--654},
  year={2019},
  organization={Springer}
}

@inproceedings{he2016deep,
  title={Deep residual learning for image recognition},
  author={He, Kaiming and Zhang, Xiangyu and Ren, Shaoqing and Sun, Jian},
  booktitle={Proceedings of the IEEE conference on computer vision and pattern recognition},
  pages={770--778},
  year={2016}
}

@article{kavur2019chaos,
  title={CHAOS-combined (CT-MR) healthy abdominal organ segmentation challenge data},
  author={Kavur, Ali Emre and Selver, M Alper and Dicle, O{\u{g}}uz and Bar{\i}{\c{s}}, Mustafa and Gezer, N Sinem},
  journal={(No Title)},
  year={2019},
  publisher={Zenodo}
}

@article{zheng2024residual,
  title={Residual aligner-based network (ran): Motion-separable structure for coarse-to-fine discontinuous deformable registration},
  author={Zheng, Jian-Qing and Wang, Ziyang and Huang, Baoru and Lim, Ngee Han and Papie{\.z}, Bart{\l}omiej W},
  journal={Medical Image Analysis},
  volume={91},
  pages={103038},
  year={2024},
  publisher={Elsevier}
}

@article{xu2016evaluation,
  title={Evaluation of six registration methods for the human abdomen on clinically acquired CT},
  author={Xu, Zhoubing and Lee, Christopher P and Heinrich, Mattias P and Modat, Marc and Rueckert, Daniel and Ourselin, Sebastien and Abramson, Richard G and Landman, Bennett A},
  journal={IEEE Transactions on Biomedical Engineering},
  volume={63},
  number={8},
  pages={1563--1572},
  year={2016},
  publisher={IEEE}
}

@article{mavska2023cell,
  title={The cell tracking challenge: 10 years of objective benchmarking},
  author={Ma{\v{s}}ka, Martin and Ulman, Vladim{\'\i}r and Delgado-Rodriguez, Pablo and G{\'o}mez-de-Mariscal, Estibaliz and Ne{\v{c}}asov{\'a}, Tereza and Guerrero Pe{\~n}a, Fidel A and Ren, Tsang Ing and Meyerowitz, Elliot M and Scherr, Tim and L{\"o}ffler, Katharina and others},
  journal={Nature Methods},
  volume={20},
  number={7},
  pages={1010--1020},
  year={2023},
  publisher={Nature Publishing Group US New York}
}

@article{bartova2011recruitment,
  title={Recruitment of Oct4 protein to UV-damaged chromatin in embryonic stem cells},
  author={B{\'a}rtov{\'a}, Eva and {\v{S}}ust{\'a}{\v{c}}kov{\'a}, Gabriela and Stixov{\'a}, Lenka and Kozubek, Stanislav and Legartov{\'a}, So{\v{n}}a and Folt{\'a}nkov{\'a}, Veronika},
  journal={PLoS One},
  volume={6},
  number={12},
  pages={e27281},
  year={2011},
  publisher={Public Library of Science San Francisco, USA}
}

@article{liu2023contrastive,
  title={Contrastive registration for unsupervised medical image segmentation},
  author={Liu, Lihao and Aviles-Rivero, Angelica I and Sch{\"o}nlieb, Carola-Bibiane},
  journal={IEEE Transactions on Neural Networks and Learning Systems},
  year={2023},
  publisher={IEEE}
}

@article{klein2012101,
  title={101 labeled brain images and a consistent human cortical labeling protocol},
  author={Klein, Arno and Tourville, Jason},
  journal={Frontiers in neuroscience},
  volume={6},
  pages={171},
  year={2012},
  publisher={Frontiers Media SA}
}

@article{beg2005computing,
  title={Computing large deformation metric mappings via geodesic flows of diffeomorphisms},
  author={Beg, M Faisal and Miller, Michael I and Trouv{\'e}, Alain and Younes, Laurent},
  journal={International Journal of Computer Vision},
  volume={61},
  number={2},
  pages={139--157},
  year={2005},
  publisher={Springer}
}

@article{ashburner2007fast,
  title={A fast diffeomorphic image registration algorithm},
  author={Ashburner, John},
  journal={NeuroImage},
  volume={38},
  number={1},
  pages={95--113},
  year={2007},
  publisher={Elsevier}
}

@ARTICLE{balakrishnan2019voxelmorph,
author={G. {Balakrishnan} and A. {Zhao} and M. R. {Sabuncu} and J. {Guttag} and A. V. {Dalca}},
journal={IEEE Transactions on Medical Imaging},
title={VoxelMorph: A Learning Framework for Deformable Medical Image Registration},
year={2019},
volume={38},
number={8},
pages={1788-1800},
ISSN={1558-254X},
month={Aug},}

@inproceedings{kirillov2023segment,
  title={Segment anything},
  author={Kirillov, Alexander and Mintun, Eric and Ravi, Nikhila and Mao, Hanzi and Rolland, Chloe and Gustafson, Laura and Xiao, Tete and Whitehead, Spencer and Berg, Alexander C and Lo, Wan-Yen and others},
  booktitle={Proceedings of the IEEE/CVF international conference on computer vision},
  pages={4015--4026},
  year={2023}
}

@inproceedings{jia2023fourier,
  title={Fourier-net: Fast image registration with band-limited deformation},
  author={Jia, Xi and Bartlett, Joseph and Chen, Wei and Song, Siyang and Zhang, Tianyang and Cheng, Xinxing and Lu, Wenqi and Qiu, Zhaowen and Duan, Jinming},
  booktitle={Proceedings of the AAAI Conference on Artificial Intelligence},
  volume={37},
  number={1},
  pages={1015--1023},
  year={2023}
}

@article{ravi2024sam,
  title={Sam 2: Segment anything in images and videos},
  author={Ravi, Nikhila and Gabeur, Valentin and Hu, Yuan-Ting and Hu, Ronghang and Ryali, Chaitanya and Ma, Tengyu and Khedr, Haitham and R{\"a}dle, Roman and Rolland, Chloe and Gustafson, Laura and others},
  journal={arXiv preprint arXiv:2408.00714},
  year={2024}
}

@article{chen2017rethinking,
  title={Rethinking atrous convolution for semantic image segmentation},
  author={Chen, Liang-Chieh and Papandreou, George and Schroff, Florian and Adam, Hartwig},
  journal={arXiv preprint arXiv:1706.05587},
  year={2017}
}

@article{heinrich2012mind,
  title={MIND: Modality independent neighbourhood descriptor for multi-modal deformable registration},
  author={Heinrich, Mattias P and Jenkinson, Mark and Bhushan, Manav and Matin, Tahreema and Gleeson, Fergus V and Brady, Michael and Schnabel, Julia A},
  journal={Medical image analysis},
  volume={16},
  number={7},
  pages={1423--1435},
  year={2012},
  publisher={Elsevier}
}

@article{kavur2021chaos,
  title={CHAOS challenge-combined (CT-MR) healthy abdominal organ segmentation},
  author={Kavur, A Emre and Gezer, N Sinem and Bar{\i}{\c{s}}, Mustafa and Aslan, Sinem and Conze, Pierre-Henri and Groza, Vladimir and Pham, Duc Duy and Chatterjee, Soumick and Ernst, Philipp and {\"O}zkan, Sava{\c{s}} and others},
  journal={Medical Image Analysis},
  volume={69},
  pages={101950},
  year={2021},
  publisher={Elsevier}
}

@article{song2025dino,
  title={Dino-Reg: Efficient Multimodal Image Registration with Distilled Features},
  author={Song, Xinrui and Xu, Xuanang and Zhang, Jiajin and Reyes, Diego Machado and Yan, Pingkun},
  journal={IEEE Transactions on Medical Imaging},
  year={2025},
  publisher={IEEE}
}

@article{baker2011database,
  title={A database and evaluation methodology for optical flow},
  author={Baker, Simon and Scharstein, Daniel and Lewis, James P and Roth, Stefan and Black, Michael J and Szeliski, Richard},
  journal={International journal of computer vision},
  volume={92},
  pages={1--31},
  year={2011},
  publisher={Springer}
}

@article{bernard2018deep,
  title={Deep learning techniques for automatic MRI cardiac multi-structures segmentation and diagnosis: is the problem solved?},
  author={Bernard, Olivier and Lalande, Alain and Zotti, Clement and Cervenansky, Frederick and Yang, Xin and Heng, Pheng-Ann and Cetin, Irem and Lekadir, Karim and Camara, Oscar and Ballester, Miguel Angel Gonzalez and others},
  journal={IEEE Transactions on Medical Imaging},
  volume={37},
  number={11},
  pages={2514--2525},
  year={2018},
  publisher={ieee}
}

@article{hering2022learn2reg,
  title={Learn2Reg: comprehensive multi-task medical image registration challenge, dataset and evaluation in the era of deep learning},
  author={Hering, Alessa and Hansen, Lasse and Mok, Tony CW and Chung, Albert CS and Siebert, Hanna and H{\"a}ger, Stephanie and Lange, Annkristin and Kuckertz, Sven and Heldmann, Stefan and Shao, Wei and others},
  journal={IEEE Transactions on Medical Imaging},
  year={2022},
  publisher={IEEE}
}

@article{marcus2007open,
  title={Open Access Series of Imaging Studies (OASIS): cross-sectional MRI data in young, middle aged, nondemented, and demented older adults},
  author={Marcus, Daniel S and Wang, Tracy H and Parker, Jamie and Csernansky, John G and Morris, John C and Buckner, Randy L},
  journal={Journal of cognitive neuroscience},
  volume={19},
  year={2007},
  publisher={MIT Press One Rogers Street, Cambridge, MA 02142-1209, USA journals-info~…}
}

@inproceedings{hoopes2021hypermorph,
  title={Hypermorph: Amortized hyperparameter learning for image registration},
  author={Hoopes, Andrew and Hoffmann, Malte and Fischl, Bruce and Guttag, John and Dalca, Adrian V},
  booktitle={International Conference on Information Processing in Medical Imaging},
  pages={3--17},
  year={2021},
  organization={Springer}
}

@inproceedings{long2015fully,
  title={Fully convolutional networks for semantic segmentation},
  author={Long, Jonathan and Shelhamer, Evan and Darrell, Trevor},
  booktitle={Proceedings of the IEEE conference on computer vision and pattern recognition},
  pages={3431--3440},
  year={2015}
}

@article{jia2025decoder,
  title={Decoder-only image registration},
  author={Jia, Xi and Lu, Wenqi and Cheng, Xinxing and Duan, Jinming},
  journal={IEEE Transactions on Medical Imaging},
  year={2025},
  publisher={IEEE}
}

@article{ma2024segment,
  title={Segment anything in medical images},
  author={Ma, Jun and He, Yuting and Li, Feifei and Han, Lin and You, Chenyu and Wang, Bo},
  journal={Nature Communications},
  volume={15},
  number={1},
  pages={654},
  year={2024},
  publisher={Nature Publishing Group UK London}
}

@article{simeoni2025dinov3,
  title={Dinov3},
  author={Sim{\'e}oni, Oriane and Vo, Huy V and Seitzer, Maximilian and Baldassarre, Federico and Oquab, Maxime and Jose, Cijo and Khalidov, Vasil and Szafraniec, Marc and Yi, Seungeun and Ramamonjisoa, Micha{\"e}l and others},
  journal={arXiv preprint arXiv:2508.10104},
  year={2025}
}

@article{allen2024prospective,
  title={Prospective study design and data analysis in UK Biobank},
  author={Allen, Naoimi and Lacey, Ben and Lawlor, Deborah and Pell, Jill and Gallacher, John and Smeeth, Liam and Elliott, Paul and Matthews, Paul and Lyons, Ronan and Whetton, Anthony and others},
  journal={Science translational medicine},
  volume={16},
  number={729},
  year={2024},
  publisher={American Association for the Advancement of Science}
}

@article{cao2026all,
  title={An all-in-one foundation model for multimodal medical image restoration and enhancement},
  author={Cao, Luyang and Wu, Zhijian and Zhang, Jian and Qi, Lei and Gao, Yang and Zheng, Hairong and Zheng, Yefeng and Shi, Yinghuan},
  journal={Nature Machine Intelligence},
  pages={1--20},
  year={2026},
  publisher={Nature Publishing Group UK, London}
}

@inproceedings{zeiler2014visualizing,
  title={Visualizing and understanding convolutional networks},
  author={Zeiler, Matthew D and Fergus, Rob},
  booktitle={European conference on computer vision},
  pages={818--833},
  year={2014},
  organization={Springer}
}

@article{zhao2025foundation,
  title={A foundation model for joint segmentation, detection and recognition of biomedical objects across nine modalities},
  author={Zhao, Theodore and Gu, Yu and Yang, Jianwei and Usuyama, Naoto and Lee, Ho Hin and Kiblawi, Sid and Naumann, Tristan and Gao, Jianfeng and Crabtree, Angela and Abel, Jacob and others},
  journal={Nature methods},
  volume={22},
  number={1},
  pages={166--176},
  year={2025},
  publisher={Nature Publishing Group US, New York}
}

@article{avants2008symmetric,
  title={Symmetric diffeomorphic image registration with cross-correlation: evaluating automated labeling of elderly and neurodegenerative brain},
  author={Avants, Brian B and Epstein, Charles L and Grossman, Murray and Gee, James C},
  journal={Medical image analysis},
  volume={12},
  number={1},
  pages={26--41},
  year={2008},
  publisher={Elsevier}
}

@inproceedings{mok2020large,
  title={Large deformation diffeomorphic image registration with laplacian pyramid networks},
  author={Mok, Tony CW and Chung, Albert},
  booktitle={International Conference on Medical Image Computing and Computer-Assisted Intervention},
  pages={211--221},
  year={2020},
  organization={Springer}
}

@InProceedings{Zhao_2019_ICCV,
author = {Zhao, Shengyu and Dong, Yue and Chang, Eric I-Chao and Xu, Yan},
title = {Recursive Cascaded Networks for Unsupervised Medical Image Registration},
booktitle = {The IEEE International Conference on Computer Vision (ICCV)},
month = {October},
year = {2019}
}

@inproceedings{arsigny2006log,
  title={A log-euclidean framework for statistics on diffeomorphisms},
  author={Arsigny, Vincent and Commowick, Olivier and Pennec, Xavier and Ayache, Nicholas},
  booktitle={International Conference on Medical Image Computing and Computer-Assisted Intervention},
  pages={924--931},
  year={2006},
  organization={Springer}
}

@inproceedings{tian2024unigradicon,
  title={unigradicon: A foundation model for medical image registration},
  author={Tian, Lin and Greer, Hastings and Kwitt, Roland and Vialard, Fran{\c{c}}ois-Xavier and San Jos{\'e} Est{\'e}par, Ra{\'u}l and Bouix, Sylvain and Rushmore, Richard and Niethammer, Marc},
  booktitle={International Conference on Medical Image Computing and Computer-Assisted Intervention},
  pages={749--760},
  year={2024},
  organization={Springer}
}

@article{chen2025survey,
  title={A survey on deep learning in medical image registration: New technologies, uncertainty, evaluation metrics, and beyond},
  author={Chen, Junyu and Liu, Yihao and Wei, Shuwen and Bian, Zhangxing and Subramanian, Shalini and Carass, Aaron and Prince, Jerry L and Du, Yong},
  journal={Medical Image Analysis},
  volume={100},
  pages={103385},
  year={2025},
  publisher={Elsevier}
}

@misc{schnabel2016advances,
  title={Advances and challenges in deformable image registration: From image fusion to complex motion modelling},
  author={Schnabel, Julia A and Heinrich, Mattias P and Papie{\.z}, Bart{\l}omiej W and Brady, J Michael},
  journal={Medical image analysis},
  volume={33},
  pages={145--148},
  year={2016},
  publisher={Elsevier}
}

@article{ganjgahi2025ai,
  title={AI-driven reclassification of multiple sclerosis progression},
  author={Ganjgahi, Habib and H{\"a}ring, Dieter A and Aarden, Piet and Graham, Gordon and Sun, Yang and Gardiner, Stephen and Su, Wendy and Berge, Claude and Bischof, Antje and Fisher, Elizabeth and others},
  journal={Nature Medicine},
  volume={31},
  number={10},
  pages={3414--3424},
  year={2025},
  publisher={Nature Publishing Group US New York}
}

@article{xie2025multi,
  title={A multi-cohort study of longitudinal and cross-sectional Alzheimer's disease biomarkers in cognitively unimpaired older adults},
  author={Xie, Long and Das, Sandhitsu R and Li, Yue and Wisse, Laura EM and McGrew, Emily and Lyu, Xueying and DiCalogero, Michael and Shah, Ujashi and Ilesanmi, Ademola and Denning, Amanda E and others},
  journal={Alzheimer's \& Dementia},
  volume={21},
  number={2},
  pages={e14492},
  year={2025},
  publisher={Wiley Online Library}
}

@article{namburete2023normative,
  title={Normative spatiotemporal fetal brain maturation with satisfactory development at 2 years},
  author={Namburete, Ana IL and Papie{\.z}, Bart{\l}omiej W and Fernandes, Michelle and Wyburd, Madeleine K and Hesse, Linde S and Moser, Felipe A and Ismail, Leila Cheikh and Gunier, Robert B and Squier, Waney and Ohuma, Eric O and others},
  journal={Nature},
  volume={623},
  number={7985},
  pages={106--114},
  year={2023},
  publisher={Nature Publishing Group UK London}
}

@dataset{hering2020learn2reg,
  author    = {Hering, Alessa and Murphy, Keelin and van Ginneken, Bram},
  title     = {Learn2Reg Challenge: CT Lung Registration - Training Data},
  publisher = {Zenodo},
  year      = {2020},
  month     = may,
  day       = {20},
  doi       = {10.5281/zenodo.3835682},
  url       = {https://doi.org/10.5281/zenodo.3835682}
}

@article{oquab2023dinov2,
  title={Dinov2: Learning robust visual features without supervision},
  author={Oquab, Maxime and Darcet, Timoth{\'e}e and Moutakanni, Th{\'e}o and Vo, Huy and Szafraniec, Marc and Khalidov, Vasil and Fernandez, Pierre and Haziza, Daniel and Massa, Francisco and El-Nouby, Alaaeldin and others},
  journal={arXiv preprint arXiv:2304.07193},
  year={2023}
}

@inproceedings{jia2022u,
  title={U-net vs transformer: Is u-net outdated in medical image registration?},
  author={Jia, Xi and Bartlett, Joseph and Zhang, Tianyang and Lu, Wenqi and Qiu, Zhaowen and Duan, Jinming},
  booktitle={International Workshop on Machine Learning in Medical Imaging},
  pages={151--160},
  year={2022},
  organization={Springer}
}

@article{chen2022transmorph,
  title={Transmorph: Transformer for unsupervised medical image registration},
  author={Chen, Junyu and Frey, Eric C and He, Yufan and Segars, William P and Li, Ye and Du, Yong},
  journal={Medical Image Analysis},
  volume={82},
  pages={102615},
  year={2022},
  publisher={Elsevier}
}

@inproceedings{rueckert2006diffeomorphic,
  title={Diffeomorphic registration using B-splines},
  author={Rueckert, Daniel and Aljabar, Paul and Heckemann, Rolf A and Hajnal, Joseph V and Hammers, Alexander},
  booktitle={International Conference on Medical Image Computing and Computer-Assisted Intervention. Proceedings, Part II 9},
  pages={702--709},
  year={2006},
  organization={Springer}
}

@article{vercauteren2009diffeomorphic,
  title={Diffeomorphic demons: Efficient non-parametric image registration},
  author={Vercauteren, Tom and Pennec, Xavier and Perchant, Aymeric and Ayache, Nicholas},
  journal={NeuroImage},
  volume={45},
  number={1},
  pages={S61--S72},
  year={2009},
  publisher={Elsevier}
}

@article{sotiras2013deformable,
  title={Deformable medical image registration: A survey},
  author={Sotiras, Aristeidis and Davatzikos, Christos and Paragios, Nikos},
  journal={IEEE Transactions on Medical Imaging},
  volume={32},
  number={7},
  pages={1153--1190},
  year={2013},
  publisher={IEEE}
}

\end{document}